\documentclass[10pt,twocolumn,letterpaper]{article}

\usepackage{iccv}
\usepackage{times}
\usepackage{epsfig}
\usepackage{graphicx}
\usepackage{amsmath}
\usepackage{amssymb}
\usepackage{mathtools}

\usepackage{graphicx}
\usepackage{tikz}
\usetikzlibrary{spy}
\usepackage{comment}

\usepackage{color}

\usepackage{array}
\usepackage{tabu}
\usepackage{array}
\usepackage{booktabs}
\usepackage{soul}

\usepackage{xcolor}

\usepackage{bbm}

\usepackage{algorithmic}
\usepackage[linesnumbered,ruled,vlined]{algorithm2e}
\usepackage{graphicx}
\usepackage{caption}

\newif\ifdraft
\drafttrue
\ifdraft
\newcommand{\npc}[1]{{\color{orange}[\textbf{NP:} #1]}}
\newcommand{\dcc}[1]{{\color{purple}[\textbf{DC:} #1]}}
\newcommand{\dlc}[1]{{\color{blue}[\textbf{DL:} #1]}}
\newcommand{\dbc}[1]{{\color{teal}[\textbf{DB:} #1]}}
\newcommand{\ybc}[1]{{\color{olive}[\textbf{YB:} #1]}}
\newcommand{\arc}[1]{{\color{blue}[\textbf{AR:} #1]}}

\else
\newcommand{\npc}[1]{}
\newcommand{\dcc}[1]{}
\newcommand{\dlc}[1]{}
\newcommand{\dbc}[1]{}
\newcommand{\ybc}[1]{}
\newcommand{\arc}[1]{}

\fi

\newcommand{\T}{T^{*}}
\newcommand{\tmap}{\mathbf{\T}}
\newcommand{\esttmap}{\mathbf{\hat{T}}^{*}}
\newcommand{\estx}{\mathbf{\hat{x}_0}}
\newcommand{\y}{\mathbf{y}}
\newcommand{\x}{\mathbf{x}_0}
\newcommand{\xt}[1]{\mathbf{x}_{#1}}
\newcommand{\n}[1]{\boldsymbol{\epsilon}_{#1}}
\newcommand{\ntilde}[1]{\boldsymbol{\tilde{\epsilon}}_{#1}}
\newcommand{\nbar}[1]{\boldsymbol{\bar{\epsilon}}_{#1}}
\newcommand{\N}[1]{\mathcal{N} \left( #1 \right)}
\newcommand{\Nmat}{\N{\mathbf{0}, \mathbf{I}}}

\newcommand{\etamat}[1]{\boldsymbol{\eta_{#1}}}
\newcommand{\gammamat}[1]{\boldsymbol{\gamma}_{#1}}
\newcommand{\sigmap}{\boldsymbol{\sigma_{p}}}
\newcommand{\sigmapsquare}{\boldsymbol{\sigma}_{\boldsymbol{p}}^2}
\newcommand{\estsigmapsquare}{\boldsymbol{\hat{\sigma}}_{\boldsymbol{p}}^2}

\newcommand{\tk}[1]{\mathbf{t}_{#1}}
\newcommand{\promt}{\mathbf{\hat{t}}}

\newcommand{\svdd}{\text{SVNR}\xspace}
\newcommand{\svddalg}[1]{\text{SVNR} \! \left( #1 \right)}
\newcommand{\clip}[1]{\mathrm{clip} \left( #1 \right)}
\newcommand{\var}[1]{\mathrm{Var} \left( #1 \right)}
\newcommand{\clipzero}[1]{(#1)^+}

\newlength{\ww}

\definecolor{mygray}{RGB}{140, 140, 140}

\definecolor{dashedpurple}{RGB}{118,0,103}
\definecolor{dashedorange}{RGB}{181,100,13}

\DeclareMathOperator{\R}{\mathbb{R}}

\makeatletter
\newcommand{\thickhline}{%
    \noalign {\ifnum 0=`}\fi \hrule height 1pt
    \futurelet \reserved@a \@xhline
}
\makeatletter
\newcommand{\thickvline}{%
    \noalign {\ifnum 0=`}\fi \vrule height 1pt
    \futurelet \reserved@a \@xvline
}
\newcolumntype{"}{@{\hskip\tabcolsep\vrule width 1pt\hskip\tabcolsep}}
\makeatother

\newcommand\figref[1]{Fig.~\ref{#1}}
\newcommand\tabref[1]{Table~\ref{#1}}
\newcommand\secref[1]{Section~\ref{#1}}

\DeclareMathAlphabet\mathbfcal{OMS}{cmsy}{b}{n}

\graphicspath{{./figures/}}
\usepackage[pagebackref=true,breaklinks=true,letterpaper=true,colorlinks,bookmarks=false]{hyperref}

\iccvfinalcopy % *** Uncomment this line for the final submission

\def\iccvPaperID{6107} % *** Enter the ICCV Paper ID here

\ificcvfinal\fi

\usepackage{svg}

\newcommand*{\affaddr}[1]{#1} 
\newcommand*{\affmark}[1][*]{\textsuperscript{#1}}

\usepackage[symbol]{footmisc}

\begin{document}

%%%%%%%%% TITLE
\title{\svdd: Spatially-variant Noise Removal with Denoising Diffusion }

\author{%
Naama Pearl\footref{note1}~~\affmark[1,]\affmark[2], 
Yaron Brodsky\affmark[1], 
Dana Berman\affmark[1], 
Assaf Zomet\affmark[1],
Alex Rav Acha\affmark[1], \\
Daniel Cohen-Or\footref{note1}~~\affmark[1,]\affmark[3]
and Dani Lischinski\footref{note1}~~\affmark[1,]\affmark[4]\\
[0.5em]
% 
% \vspace{1em} 
% 
\small{\affaddr{\affmark[1]Google Research,~~}}
\small{\affaddr{\affmark[2]University of Haifa,~~}} \\
\small{\affaddr{\affmark[3]Tel Aviv University,~~}}
\small{\affaddr{\affmark[4]The Hebrew University of Jerusalem}}
% \small{\affaddr{\affmark[2]Dept. of Marine Technologies, University of Haifa,~~}} \\
% \small{\affaddr{\affmark[3]The Blavatnik School of Computer Science, Tel Aviv University,~~}}
% \small{\affaddr{\affmark[3]The Hebrew University of Jerusalem,~~}}
}

\maketitle
\ificcvfinal\thispagestyle{empty}\fi

\footnotetext[2]{\label{note1} Performed this work while working at Google.}
\begin{abstract}
Denoising diffusion models have recently shown impressive results in generative tasks.
By learning powerful priors from huge collections of training images, such models are able to gradually modify complete noise to a clean natural image via a sequence of small denoising steps, seemingly making them well-suited for single image denoising. 
However, effectively applying denoising diffusion models to removal of realistic noise is more challenging than it may seem, since their formulation is based on additive white Gaussian noise, unlike noise in real-world images.   
In this work, we present $\svdd$, a novel formulation of denoising diffusion that assumes a more realistic, spatially-variant noise model.
$\svdd$ enables using the noisy input image as the starting point for the denoising diffusion process, in addition to conditioning the process on it.
To this end, we adapt the diffusion process to allow each pixel to have its own time embedding, and propose training and inference schemes that support spatially-varying time maps.
Our formulation also accounts for the correlation that exists between the condition image and the samples along the modified diffusion process.
In our experiments we demonstrate the advantages of our approach over a strong diffusion model baseline, as well as over a state-of-the-art single image denoising method.
\end{abstract}

\section{Introduction}
\label{sec:Intro}

\begin{figure}[t]
    \centering
    \setlength{\tabcolsep}{0.5pt}
    \renewcommand{\arraystretch}{0.5}
    \setlength{\ww}{0.325\columnwidth}
    \begin{tabular}{c c c}
        % \rotatebox{90}{\phantom{AAa}\scriptsize{}} &
        % trim={<left> <lower> <right> <upper>}
        \includegraphics[width=\ww,trim={10cm 10cm 1cm 1cm},clip]{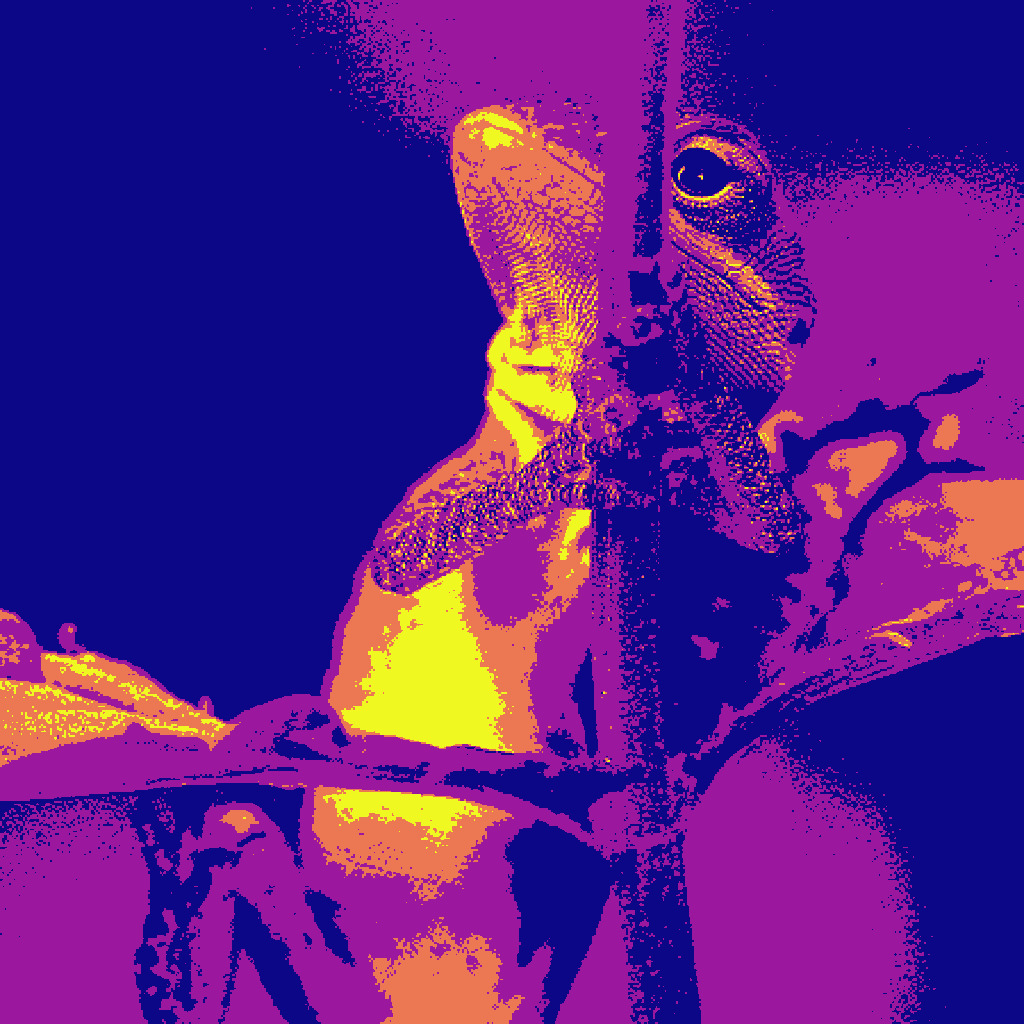}
        &
        \includegraphics[width=\ww,trim={10cm 10cm 1cm 1cm},clip]{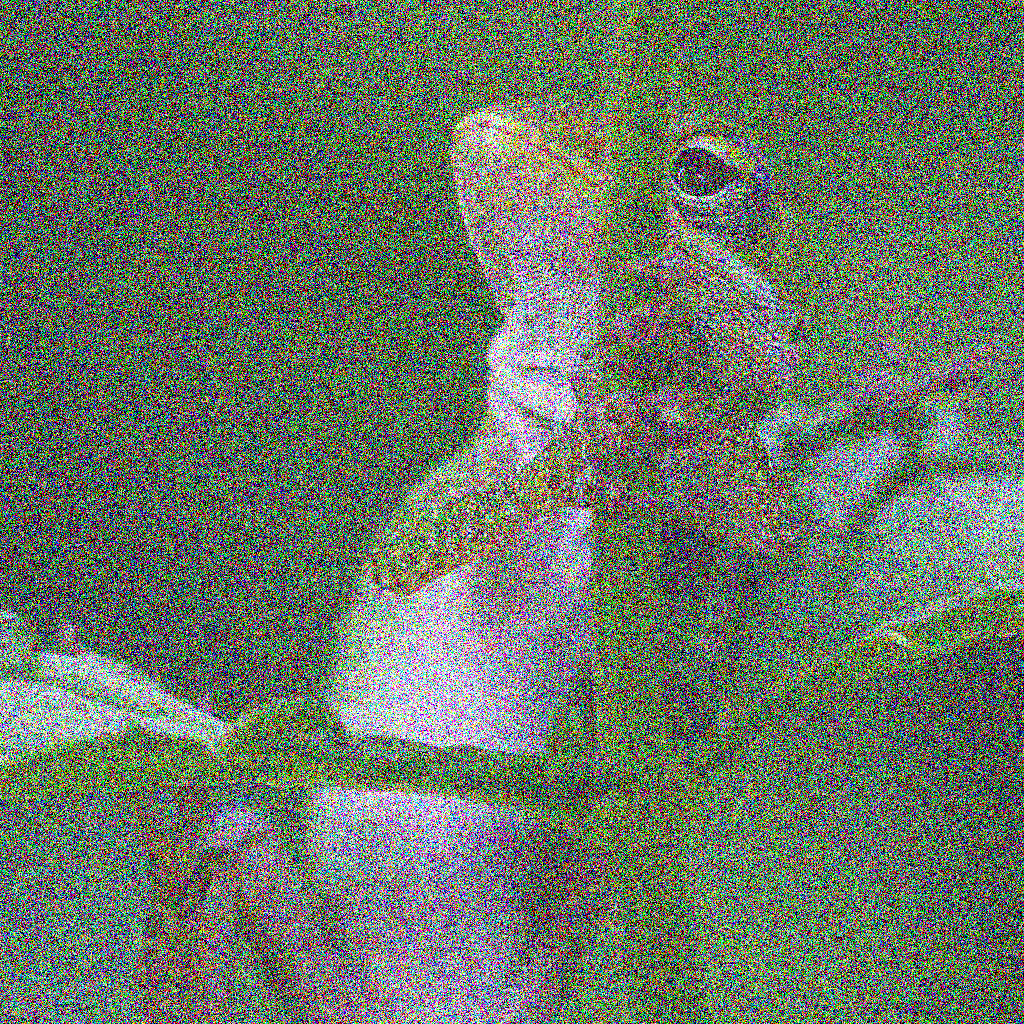}
        &
        \includegraphics[width=\ww,trim={10cm 10cm 1cm 1cm},clip]{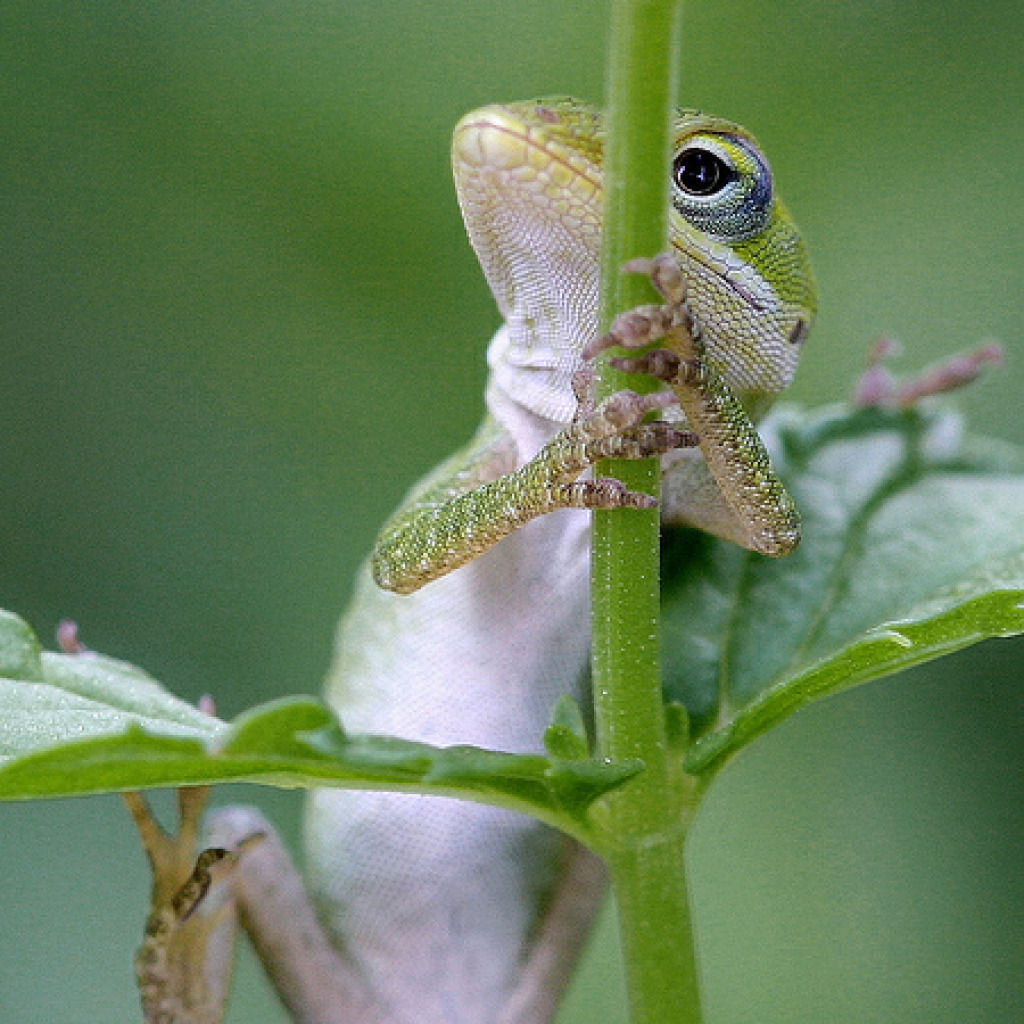}
        
        \\
        \scriptsize{Noise std} &
        \scriptsize{Noisy image} &
        \scriptsize{Clean image (ground truth)}
        \\
        \includegraphics[width=\ww,trim={10cm 10cm 1cm 1cm},clip]{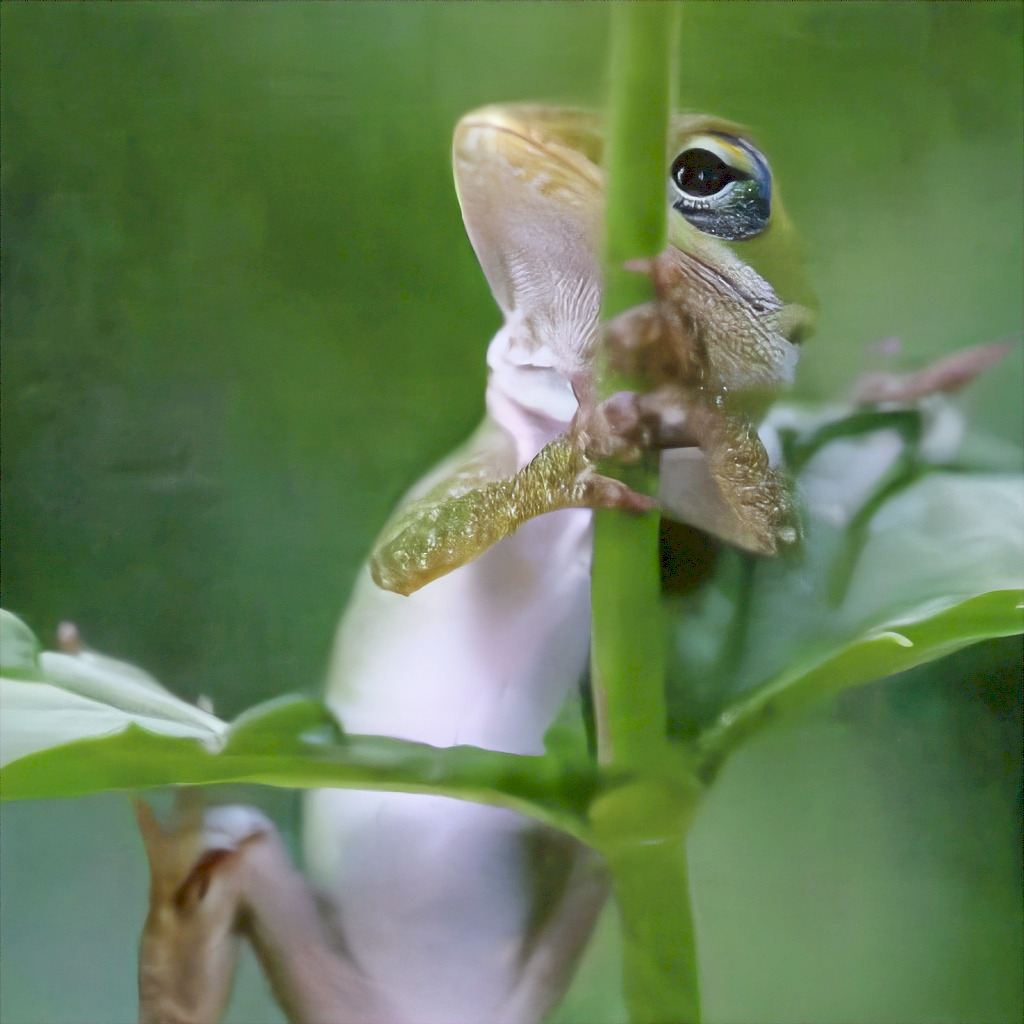} &
        \includegraphics[width=\ww,trim={10cm 10cm 1cm 1cm},clip]{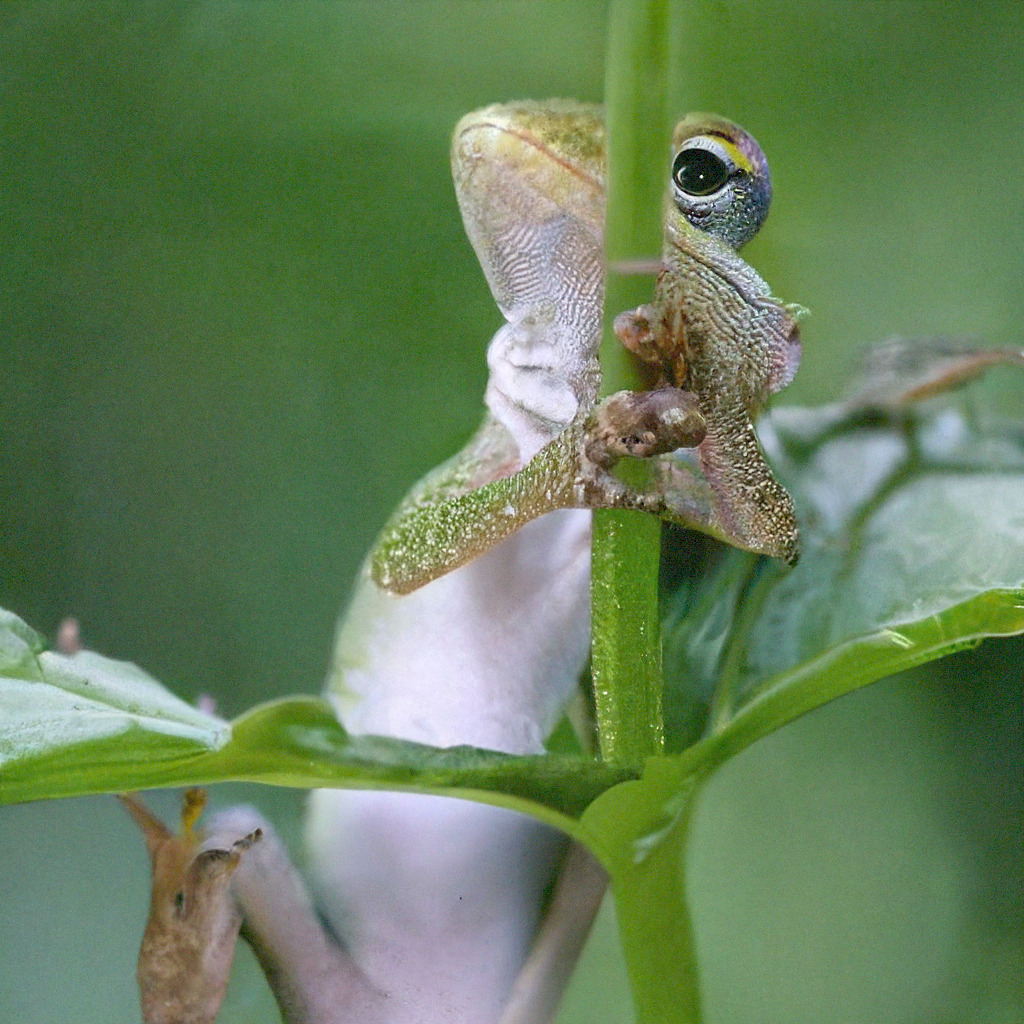} &
        \includegraphics[width=\ww,trim={10cm 10cm 1cm 1cm},clip]{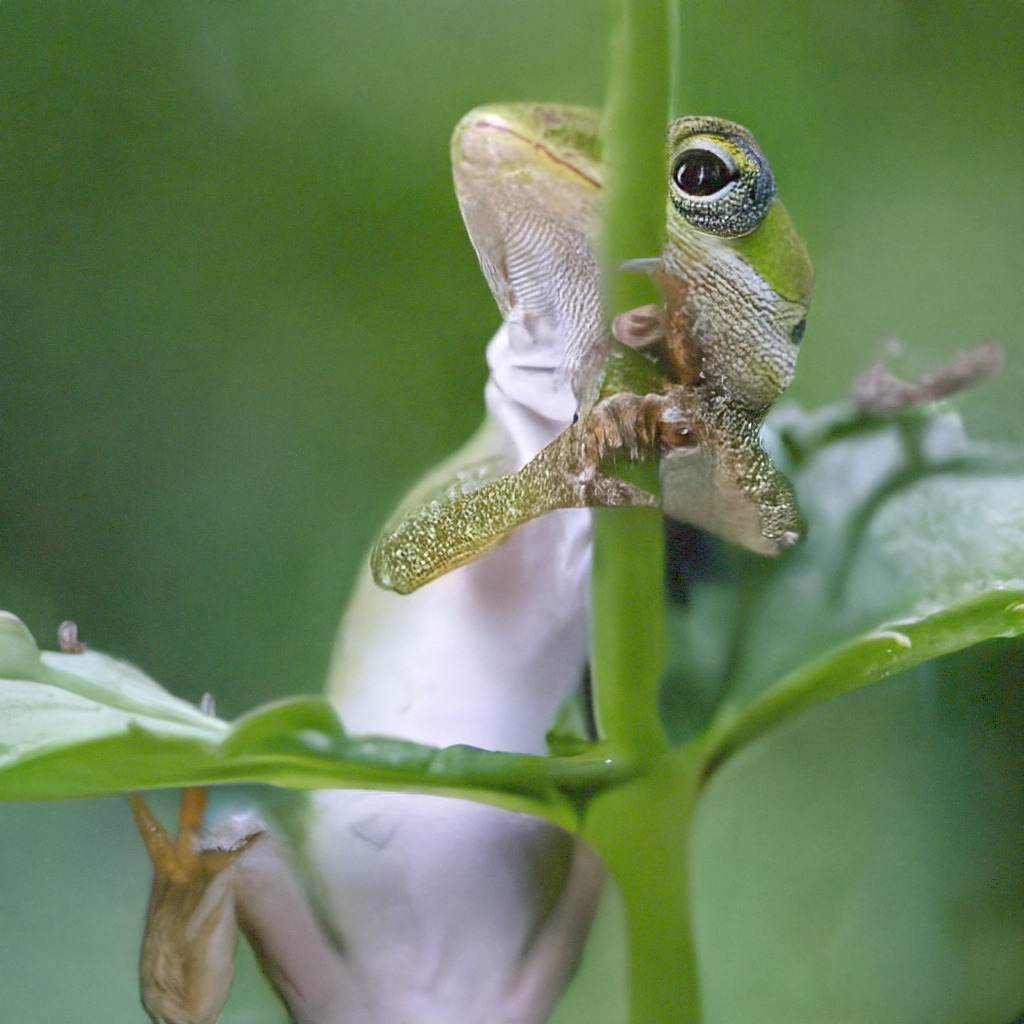} 
        \\
        \scriptsize{SoTA denoising~\cite{chen2021hinet}} &
        \scriptsize{Baseline result (1000 steps)} &
        \scriptsize{Ours (25 steps)}
        \\
    \end{tabular}
    \caption{\textbf{Top:} \emph{spatially-variant} standard deviation of noise (quantized), the resulting noisy image, and the ground truth clean image. 
    Our \svdd formulation handles such noise by applying a pixel-wise time embedding.
    \textbf{Bottom:} 
    state-of-the-art denoising methods manage to remove high levels of noise but over-smooth fine details.
    Diffusion based models are able to recover textures in the image even when they are hard to distinguish in the noisy image. 
    \svdd yields clean images of higher fidelity (part of the lizard's head is missing in the baseline result), while reducing the runtime $\sim \! \times 10$.}
    \label{fig:teaser}
    % \vspace{-1em}
\end{figure}
Image denoising, the task of removing unwanted noise from an image, while preserving its original features, is one of the most longstanding problems in image processing.
Over the years, numerous image denoising techniques have been developed, ranging from traditional filtering-based methods to more recent deep learning-based approaches, \eg, \cite{osher2005iterative,dabov2007image,zhang2017beyond,chen2021hinet,elad2023image}.

In modern real-world digital photographs, noise most commonly arises from the imaging sensor, and is particularly evident when images are captured in low-light conditions.
Yet, many of the proposed approaches make unrealistic assumptions regarding the noise
and/or assess the denoising performance using metrics such as PSNR or SSIM.
Such metrics struggle with the distortion-perception trade-off~\cite{blau2018perception} as they are sensitive to pixel alignment and do not emphasize the restoration of fine details or high-frequency textures, which may be difficult to distinguish from noise. 

In this paper, we propose a new denoising approach that leverages the natural image prior learned by today's powerful diffusion-based generative models \cite{ho2020denoising,dhariwal2021diffusion}.
Such models have been successfully applied to a variety of image restoration tasks \cite{saharia2022imagesr,saharia2022palette,kawar2022denoising,kawar2021stochastic}.
Furthermore, they possess innate denoising capabilities, since the entire generation process is based on gradual denoising of images. Thus, one might expect that it should be possible to reconstruct a clean image simply by starting the diffusion process from the noisy input image.
However, the diffusion process is based on additive white Gaussian noise (AWGN), while realistic noise models involve a signal-dependent component, the so-called shot-noise, which leads to higher noise levels in brighter parts of the image \cite{mildenhall2018burst}. 
This violates the denoising diffusion formulation that associates a single scalar noise level (time) with each step, making it non-trivial to apply the diffusion process to realistic noise removal.

In this work, we present \svdd, a novel denoising diffusion formulation that handles \textit{spatially-varying noise}, thereby enabling the reverse process to start from realistic noisy images, while significantly reducing the number of necessary diffusion steps.

Specifically, \svdd adapts the denoising diffusion framework to utilize the noisy input image as both the condition and the starting point.
We assume a realistic signal-dependent noise model (\secref{sec:noise_model}), with a  spatially-variant noise distribution.
To cope with such a noise distribution, we adapt the diffusion process to allow each pixel to have its own time embedding, effectively assuming that the denoising time step is spatially-varying, rather than constant, across the image. 
We further present training and inference schemes that support such spatially-varying time maps. Our training scheme also accounts for correlation between the condition image and the samples of the diffusion process, which stems from the fact that the reverse process starts with the same image it is conditioned on.

The spatially-variant time embedding, together with the associated training scheme, enables using the noisy input image as both the condition and the starting point for the denoising process, yielding higher quality clean images (\figref{fig:teaser}), while allowing significantly fewer denoising steps (\figref{fig:init_with_y}). 
We demonstrate the power of the \svdd framework on simulated noisy images exhibiting a wide variety of noise levels and show its ability to generate fine details, such as fur and intricate textures. 
We show that our framework outperforms the standard conditioned diffusion baseline quantitatively, as well as visually, while avoiding the over-smoothing of a state-of-the-art single-image denoising method~\cite{chen2021hinet} .
\section{Background and Related Work}
% 
% -------------------------------
\subsection{Image noise models}
Cameras sensors convert incident photons to voltage readings, which are then converted to bits by an analog to digital converter (ADC). Throughout this process, noise is unavoidably added to the measurement, depending both on photon statistics and the sensor's circuits. Sensor noise is often modeled as a combination of two primary components~\cite{nakamura2017image}: shot noise, which originates from photon arrival statistics and is modeled as a Poisson process depending on signal intensity, and read noise, which is caused by imperfections in the readout circuitry and is modeled as a Gaussian noise with standard deviation $\sigma_r$.
\subsection{Single image denoising}
Early works for single image denoising used prior knowledge like non-local self-similarity in BM3D~\cite{dabov2007image} or total variation~\cite{osher2005iterative}.

Recently, convolutional neural networks (CNNs) have shown their success in single image denoising, as summarized in this comprehensive survey \cite{elad2023image}. 
The following methods require a clean target image to train the CNNs.  Initially, they were trained on synthetically added i.i.d. Gaussian noise, however that practice fails to generalize to real noisy images~\cite{plotz2017benchmarking}. Later, datasets of real noisy images with their clean counterparts were collected (SIDD \cite{abdelhamed2018high}, RENOIR \cite{anaya2018renoir}), and are commonly used for denoising evaluation.
As shown in \cite{tran2020gan}, learning the noise distribution of real images via a GAN, which is used to synthesize noise for a denoising network, significantly improves performance. 
DnCNN~\cite{zhang2017beyond} predicts the residual image (the noise) of a noisy image.
Many works improved the performance by choosing better architectural components:  
SADNet~\cite{chang2020spatial} proposes a deformable convolution
to adjust for different textures and noise patterns, 
HINet~\cite{chen2021hinet} introduces instance normalization block for image restoration tasks and NAFNet~\cite{chen2022simple} suggests to replace non linear activation functions by element-wise multiplication between two sets of channels.
Some methods iteratively solve the problem in a multi-scale architecture or in multiple iterations: 
MPRNet~\cite{zamir2021multi} proposes supervised attention block between the different stages to leverage the restored image features at different scales. 
Somewhat similarly to our work, FFDNet~\cite{zhang2018ffdnet} employs a spatially-varying noise-map, and is able to remove non-uniform noise. However the architecture of FFDNet relies on downsampling and channel re-shuffle before applying a CNN to the image, which is different than the proposed approach.

Unlike the above works, which require clean target images, another line of works focuses on unsupervised or self-supervised solutions. According to N2N~\cite{lehtinen2018noise2noise}, the expected value of minimizing the objective with respect to clean samples is similar to minimizing it with respect to different noisy samples, and therefore clean images are not necessary.
Further works designed different ways for data augmentation that achieve the same purpose.
N2S~\cite{batson2019noise2self}, Noisier2noise~\cite{moran2020noisier2noise}, R2R~\cite{pang2021recorrupted}, neighbor2neighbor~\cite{huang2021neighbor2neighbor} use different subsamples of the image as instances of the noisy image.
IDR~\cite{zhang2022idr} added noise to the noisy image to create a noisier version which can be supervised by the noisy image.  

\subsubsection{Raw single image denoising / low light methods}
Some methods take into account the image formation model and aim to denoise the raw image, where the pixel values directly relate to the number of incident photons and the noise can be better modeled. To tackle the task of low-light imaging directly, SID~\cite{chen2018learning} introduces a dataset of raw short-exposure low-light images paired with corresponding long-exposure reference images. They train an end-to-end CNN to perform the majority of the steps of the image processing pipeline: color transformations, demosaicing, noise reduction, and image enhancement.
Brooks \etal~\cite{brooks2019unprocessing} present a technique to “unprocess” the image processing pipeline in order to synthesize realistic raw sensor images, which can be further used for training.
Wei \etal~\cite{wei2020physics} accurately formulate the noise formation model based on the characteristics of CMOS sensors.
Punnappurath \etal~\cite{punnappurath2022day} suggest a method that generates nighttime images from day images.
Similarly, in the field of low light video, Monakhova \etal~\cite{monakhova2022dancing} learn to generate nighttime frames of video.
\subsection{Diffusion models}
The usage of diffusion models for generative tasks grew rapidly over the past years,
and have shown great success in text-to-image generation (Imagen ~\cite{saharia2022photorealistic}, DALL$\cdot$E 2 \cite{ramesh2022dalle2}).
Denoising is a key component of the diffusion process, offering a strong image prior for both restoration and generative tasks.
% restoration tasks
SR3~\cite{saharia2022imagesr} adapts denoising diffusion probabilistic models to solve the super resolution task, conditioned on the low resolution image. 
Palette~\cite{saharia2022palette} extended this idea to a general framework for image-to-image translation tasks, including colorization, inpainting, uncropping, and JPEG restoration. In our evaluation, we compare to this method as a baseline, where the noisy image is given as a prior, but without modifying the diffusion formulation. 
Kawar \etal~\cite{kawar2021stochastic,kawar2022denoising} solve linear inverse image restoration problems by sampling from the posterior distribution, based on a pre-trained denoising diffusion model. This approach is limited to linear problems, whereas a realistic noise model is signal-dependant and not additive Gaussian.
In a concurrent work, Xie \etal \cite{xie2023diffusion} redefine the diffusion process to implement generative image denoising, however it is defined for different types of noise (Gaussian, Poisson) separately, while a realistic noise model is a combination of both.

\section{Method}
\label{sec:method}

Our main goal in this work is to leverage the powerful denoising-based diffusion framework for noise removal.
To this end, we adapt the framework to enable the noisy input image to be considered as a time step in the diffusion process.
Accounting for the more complex nature of real camera noise, we propose a diffusion formulation that unifies realistic image noise with that of the diffusion process.
In \secref{sec:noise_model}, we describe the camera noise model that we use, and in Sections~\ref{sec:diffusion_noise_model}--\ref{sec:time_embedding} we propose a diffusion process that can incorporate such noisy images as its samples.

For a more realistic modeling of noisy images, we consider a raw-sensor noise model, which is not uniform across the image.
This means that we cannot pair a step in the diffusion process with a single point in time. Instead, we pair each diffusion step with a spatially varying \emph{time map}, where each pixel may have a different time encoding (\secref{sec:time_embedding}).
The training and the inference schemes are modified to support such time maps, as described in \secref{sec:train-and-infer}. 

\begin{figure}[t]
{\centering
\includegraphics[width=\linewidth]{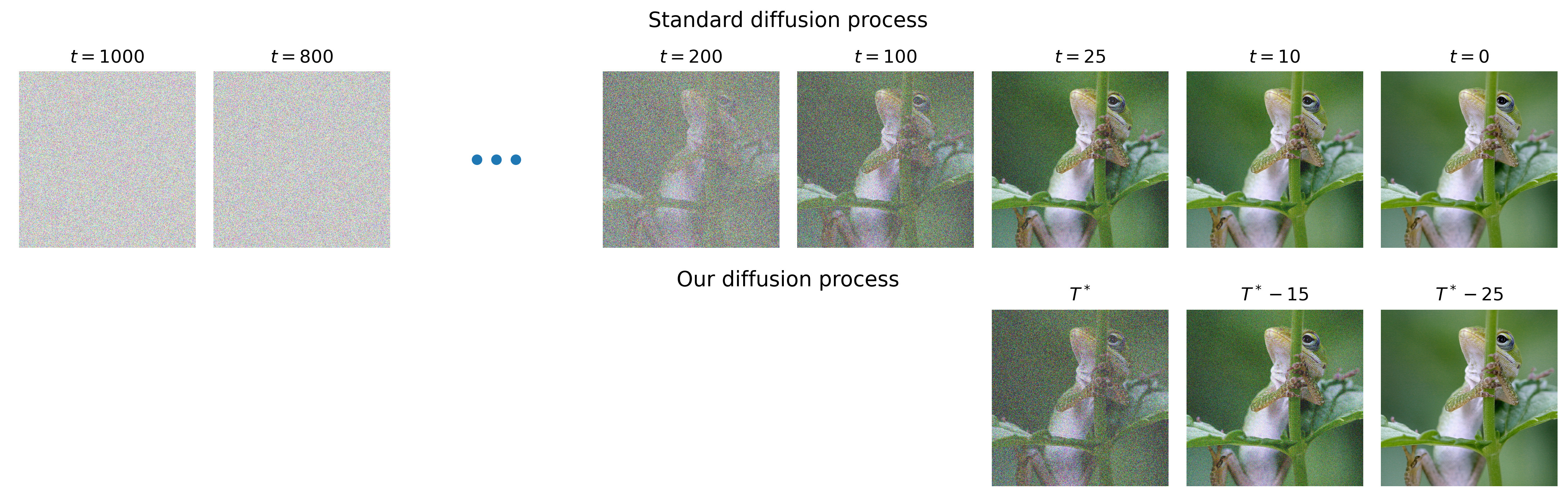}}
\caption{\textbf{Top:} standard forward diffusion process~\eqref{eq:org_diffusion_noise_model}. The reverse denoising process starts from complete noise (left) and iterates for $1000$ time-steps. \textbf{Bottom:} our diffusion formulation enables starting the reverse diffusion process from the noisy input image, requiring $\sim\!20$ iterations.}
\label{fig:init_with_y}
\vspace{-1em}
\end{figure}

In particular, the starting point of the diffusion process is set to the noisy input image, and not to an i.i.d Gaussian noise. This has the additional advantage of significantly reducing the number of diffusion steps ($\sim\!50$ times fewer steps in our experiments), see \figref{fig:init_with_y}.
However, using the same noisy input image as both the condition and the starting point of the diffusion process, introduces another challenge: there is a correlation between the condition and the samples along the reverse diffusion process at inference time, a correlation that is not reflected in the training scheme. We address this challenge in \secref{sec:noise_correlation}, give a theoretical analysis of this phenomenon and propose a modified training scheme to overcome it.

\paragraph{Notation and setting:}
Below we use small italics (\eg, $x$) to denote scalars, while bold roman letters (\eg, $\mathbf{x}$) denote vectors. 
Images and other per-pixel maps are represented as vectors in $\mathbb{R}^{H\times W \times 3}$.
In particular, $\n{} \sim \Nmat$ is a noise vector with the same dimensions, whose elements are sampled from $\N{0, 1}$.  
The operations $\mathbf{a} \cdot \mathbf{b} \mbox{ and } \frac{\mathbf{a}}{\mathbf{b}}$ between two vectors $\mathbf{a}\mbox{ and }\mathbf{b}$, denote element-wise multiplication and division respectively.
\subsection{Noise model}
\label{sec:noise_model}
We adopt a noise model that is commonly used for sensor raw data ~\cite{mildenhall2018burst,pearl2022nan}. 
The noisy version $\y \in \R^{H \times W \times 3}$ of a clean linear image  $\x \in \R^{H \times W \times 3}$ is given by:
\begin{equation}
\begin{aligned}
\label{eq:noise_model}
    &\y = \x +  \sigmap \cdot \n{\y}, \quad \n{\y} \sim \Nmat, \\
    & \sigmap \triangleq \sqrt{\sigma_r^2 + \sigma_s^2 \x},
\end{aligned}
\end{equation}
where $\n{\y} \in \R^{H \times W \times 3}$ and $\sigmap$ is the per-pixel standard deviation of the noise, defined as a combination of
$\sigma_r$, the standard deviation for the \emph{signal-independent} read-noise, and $\sigma_s$ for the \emph{signal-dependent} shot-noise.
See \secref{sec:imagenet_simulation} for further details regarding our experiments.
\subsection{Diffusion process definition}
\label{sec:diffusion_noise_model}
Given a clean image $\x$ and a noise schedule $\left\{ \beta_t \right\}_{t=1}^T$, the standard diffusion process of length $T$ is given by:
\begin{equation}
\begin{aligned}
\label{eq:org_diffusion_noise_model}
    & q \left( \xt{t} \vert \xt{t-1} \right) = \N{\xt{t}; \sqrt{1 - \beta_t} \xt{t-1}, \beta_t \mathbf{I}}, \\
    & \bar{\alpha}_t = \prod_{i=1}^t \alpha_i = \prod_{i=1}^t (1 - \beta_i), \\ 
    & q \left(\xt{t} \vert \x \right) = \N{\xt{t}; \sqrt{\bar{\alpha}_t} \x, (1 - \bar{\alpha}_t)\mathbf{I}}.
\end{aligned}
\end{equation}
Note that this formulation defines a Markovian process, i.e., the variance of $\xt{t}$ along the process is constant (assuming $\mathbb{E}(\x) = 0$ and $\var{\x}=1$).
As the noise level increases, the stationary nature of $\xt{t}$ is achieved by attenuating the clean signal by a factor of $\sqrt{\bar{\alpha}_t}$.
To be able to refer to $\y$ as a sample from the diffusion process, we need to overcome two obstacles. The first issue is that in our noise model, the signal is not attenuated, and the second is that our noise model uses a spatially-varying noise distribution. We first resolve the former issue and modify the diffusion process to be non-stationary, by considering a process which does not attenuate the signal:
\begin{equation}
\begin{aligned}
\label{eq:ve_diffusion}
& q \left(\xt{t} \lvert \xt{t-1} \right) = \N{\xt{t} ; \xt{t-1}, \eta_t \mathbf{I}}, \\
& q \left(\xt{t} \lvert \x \right) = \N{\xt{t} ; \x, \gamma_t \mathbf{I}}, \\
& \gamma_t = \sum_{i=1}^{t} \eta_i,
\end{aligned}
\end{equation}
for some noise schedule $\left\{ \eta_t \right\}_{t=1}^T$.
This process, where $\var{\xt{t}|\x}\to\infty$ as $t\to\infty$, is termed ``Variance Exploding" by Song \etal~\cite{song2021score}.

We wish to keep the noise schedule similar to the original DDPM schedule~\cite{ho2020denoising}. Hence we choose the noise schedule $\eta_t$ so that $\gamma_t$ will be a scaled version of $1-\bar{\alpha}_t$, that is, $\gamma_t = \lambda \left( 1-\bar{\alpha}_t \right)$ for some $\lambda$. This implies,
\begin{equation}
\label{eq:noise_schedule}
 \eta_t = \lambda \beta_t \Pi_{i=1}^{t-1} (1 - \beta_i).
\end{equation}
This non-stationary forward process, yields a reverse process of the same form as in the standard diffusion,
\begin{equation}
\begin{aligned}
\label{eq:ve_reverse}
& q \left(\xt{t-1} \lvert \xt{t}, \x \right) =  \N{\xt{t-1} ; \mathbf{\tilde{\boldsymbol{\mu}}_t} \left( \xt{t}, \x \right), \tilde{\eta}_t \mathbf{I}},\\
& \mathbf{\tilde{\boldsymbol{\mu}}_t} \left( \xt{t}, \x \right) = \frac{\gamma_{t-1}}{\gamma_{t}} \xt{t} + \frac{\eta_{t}}{\gamma_{t}} \x, \\
& \tilde{\eta}_t = \frac{\gamma_{t-1} \eta_{t}}{\gamma_{t}}. \\
\end{aligned}
\end{equation}
The fact that our noise model does not attenuate the clean signal $\x$ is reflected in the expression for $\mathbf{\tilde{\boldsymbol{\mu}}_t}$, that lacks the multiplication by the attenuation factor $\alpha, \bar{\alpha}$.
More details can be found in the supplementary materials.

At inference time, the diffusion process should start with $\xt{T} = \x + \sqrt{\lambda} \n{T}, ~\n{T} \sim \Nmat$. Note that in our noise model one cannot start the reverse process from pure noise (as done in standard diffusion processes), since the signal is not attenuated to $0$. However, since our goal is to start the reverse process from the input noisy image, this is not a concern. 

\subsection{Spatially-variant time embedding}
\label{sec:time_embedding}
Our noise schedule, Eq.~\eqref{eq:ve_diffusion}, defines a noise level $\gamma_t$ for every integer $t$ between $0$ and $T=1000$. As in standard diffusion models, we can extend the definition of $\gamma_t$ to non-integer $t$ using interpolation. 
Thus, given a noise level $\sigma^2$, we can find a time $t$ at which this noise level is attained.
Consider now our camera noise model, Eq.~\eqref{eq:noise_model}. 
Each pixel $p$ has a different noise level $\sigmapsquare(p)$, and thus a corresponding time value that yields this noise level. 
The maximum noise level over the three channels defines a time map $\tmap \in \mathbb{R}^{H\times W}$ for which $\gammamat{\tmap(p)}= \max_{c \in \text{R,G,B}} \sigmapsquare(p_c)$.
In other words, we think of each pixel as being at its own stage of the diffusion process. 
Note that the time map $\tmap$ encodes the spatially-varying noise of the entire input image $\y$. Hence we denote
\begin{equation}
\begin{aligned}
    & \xt{\tmap} \triangleq \y, \quad \n{\tmap} \triangleq \n{\y}, \quad \mathbf{\gammamat{\tmap}} \triangleq \max_{\text{R,G,B}}{ \sigmapsquare}.
\end{aligned}
\end{equation}

In practice, when presented with a noisy image $\y$, we do not know the actual noise level $\sigmap$, even if $\sigma_r$ and $\sigma_s$ are known, since the original clean signal $\x$ is not available. 
Thus, we follow common practice~\cite{mildenhall2018burst} and estimate it using a clipped version of the noisy image, to obtain $\esttmap$ such that
\begin{equation}
\begin{aligned}
\label{eq:estimated_tmap}
    & \gammamat{\esttmap}= \max_{\text{R,G,B}} \estsigmapsquare \\
    & \estsigmapsquare = \sqrt{\sigma_r^2 + \sigma_s^2 \; \cdot \; \clip{\y, 0, 1}}.
\end{aligned}
\end{equation}

A standard diffusion model receives as input both $\xt{t}$ and a time value $t$, indicating the signal noise level over the entire image.
An embedding vector of the time is then used to apply an affine transformation independently to each pixel feature in $\xt{t}$.
By replacing $t$ with a spatially-varying time map $\tmap$, and computing a different time embedding per pixel, we can make the model dependent on the spatially-varying noise level $\sigmap$. However, since each pixel can now be at a different stage of the diffusion process, it requires a different number of steps to reach time $0$. Hence, we need to develop new training and inference schemes to account for this, which are presented below.

\subsection{Training and inference schemes}
\label{sec:train-and-infer}
Our diffusion model receives as input a noisy image $\y$ and a time map $\tmap$. We present training and inference schemes that account for this change. Our algorithm is summarized in Algs.~\ref{alg:training_scheme} and~\ref{alg:inference_scheme}.

Note that the reverse diffusion process, Eq.~\eqref{eq:ve_reverse}, operates on each pixel independently. Thus, we can use the same reverse process even with a spatially-varying time step $\tmap$. However, each pixel may require a different number of steps before reaching time $0$. We handle this by stopping the reverse process once a pixel reaches a negative time. In other words, the time map after $t_0$ denoising steps will be $\clipzero{\tmap-t_0} \triangleq \max\{\tmap-t_0, 0\}$.

During training, given a clean image $\x$, we sample $\sigma_r$, $\sigma_s$, and a random noise $\n{\y}=\n{\T}$. 
The noisy image $\y$ is then generated according to the noise model Eq.~\eqref{eq:noise_model}, and the estimated induced time map $\esttmap$ is calculated by Eq.~\eqref{eq:estimated_tmap}.
Next, we sample a scalar $t_0$ between $0$ and the maximal value of $\esttmap$, and advance the times of all the pixels by $t_0$ steps, to obtain $\promt = \clipzero{\esttmap - t_0}$. We then sample a random Gaussian noise $\n{\promt}$ and construct a sample $\xt{\promt}=\x+\gammamat{\promt}\n{\promt}$ of the diffusion process according to Eq.~\eqref{eq:ve_diffusion}. Note that $\gammamat{\promt}$ is a matrix, so the noise level is spatially-varying. 
The network then tries to predict $\n{\promt}$ from the diffusion sample $\xt{\promt}$, the time map $\promt$, and the condition image $\y$.

At inference time, we get a noisy image $\y$ and its $\sigma_r, \sigma_s$. First, we estimate the time map $\esttmap$ by Eq.~\eqref{eq:estimated_tmap}. We feed the network with $\y$ as the condition image, $\esttmap$ as the time map, and $\y=\xt{\tmap}$ as the diffusion sample. The network outputs an estimate of the noise $\n{\esttmap}$, from which we can compute an estimate of the original image $\estx$. We then use the reverse process Eq.~\eqref{eq:ve_reverse} (replacing $\x$ by $\estx$) to produce the next sample. 
Additionally, we promote the time map $\esttmap$ by one step, \ie, we replace $\esttmap$ with $\hat{\mathbf{t}} = \clipzero{\esttmap - 1}$. We then run the network with our new sample and the promoted $\hat{\mathbf{t}}$ (using the same condition $\y$), and continue in this manner until we reach $\hat{\mathbf{t}} = 0$ for all pixels.

Explicitly, the reverse process is preformed by sampling a Gaussian noise $\n{\promt - 1}\sim \Nmat$ and computing
\begin{equation}
\label{eq:explicit_ve_reverse}   
  \xt{\promt-1} = \frac{\gammamat{\promt - 1}}{\gammamat{\promt}} \xt{\promt} + \frac{\etamat{\promt}}{\gammamat{\promt}} \estx + \sqrt{\frac{\gammamat{\promt - 1} \etamat{\promt}}{\gammamat{\promt}}} \n{\promt - 1},
\end{equation}
where in $\promt - 1$ we clip the negative values, and $\gammamat{\promt}, \gammamat{\promt - 1}, \etamat{\promt}$ are all vectors of the same dimension as $\x$, whose values depend on the initial noise in the image.
To avoid further denoising of pixels whose time has reached 0, we override their values after the prediction by the network.

\begin{algorithm}[htbp]
	\For{$ i=1, \ldots$}{
		Sample $\x,\sigma_r, \sigma_s$
		
		Sample $\y$ by Eq.~\eqref{eq:noise_model}
		
		Calculate $\esttmap$ by Eq.~\eqref{eq:estimated_tmap}
		
		Sample $t_0 \sim \mathcal{U}\left[ 0, \max{(\esttmap)} \right]$
		
		Set $\promt = \max \{\esttmap - t_0, 0\}$
		
		Calculate $\xt{\promt}$ by Eq.~\eqref{eq:train_ddpm}
		
		$\estx = \svddalg{\y, \xt{\promt}, \promt}$
		
		Calculate loss and update weights.
	}
	\caption{Training diffusion initialized with $\y$}
	\label{alg:training_scheme}
\end{algorithm}
\begin{algorithm}[htbp]
	\SetKwInOut{KwIn}{Inputs}
	% \SetKwInOut{KwOut}{Outputs}
	% \KwOut{$\B{A}_{\tau}, X_{\tau}$
		
		\KwIn{$\y, \sigma_r, \sigma_s$}
		
		Calculate $\esttmap$  by Eq.~\eqref{eq:estimated_tmap}
		
		Set $\promt = \esttmap$, $\xt{\promt} = \y$
		
		\While{$\mathrm{any}(\promt > 0)$}{
			
			$\estx = \svddalg{\y, \xt{\promt}, \promt}$
			
			Sample $\xt{\clipzero{\promt - 1}}$ by Eq.~\eqref{eq:explicit_ve_reverse}
			
			Override pixels that will reach $\clipzero{t - 1}=0$ with the values in $\estx$. These values remain fixed for the rest of the process.
			
			Set $\promt = \clipzero{\promt - 1}, \xt{\promt} = \xt{\clipzero{\promt - 1}}$
			
		}
		\caption{Inference by diffusion from $\y$}
		\label{alg:inference_scheme}
\end{algorithm}

\subsection{Noise correlation in the reverse process}
Next, we discuss a phenomenon that arises when we initialize the process with the noisy input image \emph{and} condition the process on it.
The key observation is that throughout the reverse diffusion process, there is a correlation between the noise component of the diffusion sample $\xt{\mathbf{t}}$ and the noise component of the condition image $\y=\xt{\tmap}$. 

\label{sec:noise_correlation}
When initializing the diffusion process with $\xt{\tmap}$, the first reverse step yields a sample $\xt{\tmap-1}$ derived from Eq.~\eqref{eq:ve_reverse}.
This sample is less noisy than $\xt{\tmap}$ and can be explicitly written (given $\x$) as 
\begin{equation}
\xt{\tmap\!-\!1} \!=\!  \frac{\gammamat{\tmap\!-\!1}}{\gammamat{\tmap}} \xt{\tmap} + \frac{\etamat{\tmap}}{\gammamat{\tmap}} \x + \sqrt{\frac{\gammamat{\tmap\!-\!1} \etamat{\tmap}}{\gammamat{\tmap}}} \n{\tmap\!-\!1}.
\end{equation}
Using Eq.~\eqref{eq:noise_model} it can be rewritten as a summation of $\x$ and an additional noise term, which is a linear combination between the noise $\n{\tmap}$ and the new sampled noise term $\n{\tmap-1}$,
\begin{equation}
\xt{\tmap\!-\!1} = \x + \frac{\gammamat{\tmap\!-\!1}}{\sqrt{\gammamat{\tmap}}} \n{\tmap} + \sqrt{\gammamat{\tmap\!-\!1} \!\! \left(\! 1 \!-\! \frac{\gammamat{\tmap\!-\!1}}{\gammamat{\tmap}} \! \right)} \n{\tmap\!-\!1}.
\end{equation}
After $t_0$ inference steps, the time map is $\mathbf{t}=\clipzero{\tmap - t_0}$ and $\xt{\mathbf{t}}$ can be written as
\begin{equation}
\begin{aligned}
\label{eq:train_ddpm}
\xt{\mathbf{t}} &= \x + \frac{\gammamat{\mathbf{t}}}{\sqrt{\gammamat{\tmap}}} \n{\tmap} + \sqrt{\gammamat{\mathbf{t}} \left( 1 - \frac{\gammamat{\mathbf{t}}}{\gammamat{\tmap}} \right)} \n{\mathbf{t}}, \\
&= \x + \sqrt{\gammamat{\mathbf{t}}} \tilde{\n{}}_{\mathbf{t}}.
\end{aligned}
\end{equation}
The full derivation can be found in the supplementary materials.
The modified noise $\tilde{\n{}}_{\mathbf{t}}$ is a linear combination between the initial noise of $\n{\tmap}$ and another i.i.d noise term, $\n{\mathbf{t}}$,
\begin{equation}
\label{eq:noise_correlation}
\tilde{\n{}}_{\mathbf{t}} = \sqrt{\frac{\gammamat{\mathbf{t}}}{\gammamat{\tmap}}} \n{\tmap} + \sqrt{1 - \frac{\gammamat{\mathbf{t}}}{\gammamat{\tmap}}} \n{\mathbf{t}}.
\end{equation}
This relationship describes the correlation between $\tilde{\n{}}_{\mathbf{t}}$, the noise component of the diffusion sample $\xt{\mathbf{t}}$, and $\n{\tmap}$, the noise component of the condition image $\y=\xt{\tmap}$. 

Because of the above correlation, at train time the network sees a different distribution than at inference time.
During training, the noise of the diffusion sample $\xt{\mathbf{t}}$ consists entirely of noise sampled independently from $\n{\tmap}$. Hence, at train time, the $\xt{\mathbf{t}}$ and $\y$ presented to the network are two independent degradations of the true signal $\x$. This effect is made clearer when one considers the first step (\ie, $t_0=0$). While at train time the network sees two independent samples of $\x$ noised with $\sigmap$, at inference time the two images are the same.

\begin{figure}[t]
	\centering
	\includegraphics[width=1\linewidth]{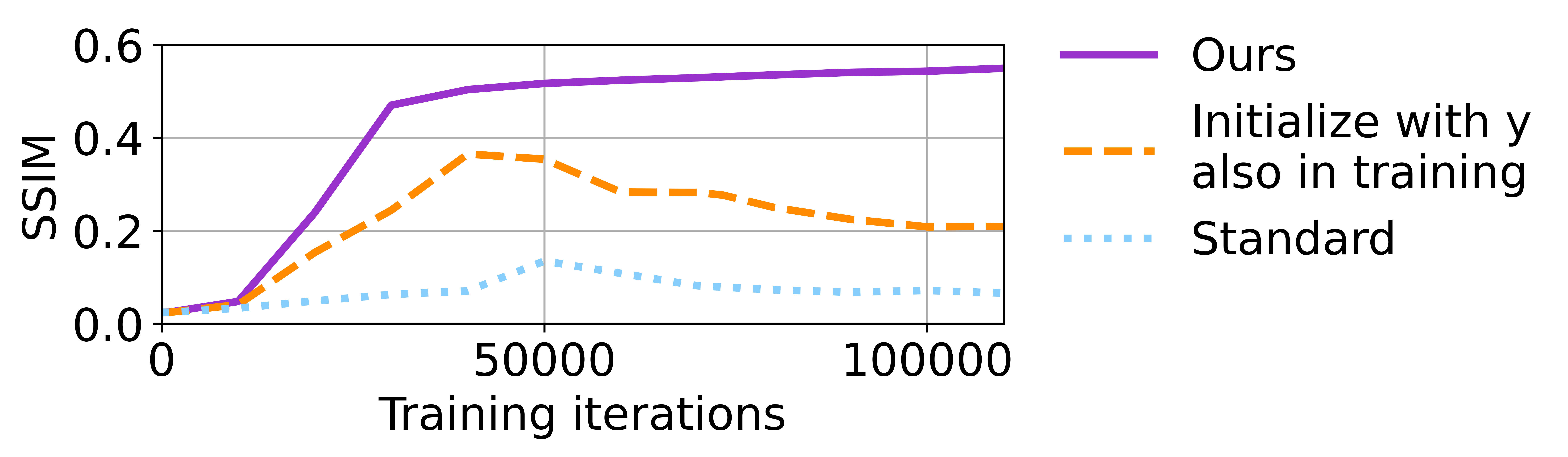}
	\caption{SSIM of validation during training. The standard training scheme (light blue) cannot restore the signal. Initializing the diffusion with the noisy image also in training (orange) partially solves the problem, but over time the network utilizes the two realizations of the noise (from the conditioned image and the diffusion sample) that are not available during inference. Our training scheme (purple) that relies on Eq.\eqref{eq:train_ddpm} yields stable training.}
	\label{fig:training_overfitting}
    \vspace{-1em}
\end{figure}

Indeed, looking at the progress of inference error in \figref{fig:training_overfitting}, we see a sudden drop of quality, which can be explained by the fact that the network may be learning to utilize its two uncorrelated inputs, which does not generalize to the inference process.

A naive solution to this problem would be to drop the conditioning entirely, however, our ablation study shows that this yields deteriorated results. The experiments suggest that it stems mainly from the clipping of negative values, which violates the noise model.

Thus, we choose to pursue a different approach and modify the training scheme to explicitly account for this correlation. Specifically, we propose to sample $\xt{\mathbf{t}}$ during training according to Eq.~\eqref{eq:train_ddpm}, in order to simulate a distribution of inputs that is similar to that of inference time. 
As noted above, a special case of this noise correlation is when $t_0=0$ and $\y=\xt{\tmap}$. We increase the probability of those cases to $1\%$ of the training iterations.

\section{Results}
We test our method on natural images from the ImageNet dataset~\cite{deng2009imagenet}, corrupted by simulated noise that was generated by our noise model (Eq.~\eqref{eq:noise_model}).
For training we use the full training set of ImageNet, and for evaluation we use a subset of 2000 images from the ImageNet validation set.

We compare our results to a strong diffusion baseline, based on the framework of~\cite{saharia2022imagesr, saharia2022palette}, that was trained to solve the task of image denoising (conditioned on the noisy image), in addition to a state-of-the-art single image denoising method~\cite{chen2021hinet}.
We report quantitative PSNR, SSIM, LPIPS~\cite{zhang2018unreasonable} and FID~\cite{heusel2017gans} metrics for all of the models and datasets.
While the former three metrics are used to compare pairs of images, the FID metric is used to compare entire distributions. We include this metric to asses the overall similarity between the distribution of the ground truth clean images and the distribution of the denoised results.
\subsection{Data and implementation details}
\label{sec:imagenet_simulation}
\paragraph{Noise simulation:}
The noise model in Eq.~\eqref{eq:noise_model} is defined with respect to linear images. Hence, we first ``linearize" the images by applying inverse gamma-correction and inverse white level. For white level values, during training we sample a value in the range $[0.1, 1]$, and use $0.5$ during validation. 

\begin{figure}[t]
	\centering
	\includegraphics[width=1.0\linewidth]{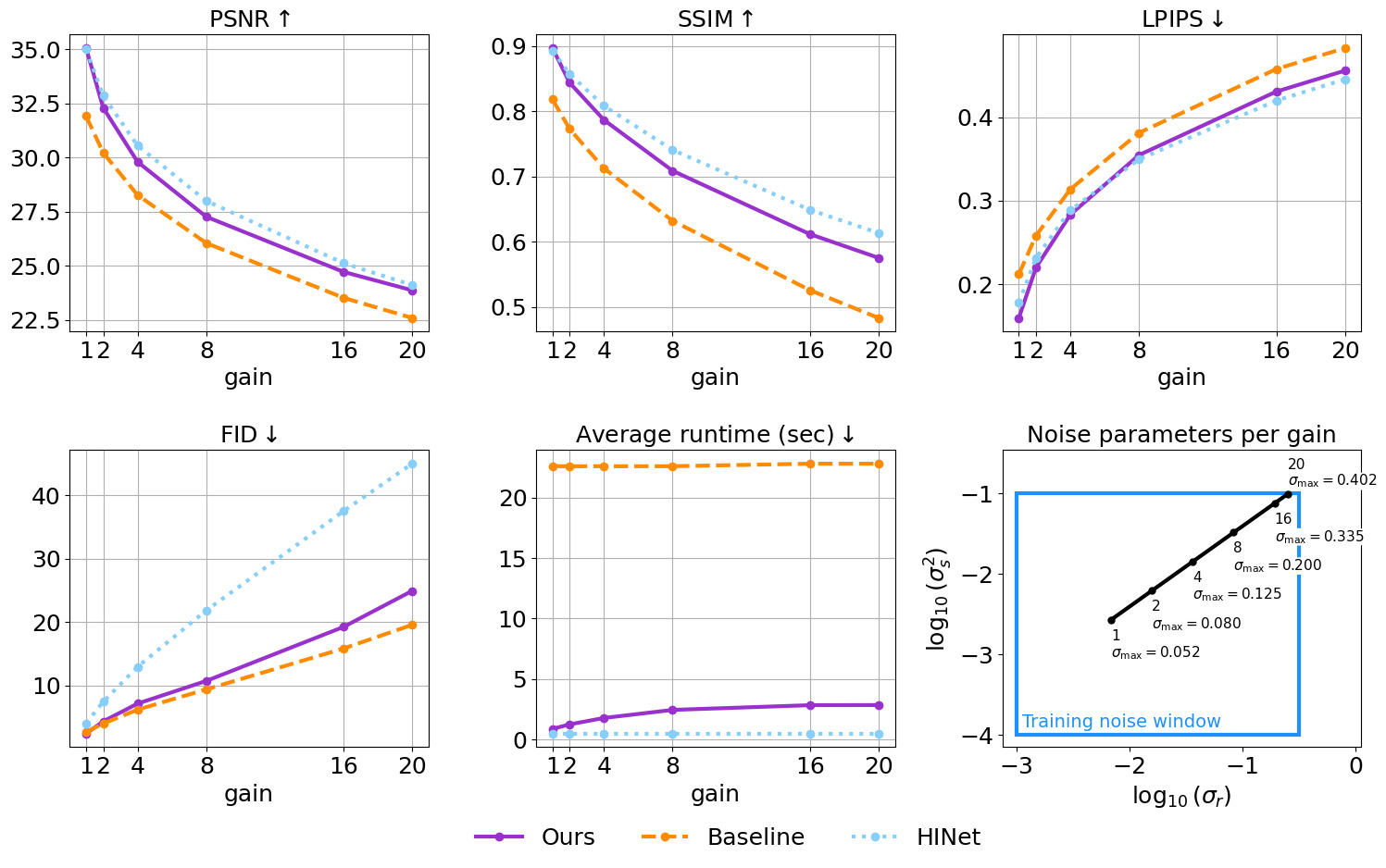}
	\caption{Quantitative results for simulated noise across different noise levels. 
    We compare the diffusion baseline, a single image denoising method~\cite{chen2021hinet} and our method. The metrics we report are PSNR, SSIM, LPIPS~\cite{zhang2018unreasonable} and FID~\cite{heusel2017gans}. In addition, average runtimes are presented for the diffusion methods.
    The noise is simulated using noise model in Eq.~\eqref{eq:noise_model}. During training, the noise parameters are sampled from the blue rectangle. At inference time, we use a set of fixed noise parameters that correspond to various gain levels of an example camera, as described in~\cite{mildenhall2018burst}.}
	\label{fig:imagenet_sim_metrics}
    % \vspace{-1em}
\end{figure}

We train the network on a range of values for $\sigma_r, \sigma_s$ and evaluate the method on fixed gain levels of an example camera, defined in~\cite{mildenhall2018burst}. 
Following~\cite{pearl2022nan}, we consider a wider training region and higher gain levels in our evaluation. See \figref{fig:imagenet_sim_metrics} for the specific values used during training and evaluation.

To make the noisy images more realistic, we further clip the images at $0$ after the addition of noise, as negative values are not attainable in real sensors. 
Our network seems to overcome this discrepancy between the theoretical model and the data distribution we use in practice.
We do not clip the image at higher values, as it can be adjusted with exposure time.
We use crops of $256 \times 256$ for training and a set of $2000$ images for validation, cropped to the maximum square and resized to $1024 \times 1024$.
The noise is added after the resizing, so we do not change the noise distribution.
\vspace{-0.5em}
\paragraph{Implementation details:}
Before being fed into the network, the input noisy images are scaled to occupy the full range of $[-1, 1]$ to match the diffusion models assumption. The noise standard deviation is scaled accordingly.
The input to the network has $6$ channels: $3$ RGB channels of the noisy image $\y$ (condition) and $3$ RGB channels of the sample in the diffusion process $\xt{\mathbf{t}}$. 
In addition, the network is also given as input the spatially-varying time map, which is computed from the known noise parameters $\sigma_r, \sigma_s$.
At inference time the sample of the diffusion process is initialized with the noise image $\y$ and the estimated $\esttmap$.

We fine-tune a fully-convolutional version of the Imagen model~\cite{saharia2022photorealistic}, disregarding the text components and conditioning it on the degraded input image, as done in~\cite{saharia2022palette, saharia2022imagesr}.
We use $\left\{ \beta_t \right\}_{t=1}^T$ that are linearly spaced in the range $[0.02, 10^{-8}]$ and $T=1000$ for the standard diffusion in Eq.~\eqref{eq:org_diffusion_noise_model}, and $\lambda = 20$ for the modified noise schedule in Eq.~\eqref{eq:noise_schedule}.
We train the network on 8 TPU-v4 chips, for $900K$ iterations and follow the training optimization of~\cite{saharia2022photorealistic}, with Adam optimizer and learning rate scheduler with linear warm-up followed by cosine decay. 
The training phase takes three days.

\subsection{Results on ImageNet}

We evaluate our method on a subset of $2000$ images from the ImageNet dataset \cite{deng2009imagenet} and report metrics for noise levels corresponding to gains ranging from 1 to 20. Note that while the input to the network are ``linearized" images, the metrics are calculated on the reprocessed images, \ie, after readjusting the white level and reapplying the gamma correction.
As mentioned before, we compare our results to a strong diffusion baseline, as well as to HINet, a state-of-the-art single image denoising method~\cite{chen2021hinet}.
For a fair comparison, we retrain HINet on the same dataset and noise levels that we used.
Quantitative results for PSNR, SSIM, LPIPS and FID metrics are reported in \figref{fig:imagenet_sim_metrics}, as well as the average runtime per example (in seconds).

Compared to the state-of-the-art model, our method (\svdd) shows slightly worse performance in all ``pixel-to-pixel" metrics, while achieving a signifcantly better FID score. On the other hand, the baseline diffusion model outperforms our model in the FID metric but exhibits signficantly worse results in all other metrics. 
This nicely demonstrates how our approach balances the perception-distortion trade-off~\cite{blau2018perception}. We can see that the baseline diffusion model favours realistic images at the expense of lower fidelity to the clean signal, while the state-of-the-art model shows the best fidelity to the signal at the cost of drifting away from the input distribution. In contrast, \svdd manages to keep a relatively high signal fidelity without the significant distribution drift. 

\begin{figure*}[htbp]
\setlength{\ww}{0.192\textwidth}
\begin{center}
\newcommand{\magn}{5.0}
\newcommand{\spyloc}{(0.28,0.15)}    
\newcommand{\spyshift}{(1.675,-3.4)}
\newcommand{\bright}{0.1 1 0.1 1 0.1 1}
\small\addtolength{\tabcolsep}{-8.5pt}
\begin{tabular}{ccccc}
    Noisy & HINet~\cite{chen2021hinet} & Baseline & Ours & Clean GT\\
        %% sky 1010
        \renewcommand{\spyloc}{(-0.6,-0.9)}
        \begin{tikzpicture}[spy using outlines={red,magnification=\magn,size=\ww}]
            \node {\includegraphics[width=\ww]{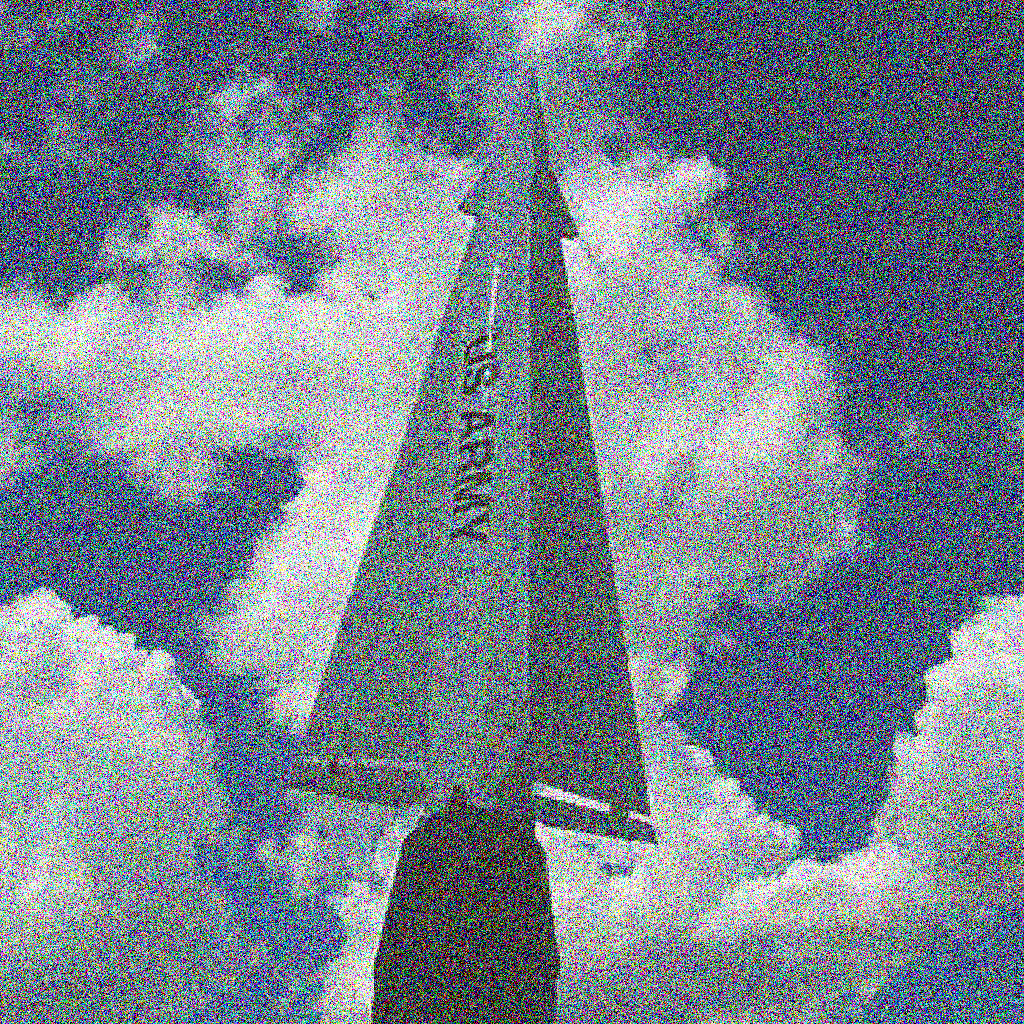}};
            \spy on \spyloc in node [left] at \spyshift;
        \end{tikzpicture} &
        \renewcommand{\spyloc}{(-0.6,-0.9)}
        \begin{tikzpicture}[spy using outlines={red,magnification=\magn,size=\ww}]
            \node {\includegraphics[width=\ww]{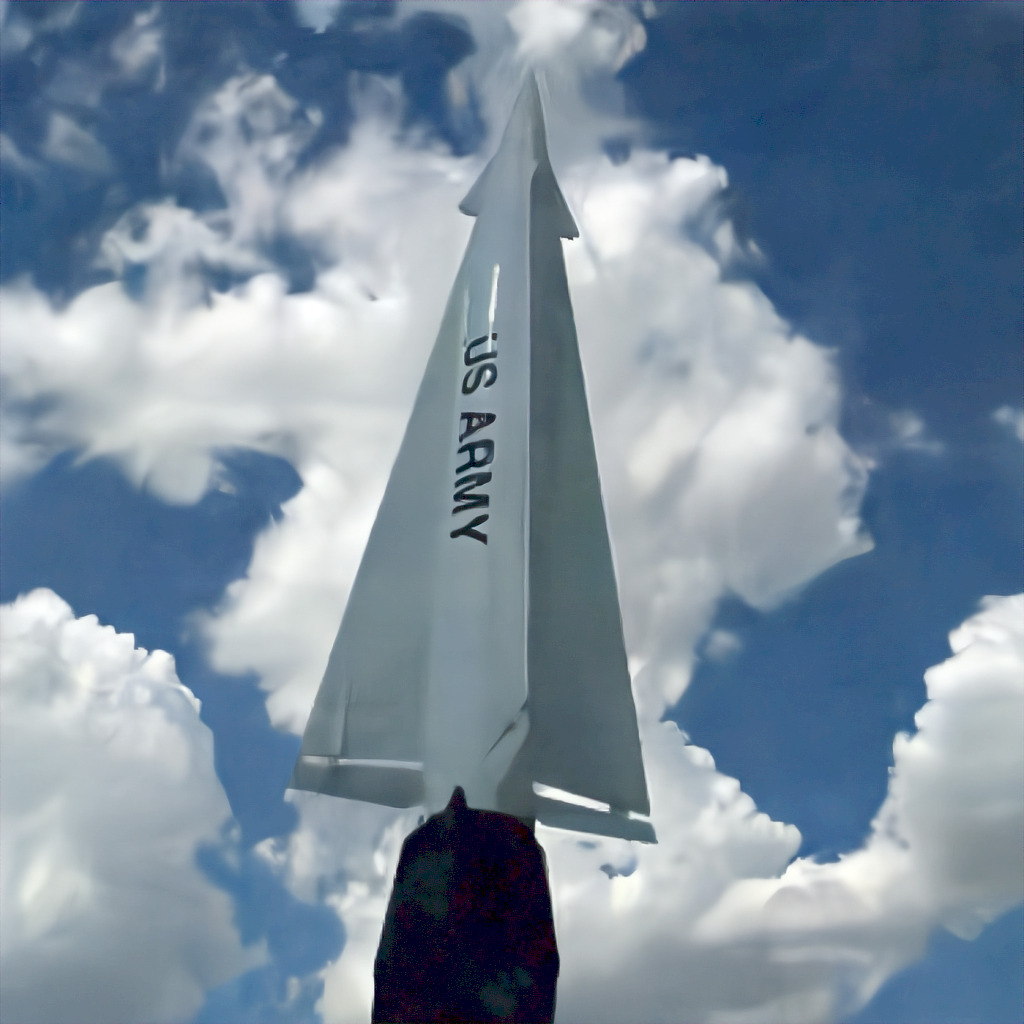}};
            \spy on \spyloc in node [left] at \spyshift;
        \end{tikzpicture} &
        \renewcommand{\spyloc}{(-0.6,-0.9)}
        \begin{tikzpicture}[spy using outlines={red,magnification=\magn,size=\ww}]
            \node {\includegraphics[width=\ww]{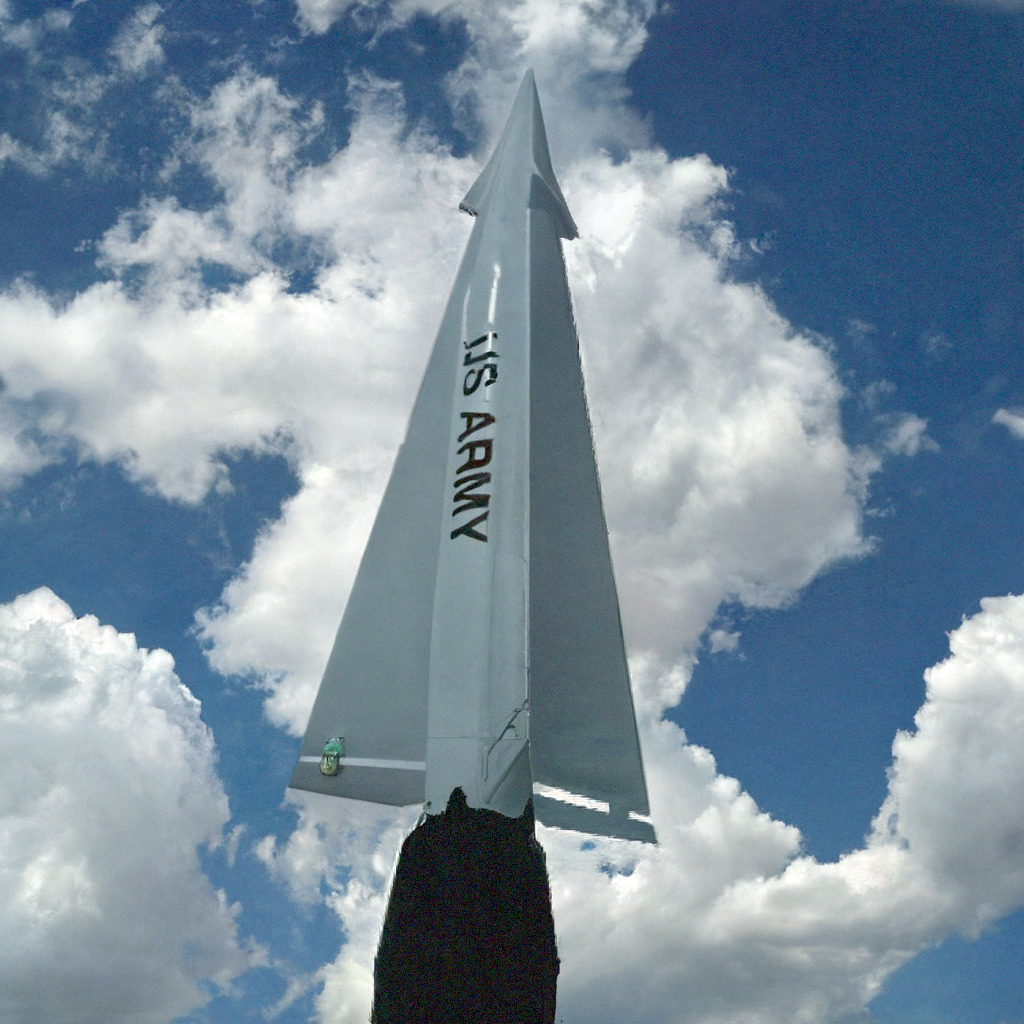}};
            \spy on \spyloc in node [left] at \spyshift;
        \end{tikzpicture} &
        \renewcommand{\spyloc}{(-0.6,-0.9)}
        \begin{tikzpicture}[spy using outlines={red,magnification=\magn,size=\ww}]
            \node {\includegraphics[width=\ww]{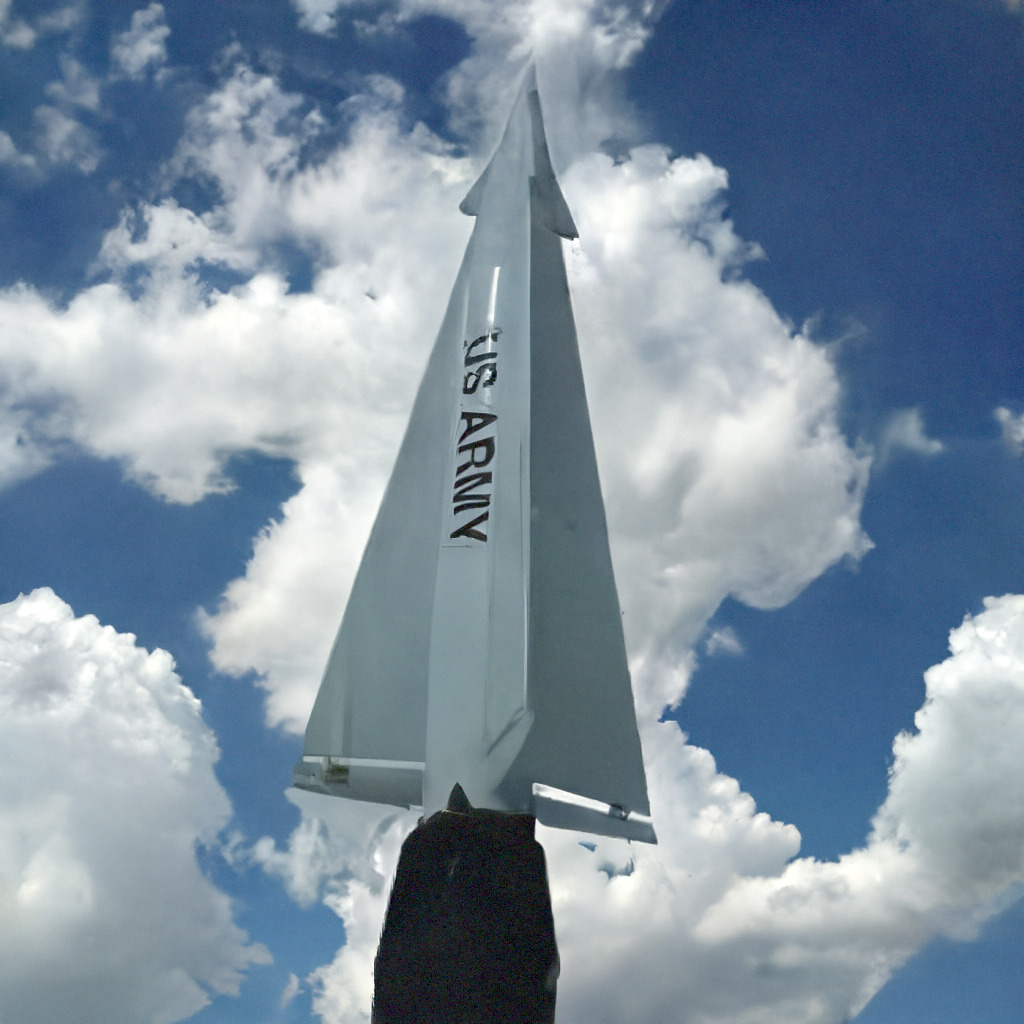}};
            \spy on \spyloc in node [left] at \spyshift;
        \end{tikzpicture} &
        \renewcommand{\spyloc}{(-0.6,-0.9)}
        \begin{tikzpicture}[spy using outlines={red,magnification=\magn,size=\ww}]
            \node {\includegraphics[width=\ww]{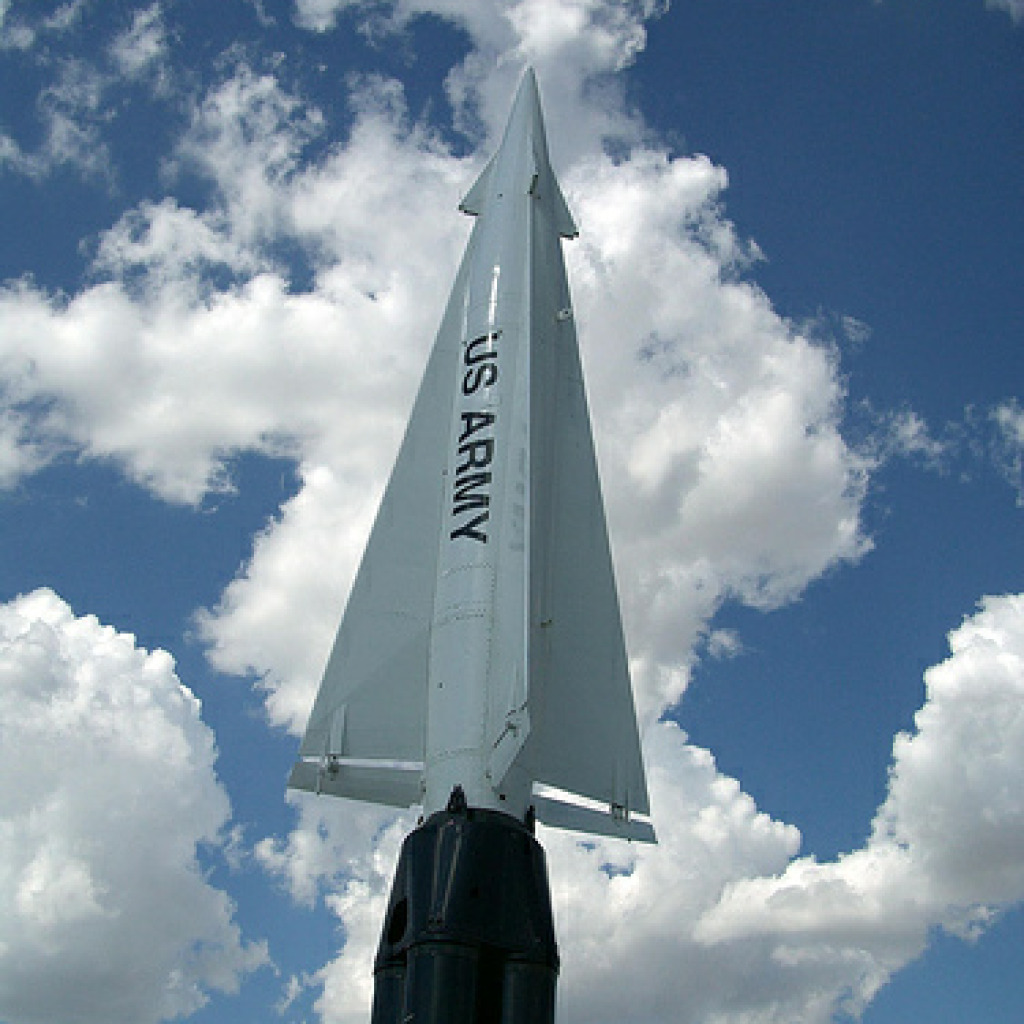}};
            \spy on \spyloc in node [left] at \spyshift;
        \end{tikzpicture} \\

        %%% cat 1016
        \renewcommand{\spyloc}{(-0.5,0.5)}
        \begin{tikzpicture}[spy using outlines={red,magnification=\magn,size=\ww}]
            \node {\includegraphics[width=\ww]{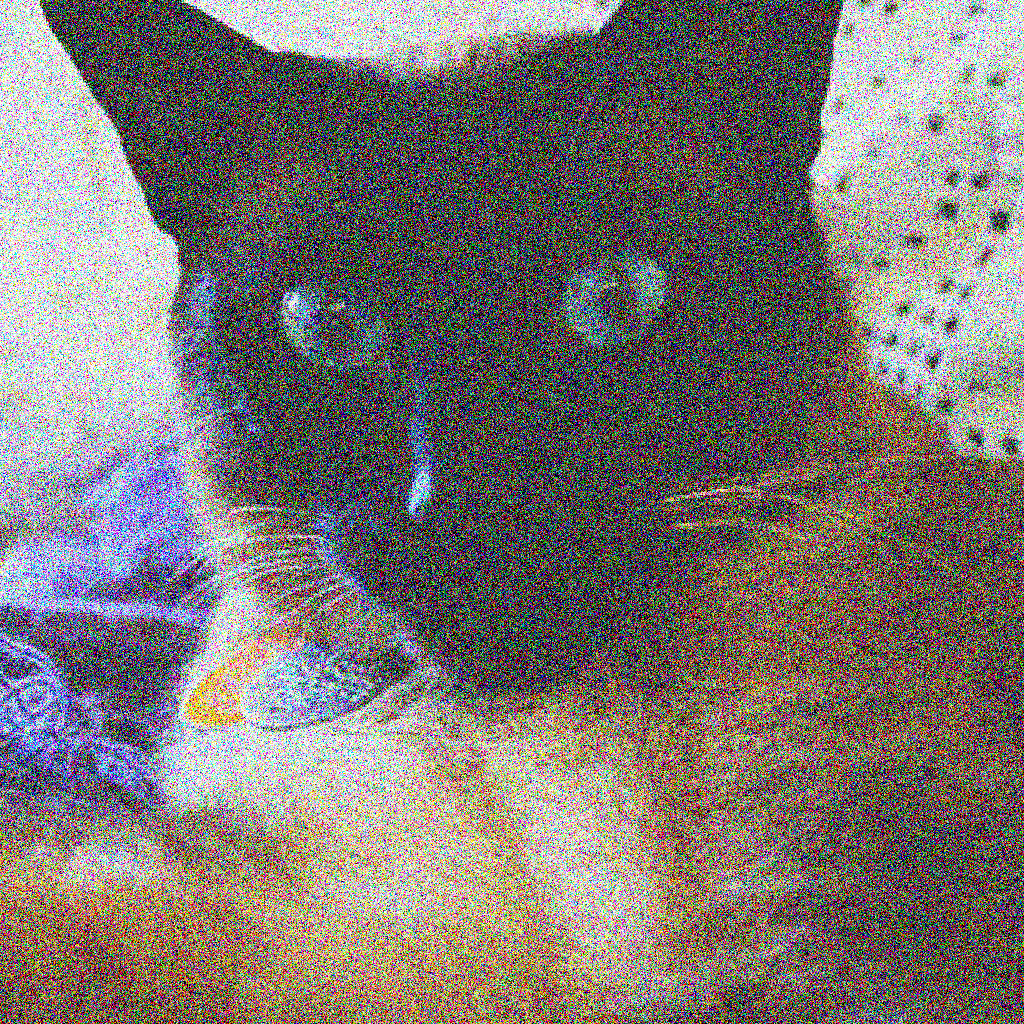}};
            \spy on \spyloc in node [left] at \spyshift;
        \end{tikzpicture} &
        \renewcommand{\spyloc}{(-0.5,0.5)}
        \begin{tikzpicture}[spy using outlines={red,magnification=\magn,size=\ww}]
            \node {\includegraphics[width=\ww]{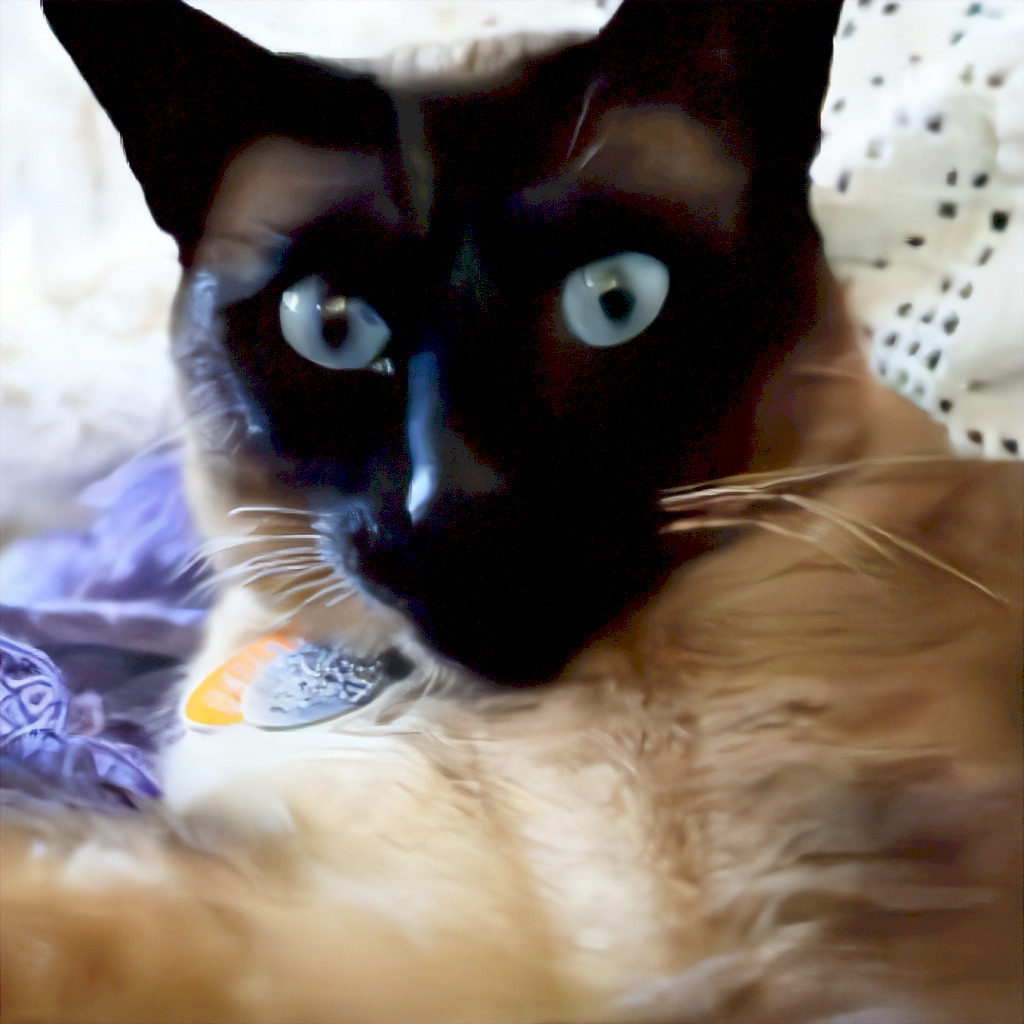}};
            \spy on \spyloc in node [left] at \spyshift;
        \end{tikzpicture} &
        \renewcommand{\spyloc}{(-0.5,0.5)}
        \begin{tikzpicture}[spy using outlines={red,magnification=\magn,size=\ww}]
            \node {\includegraphics[width=\ww]{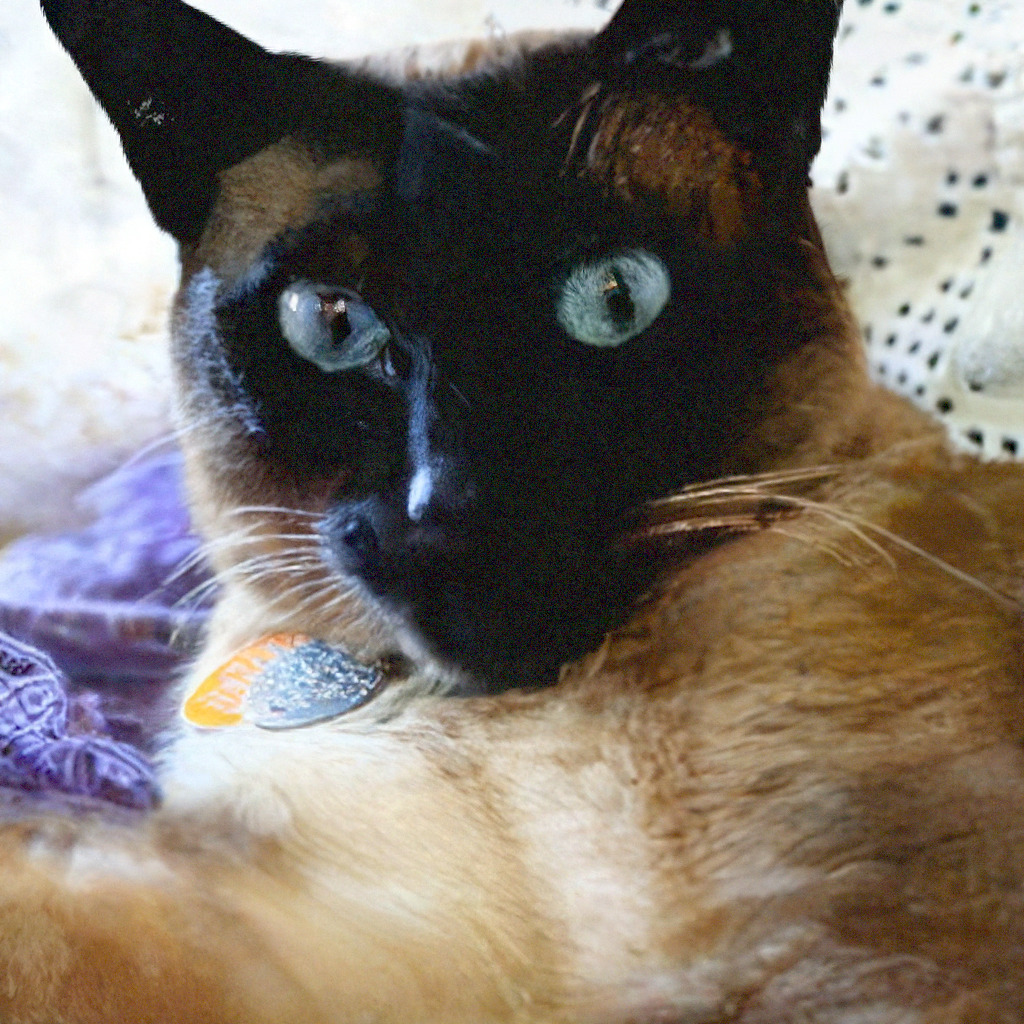}};
            \spy on \spyloc in node [left] at \spyshift;
        \end{tikzpicture} &
        \renewcommand{\spyloc}{(-0.5,0.5)}
        \begin{tikzpicture}[spy using outlines={red,magnification=\magn,size=\ww}]
            \node {\includegraphics[width=\ww]{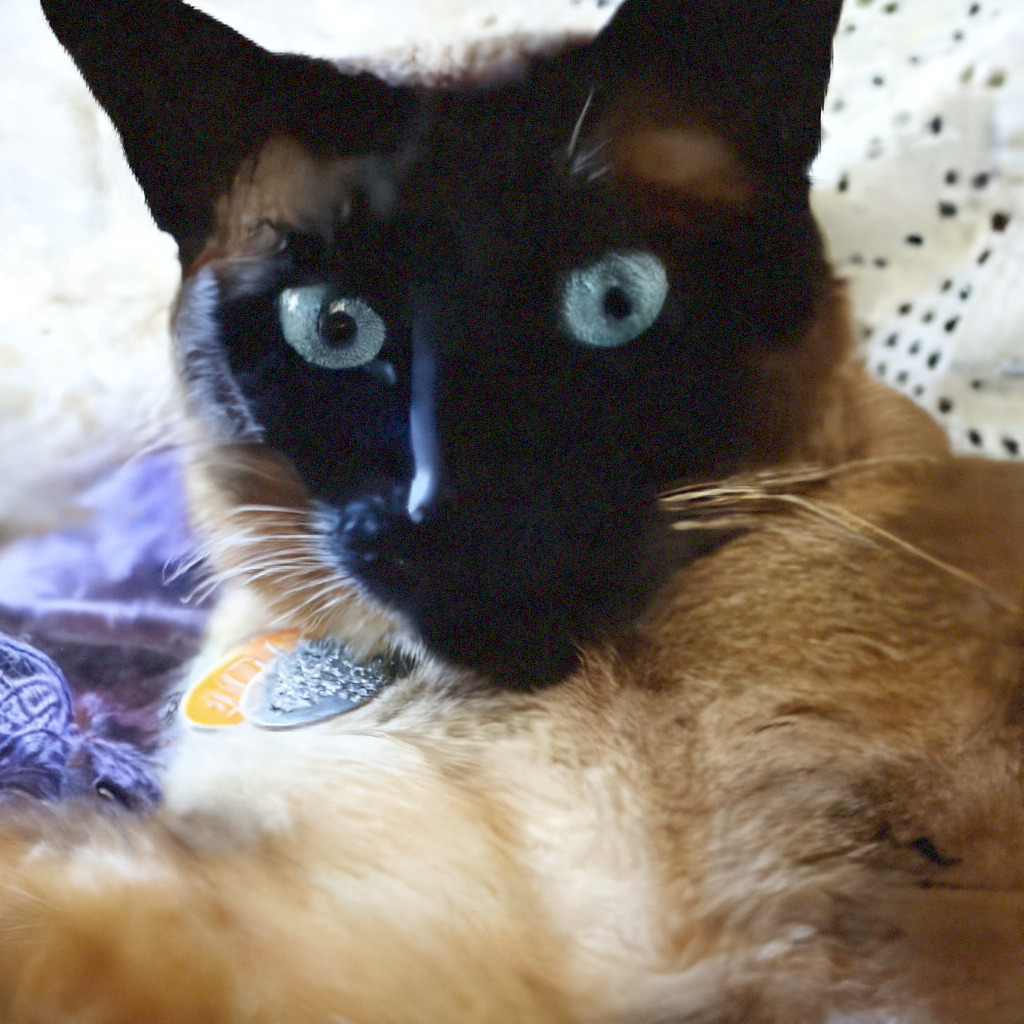}};
            \spy on \spyloc in node [left] at \spyshift;
        \end{tikzpicture} &
        \renewcommand{\spyloc}{(-0.5,0.5)}
        \begin{tikzpicture}[spy using outlines={red,magnification=\magn,size=\ww}]
            \node {\includegraphics[width=\ww]{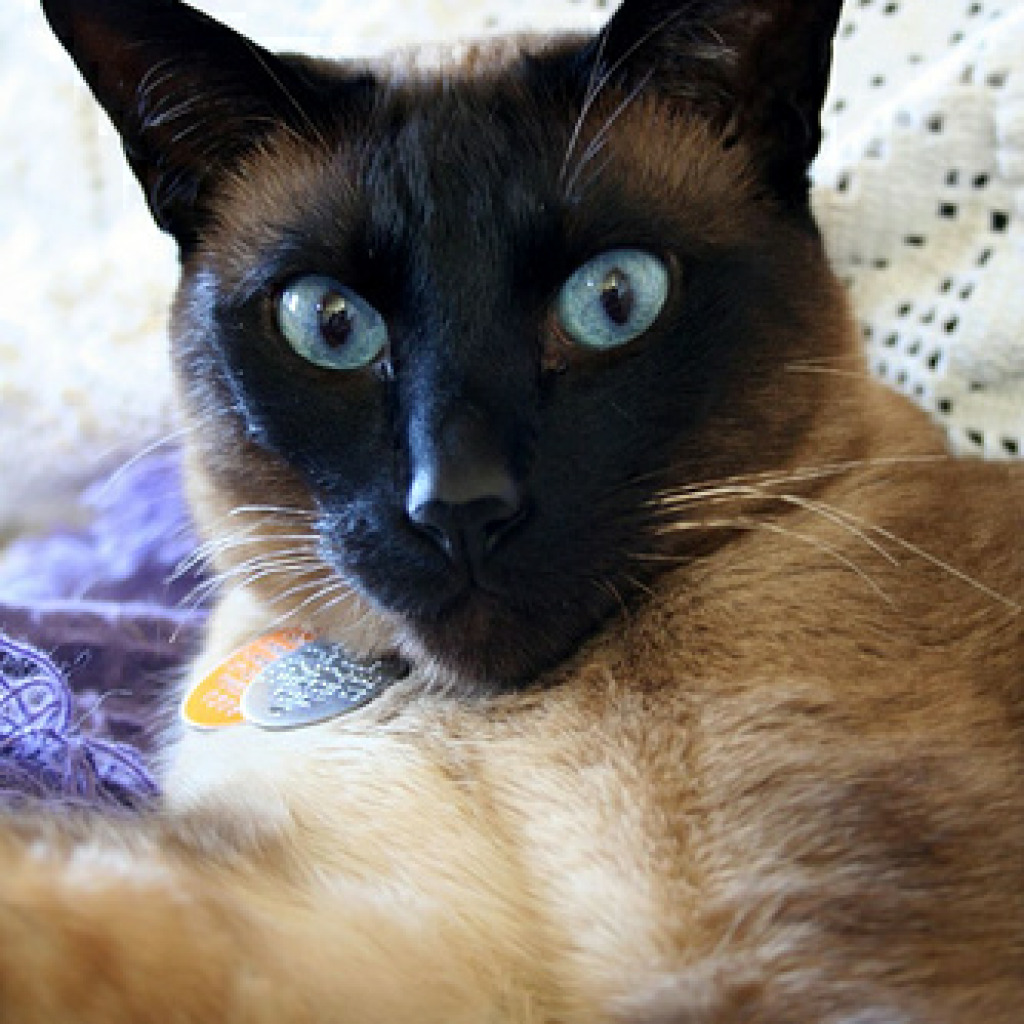}};
            \spy on \spyloc in node [left] at \spyshift;
        \end{tikzpicture} \\

        %% diver 246
    \begin{tikzpicture}[spy using outlines={red,magnification=\magn,size=\ww}]
        \node {\includegraphics[width=\ww]{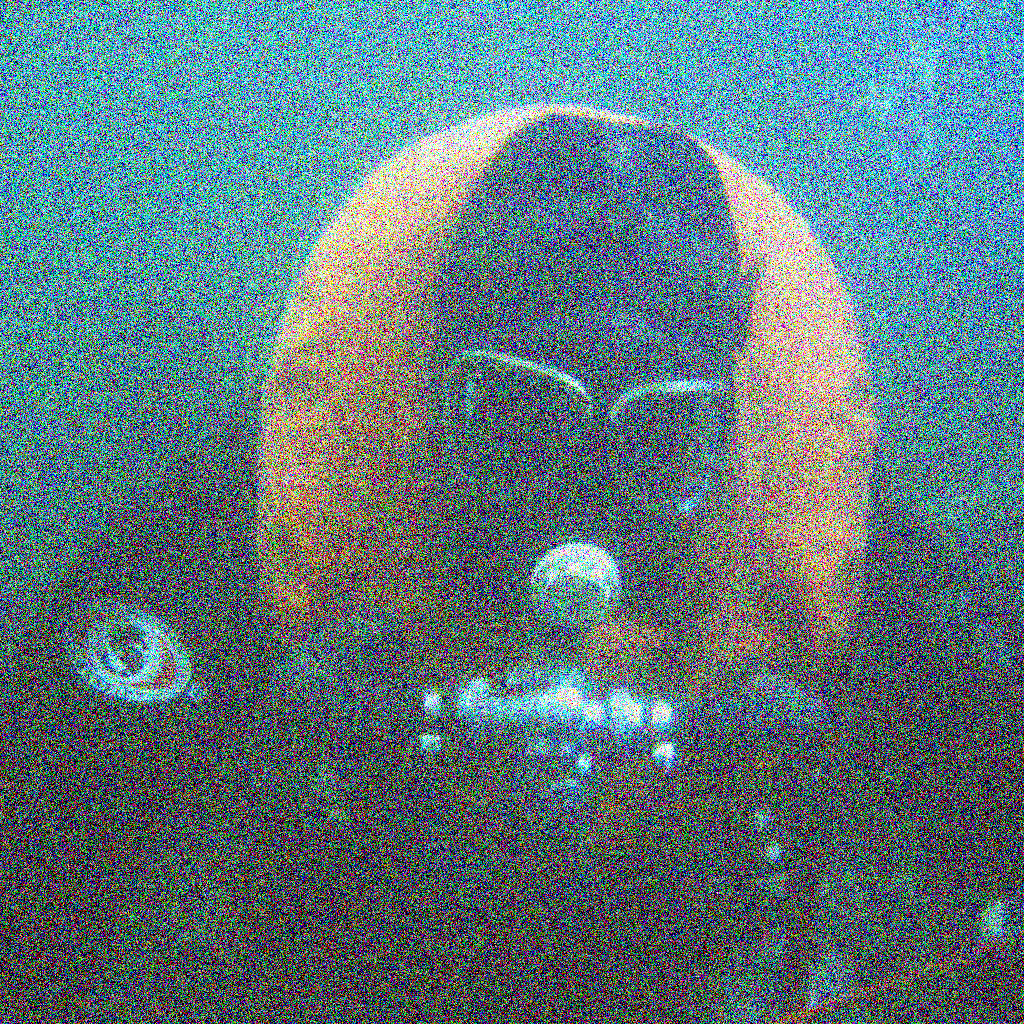}};
        \spy on \spyloc in node [left] at \spyshift;
    \end{tikzpicture} &
    \begin{tikzpicture}[spy using outlines={red,magnification=\magn,size=\ww}]
        \node {\includegraphics[width=\ww]{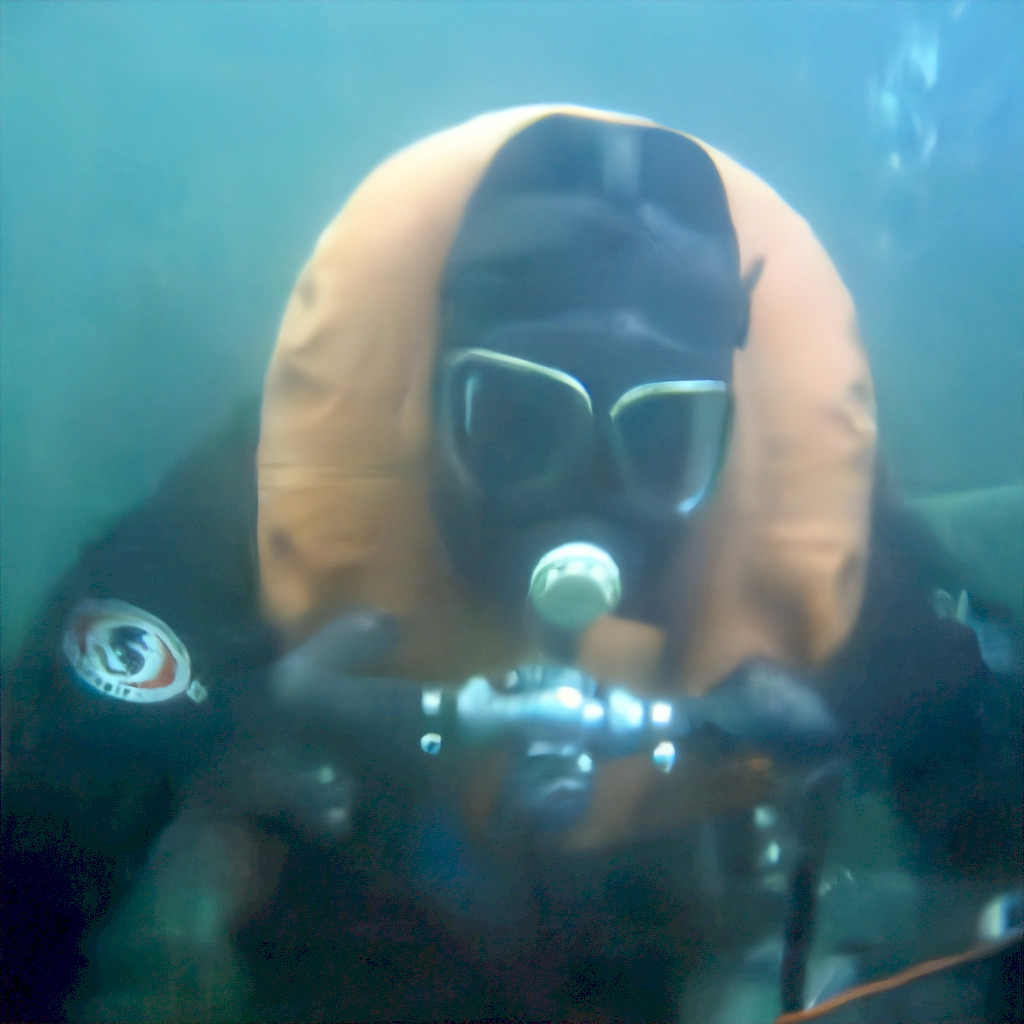}};
        \spy on \spyloc in node [left] at \spyshift;
    \end{tikzpicture} &
    \begin{tikzpicture}[spy using outlines={red,magnification=\magn,size=\ww}]
        \node {\includegraphics[width=\ww]{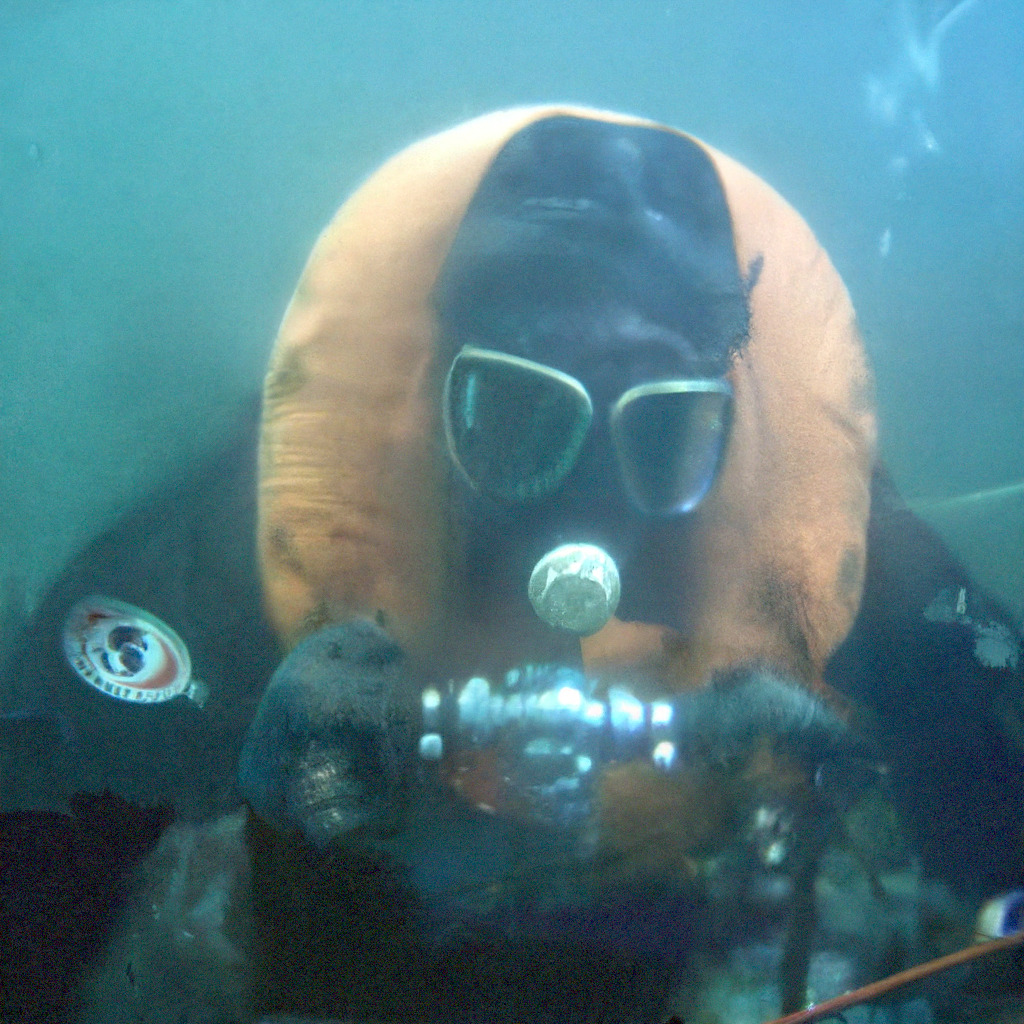}};
        \spy on \spyloc in node [left] at \spyshift;
    \end{tikzpicture} &
    \begin{tikzpicture}[spy using outlines={red,magnification=\magn,size=\ww}]
        \node {\includegraphics[width=\ww]{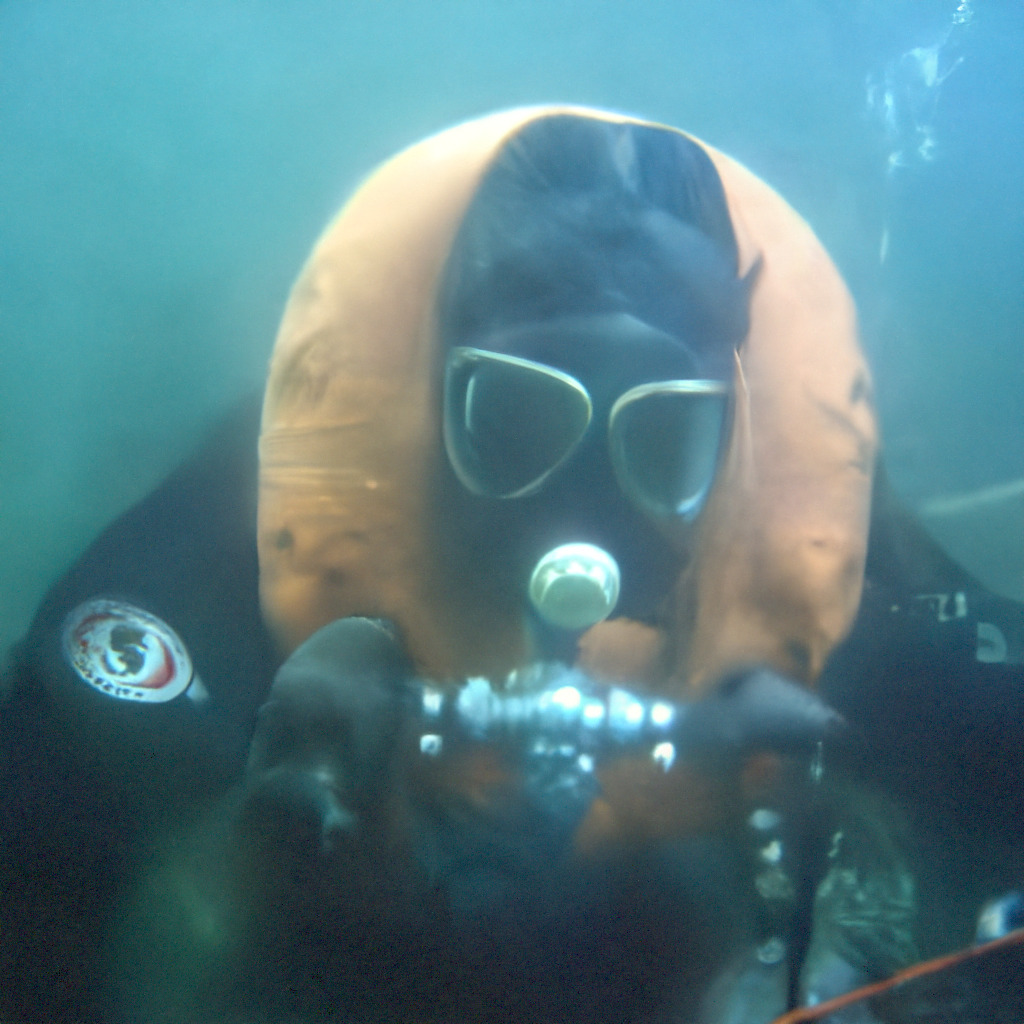}};
        \spy on \spyloc in node [left] at \spyshift;
    \end{tikzpicture} &
    \begin{tikzpicture}[spy using outlines={red,magnification=\magn,size=\ww}]
        \node {\includegraphics[width=\ww]{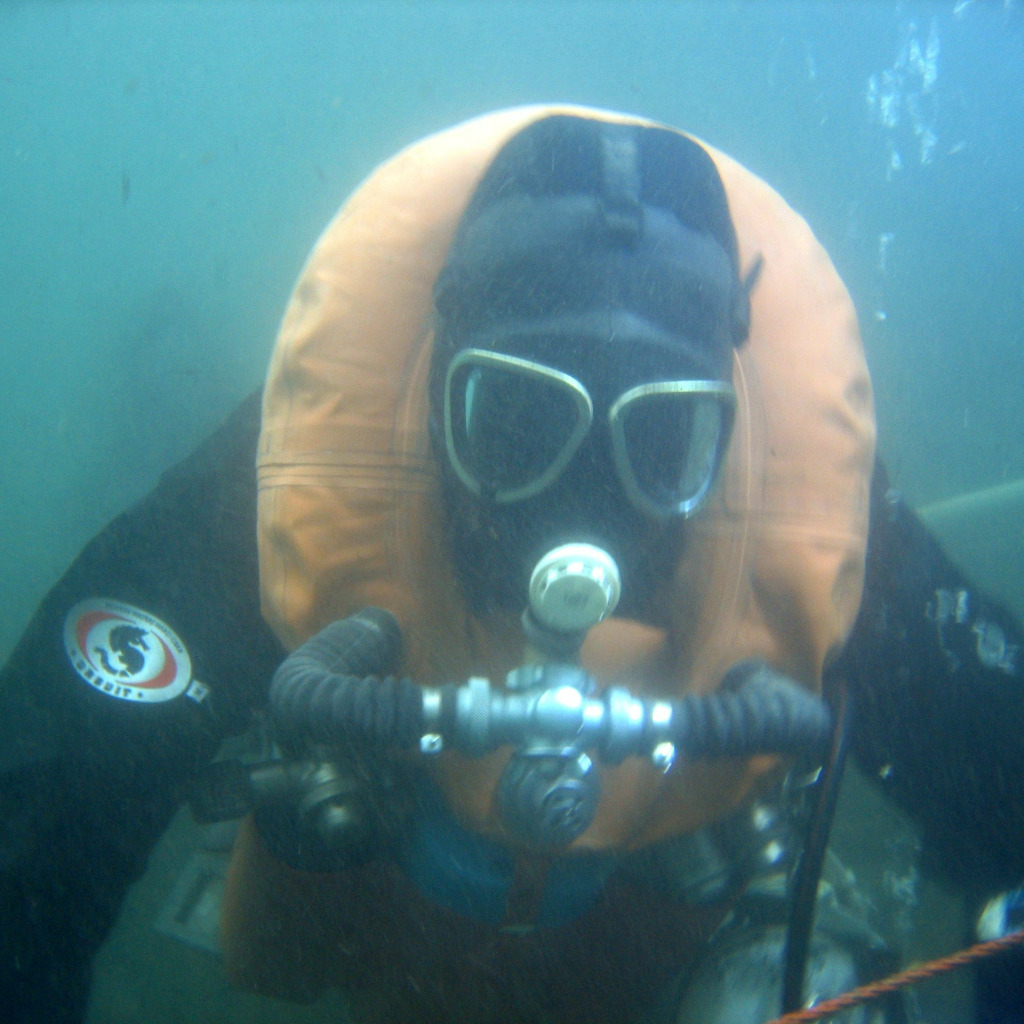}};
        \spy on \spyloc in node [left] at \spyshift;
    \end{tikzpicture} \\   
\end{tabular}
\end{center}
\vspace{-1em}
\caption{Comparison between different denoising methods on images with noise gain of 16.} 
\label{fig:imagenet_comparison-full-page-1}
\end{figure*}

\begin{figure*}[htbp]
\setlength{\ww}{0.192\textwidth}
\begin{center}
\newcommand{\magn}{5.0}
\newcommand{\spyloc}{(0.28,0.15)}    
\newcommand{\spyshift}{(1.675,-3.4)}
\newcommand{\bright}{0.1 1 0.1 1 0.1 1}
\small\addtolength{\tabcolsep}{-8.5pt}
\begin{tabular}{ccccc}
    Noisy & HINet~\cite{chen2021hinet} & Baseline & Ours & Clean GT\\
        %% dessert 1031
        \renewcommand{\spyloc}{(0.56,0.44)}
        \renewcommand{\magn}{4.0}
        \begin{tikzpicture}[spy using outlines={red,magnification=\magn,size=\ww}]
            \node {\includegraphics[width=\ww]{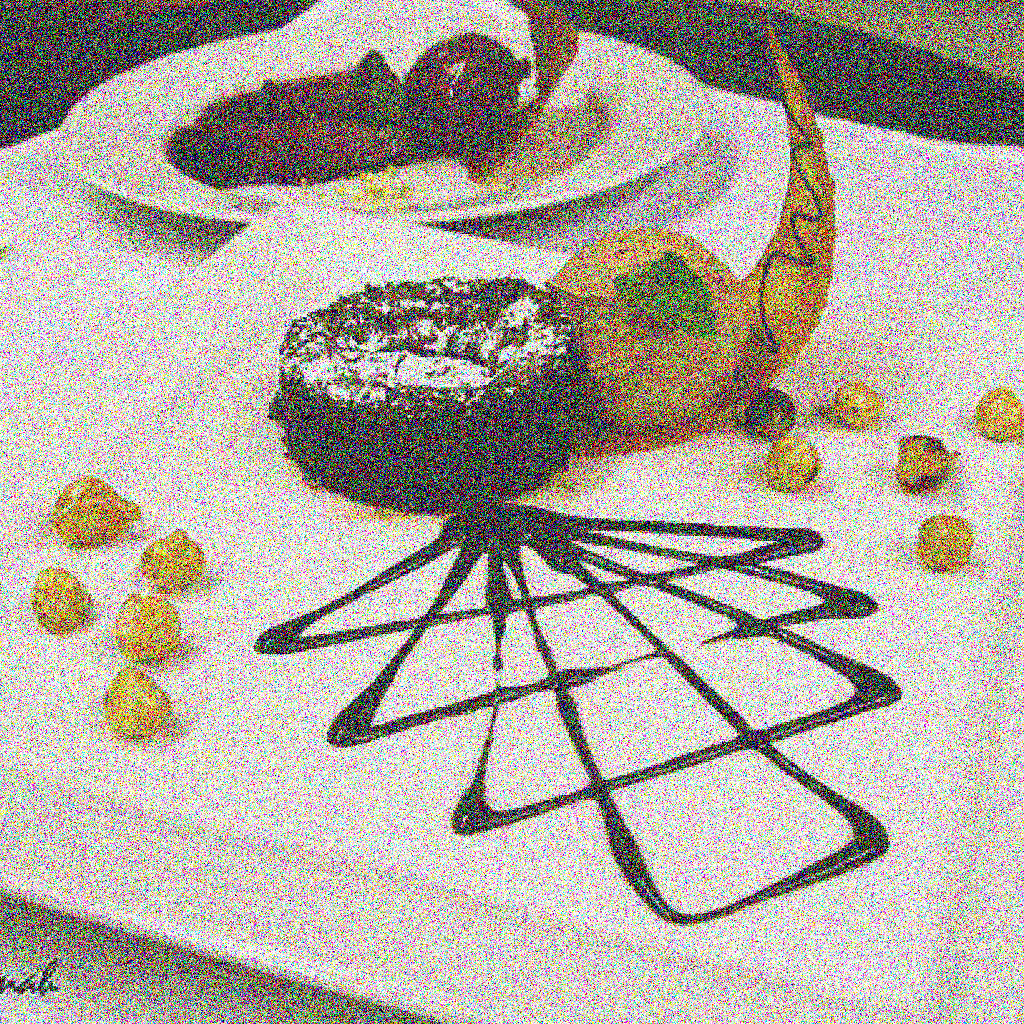}};
            \spy on \spyloc in node [left] at \spyshift;
        \end{tikzpicture} &
        \renewcommand{\spyloc}{(0.56,0.44)}
        \renewcommand{\magn}{4.0}
        \begin{tikzpicture}[spy using outlines={red,magnification=\magn,size=\ww}]
            \node {\includegraphics[width=\ww]{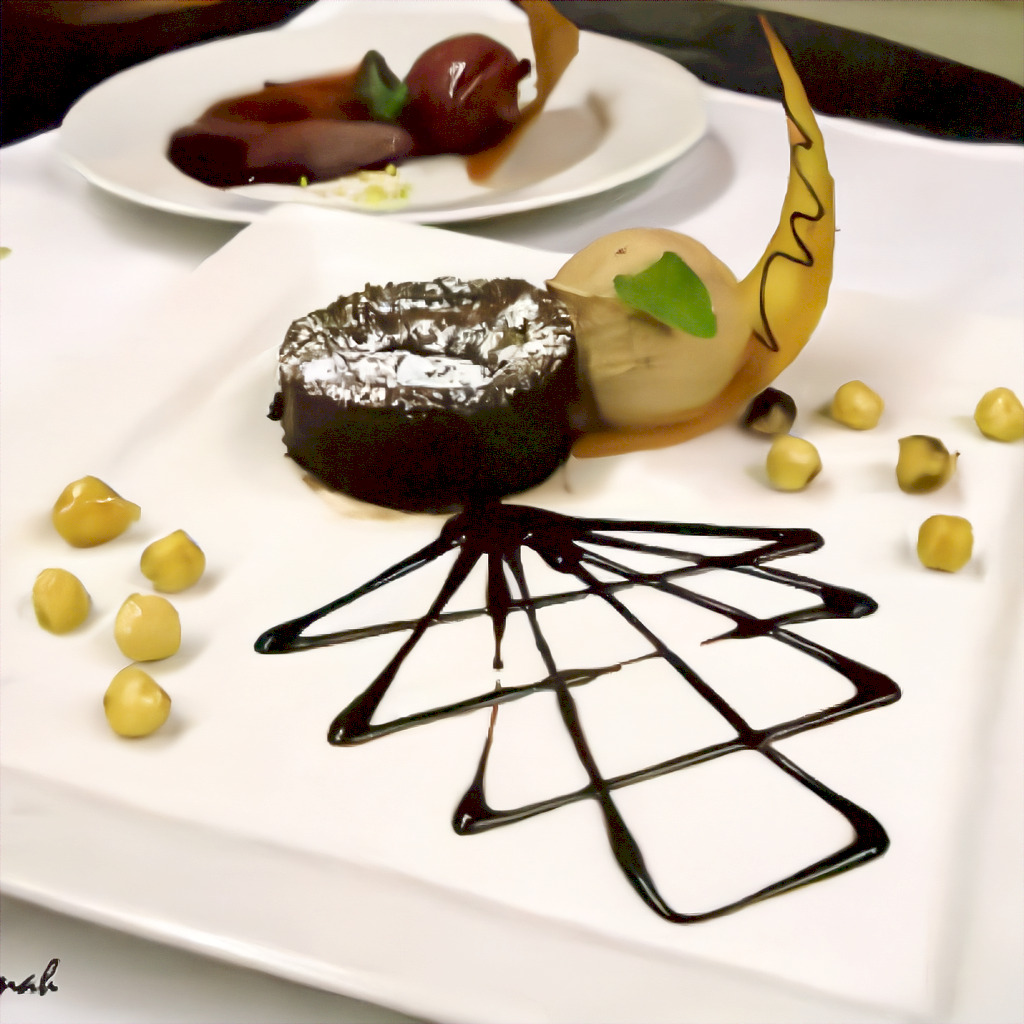}};
            \spy on \spyloc in node [left] at \spyshift;
        \end{tikzpicture} &
        \renewcommand{\spyloc}{(0.56,0.44)}
        \renewcommand{\magn}{4.0}
        \begin{tikzpicture}[spy using outlines={red,magnification=\magn,size=\ww}]
            \node {\includegraphics[width=\ww]{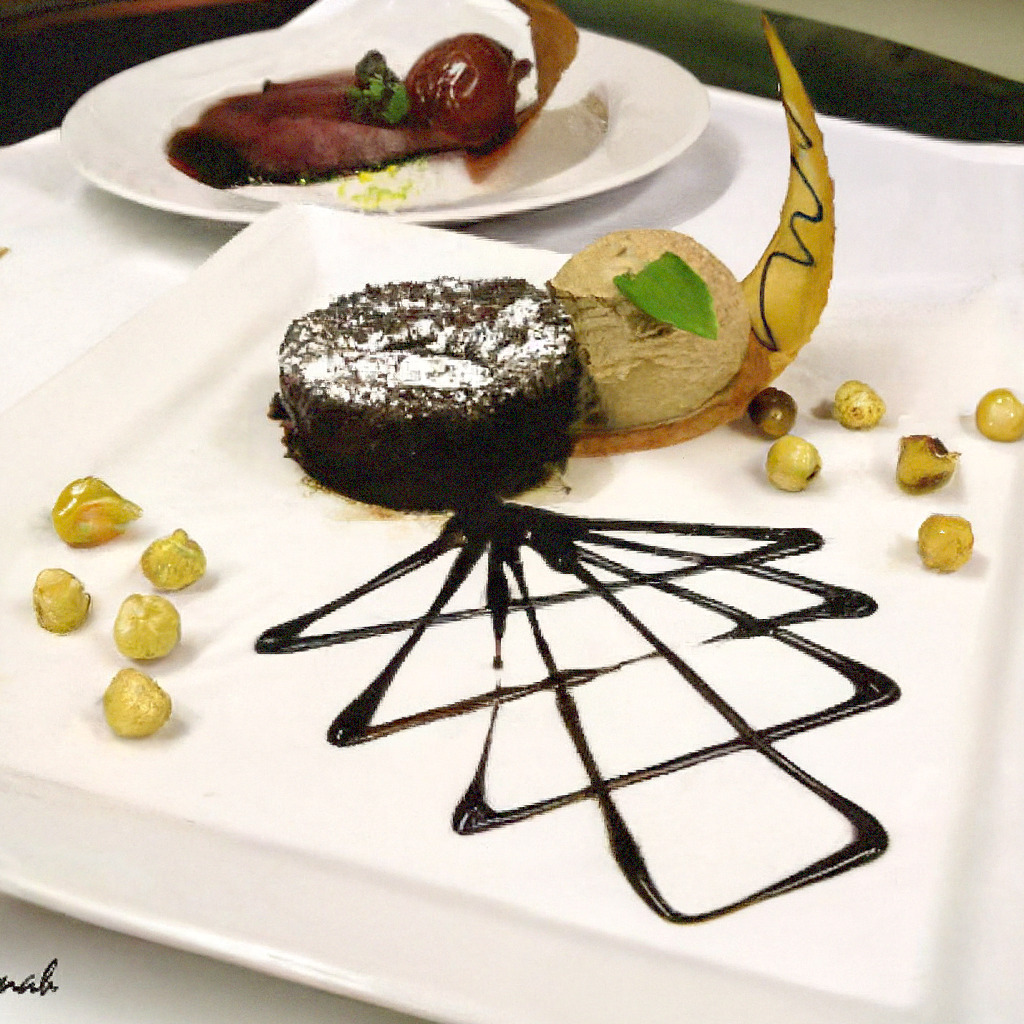}};
            \spy on \spyloc in node [left] at \spyshift;
        \end{tikzpicture} &
        \renewcommand{\spyloc}{(0.56,0.44)}
        \renewcommand{\magn}{4.0}
        \begin{tikzpicture}[spy using outlines={red,magnification=\magn,size=\ww}]
            \node {\includegraphics[width=\ww]{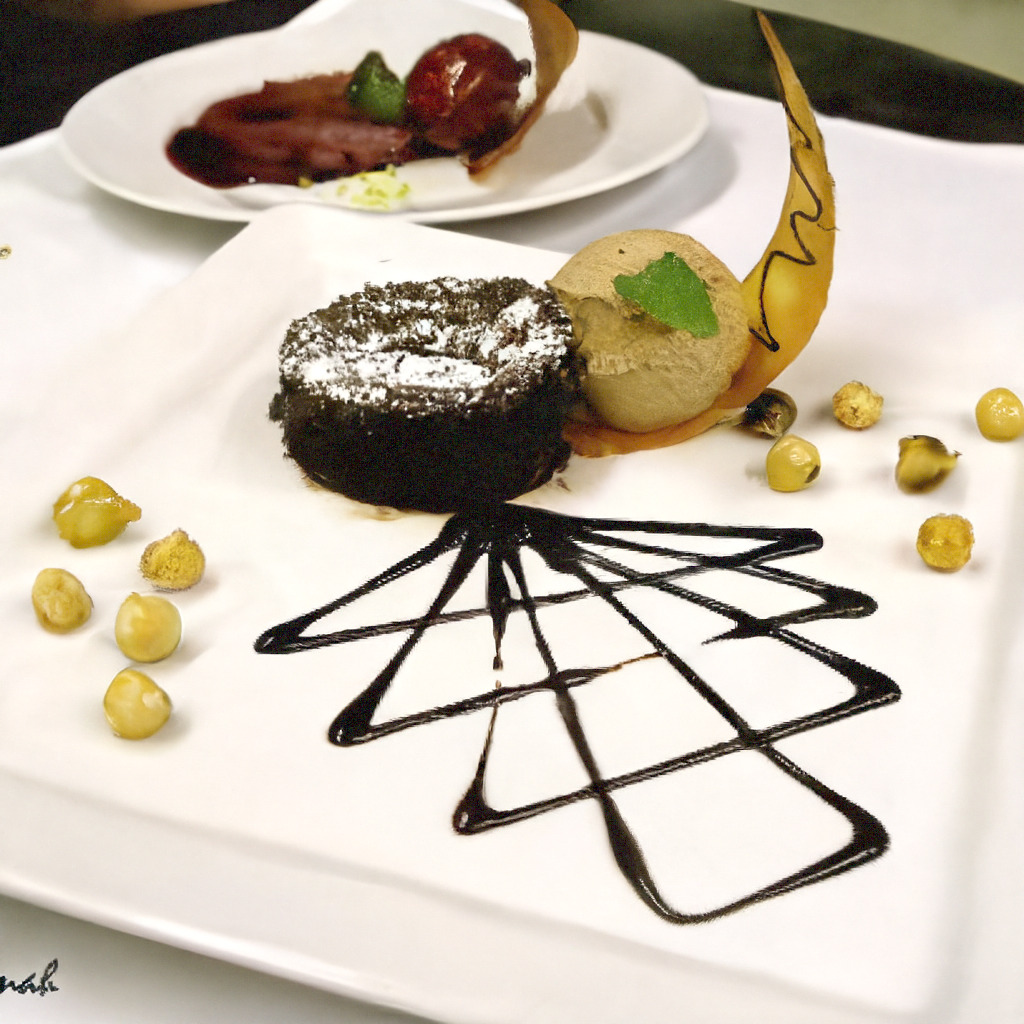}};
            \spy on \spyloc in node [left] at \spyshift;
        \end{tikzpicture} &
        \renewcommand{\spyloc}{(0.56,0.44)}
        \renewcommand{\magn}{4.0}
        \begin{tikzpicture}[spy using outlines={red,magnification=\magn,size=\ww}]
            \node {\includegraphics[width=\ww]{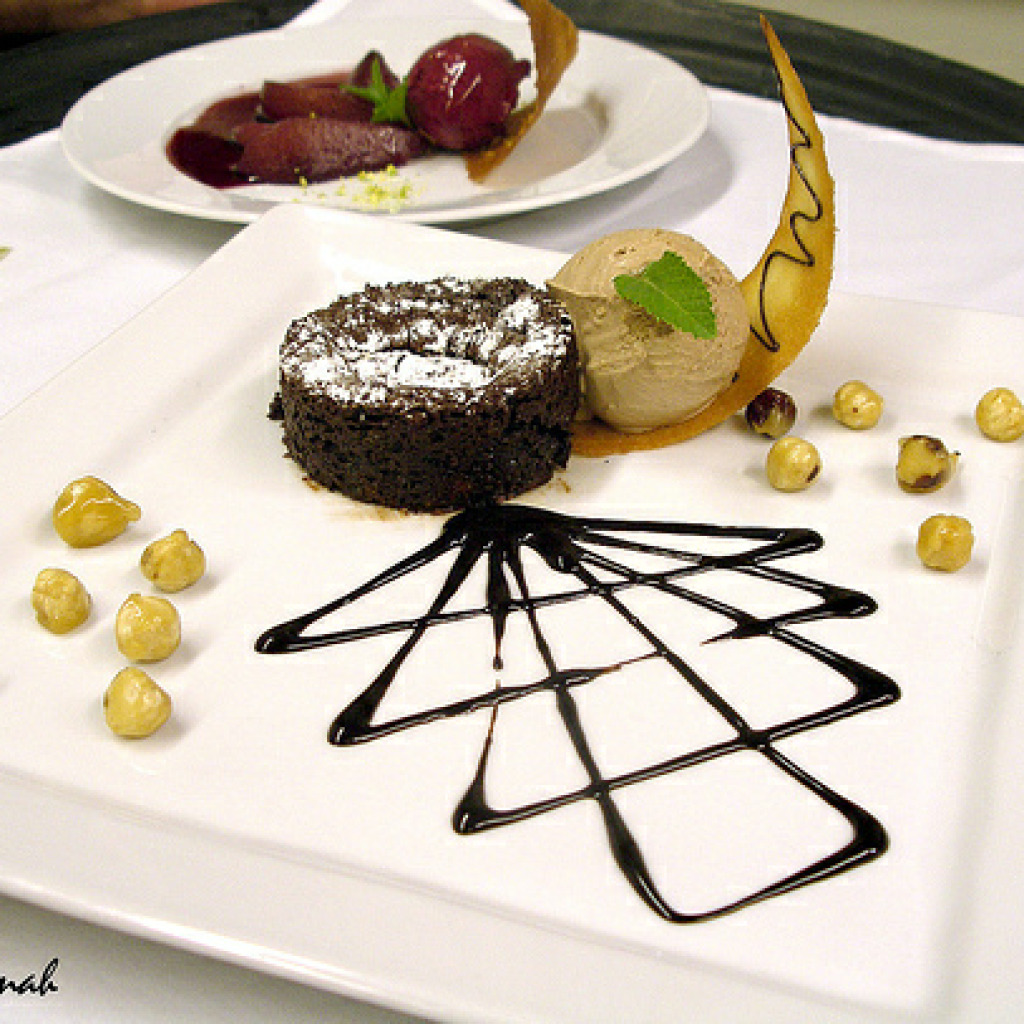}};
            \spy on \spyloc in node [left] at \spyshift;
        \end{tikzpicture} \\

        %% wolf 684
        \renewcommand{\spyloc}{(0.66,-0.6)}
        \renewcommand{\magn}{3.0}
        \begin{tikzpicture}[spy using outlines={red,magnification=\magn,size=\ww}]
            \node {\includegraphics[width=\ww]{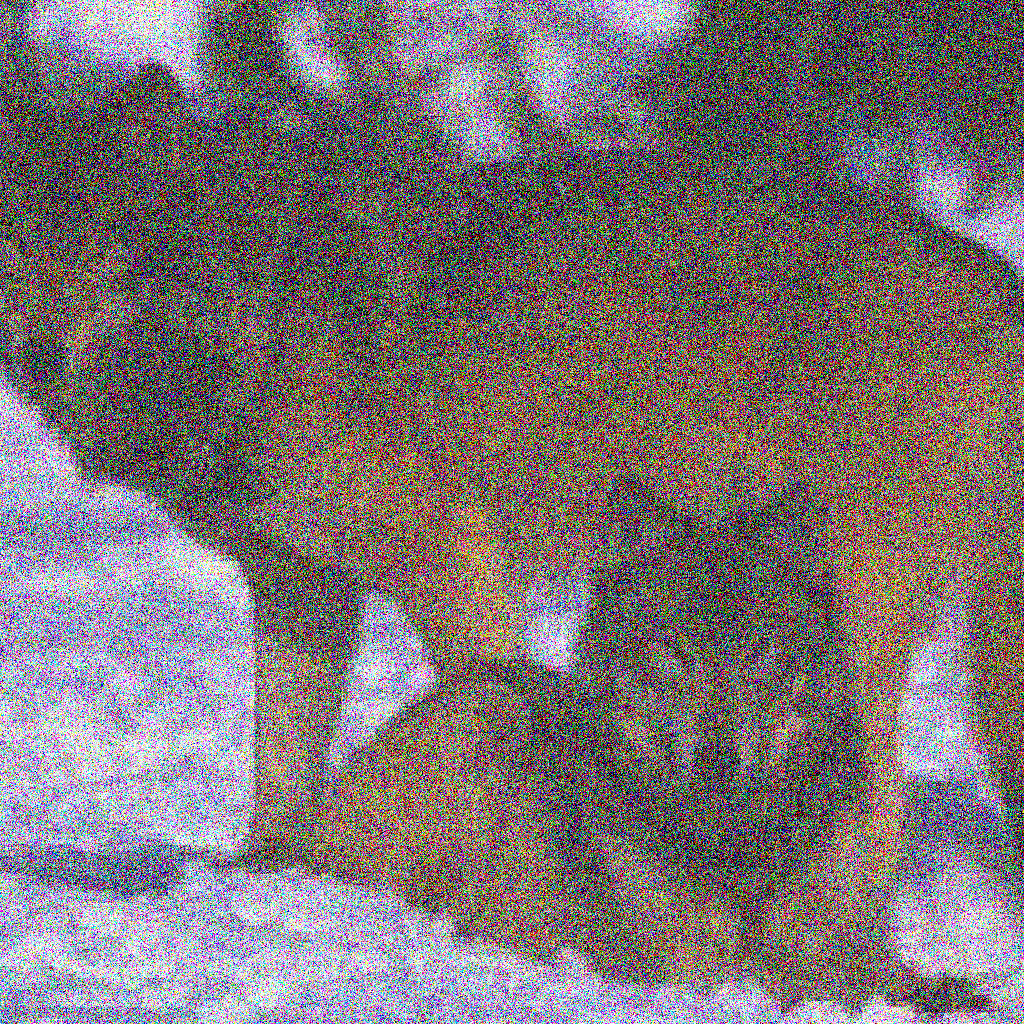}};
            \spy on \spyloc in node [left] at \spyshift;
        \end{tikzpicture} &
        \renewcommand{\spyloc}{(0.66,-0.6)}
        \renewcommand{\magn}{3.0}
        \begin{tikzpicture}[spy using outlines={red,magnification=\magn,size=\ww}]
            \node {\includegraphics[width=\ww]{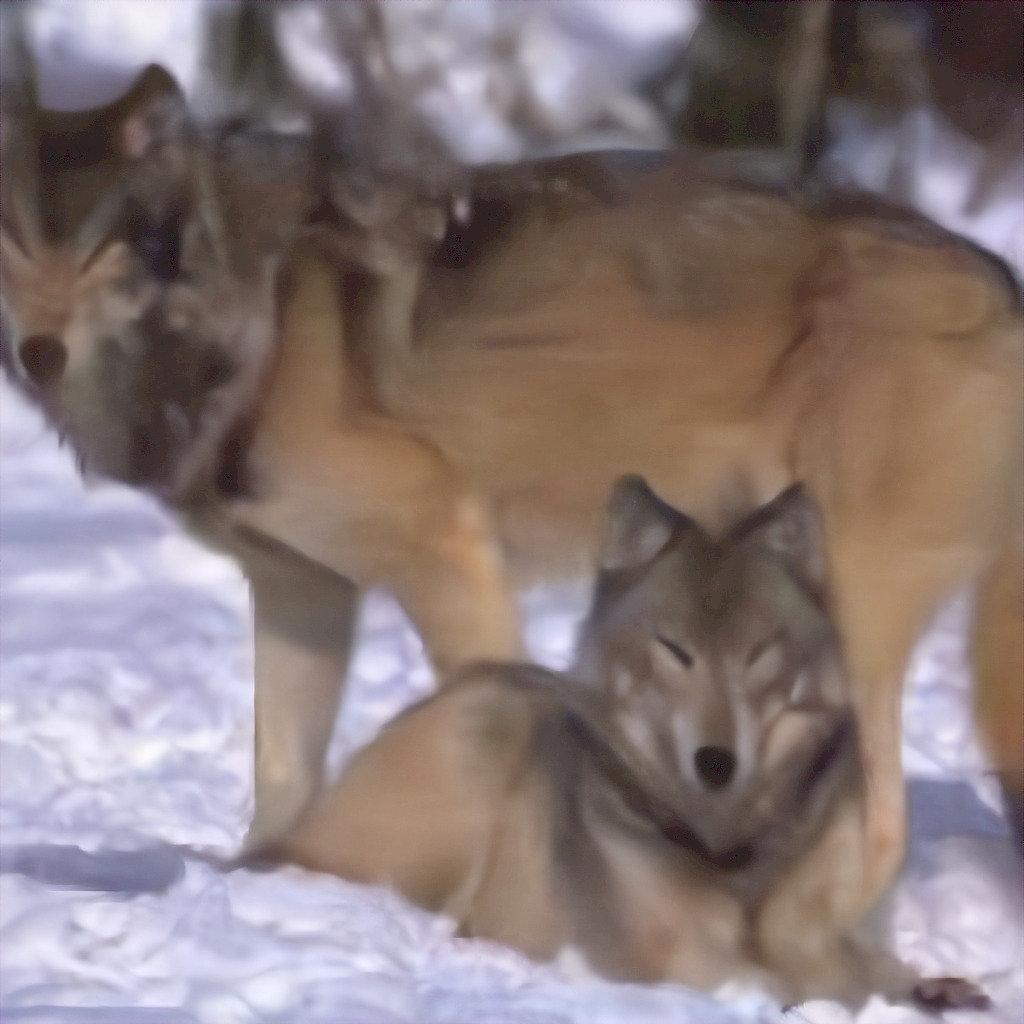}};
            \spy on \spyloc in node [left] at \spyshift;
        \end{tikzpicture} &
        \renewcommand{\spyloc}{(0.66,-0.6)}
        \renewcommand{\magn}{3.0}
        \begin{tikzpicture}[spy using outlines={red,magnification=\magn,size=\ww}]
            \node {\includegraphics[width=\ww]{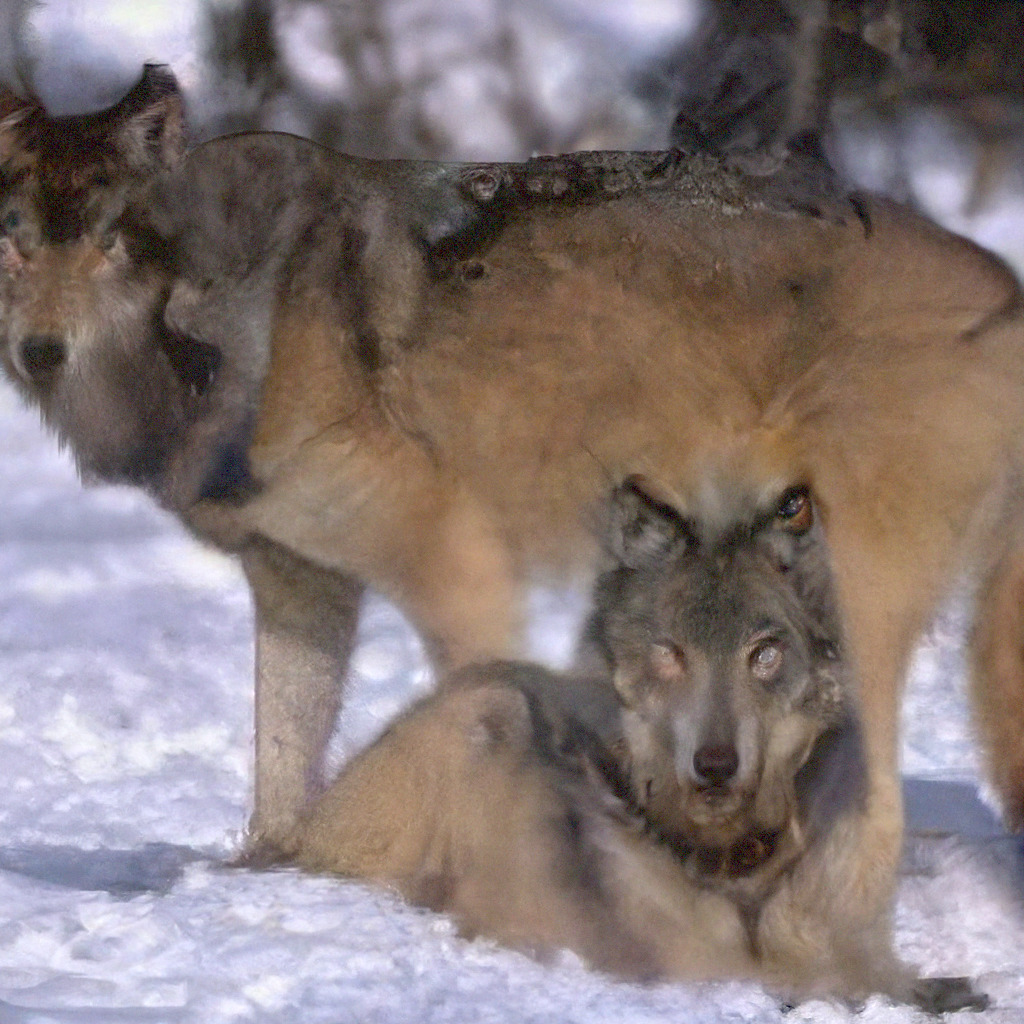}};
            \spy on \spyloc in node [left] at \spyshift;
        \end{tikzpicture} &
        \renewcommand{\spyloc}{(0.66,-0.6)}
        \renewcommand{\magn}{3.0}
        \begin{tikzpicture}[spy using outlines={red,magnification=\magn,size=\ww}]
            \node {\includegraphics[width=\ww]{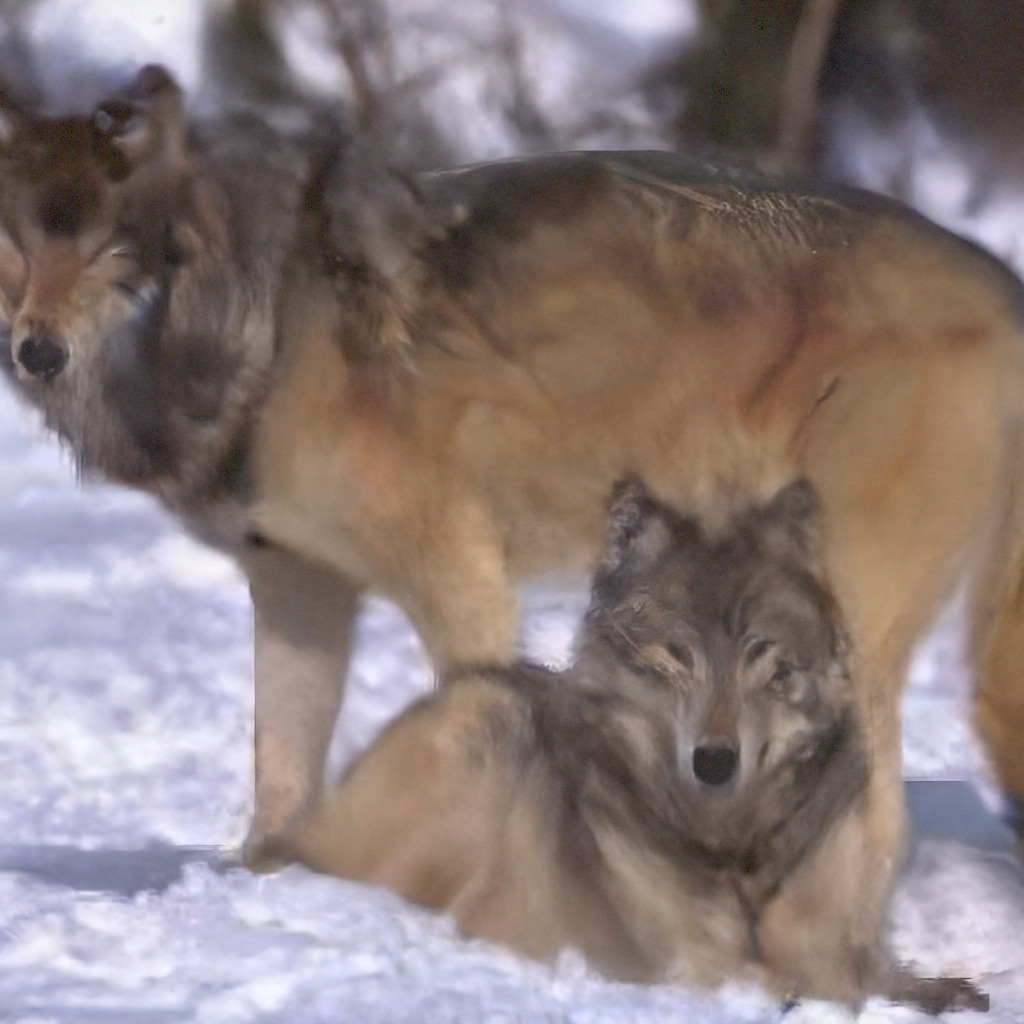}};
            \spy on \spyloc in node [left] at \spyshift;
        \end{tikzpicture} &
        \renewcommand{\spyloc}{(0.66,-0.6)}
        \renewcommand{\magn}{3.0}
        \begin{tikzpicture}[spy using outlines={red,magnification=\magn,size=\ww}]
            \node {\includegraphics[width=\ww]{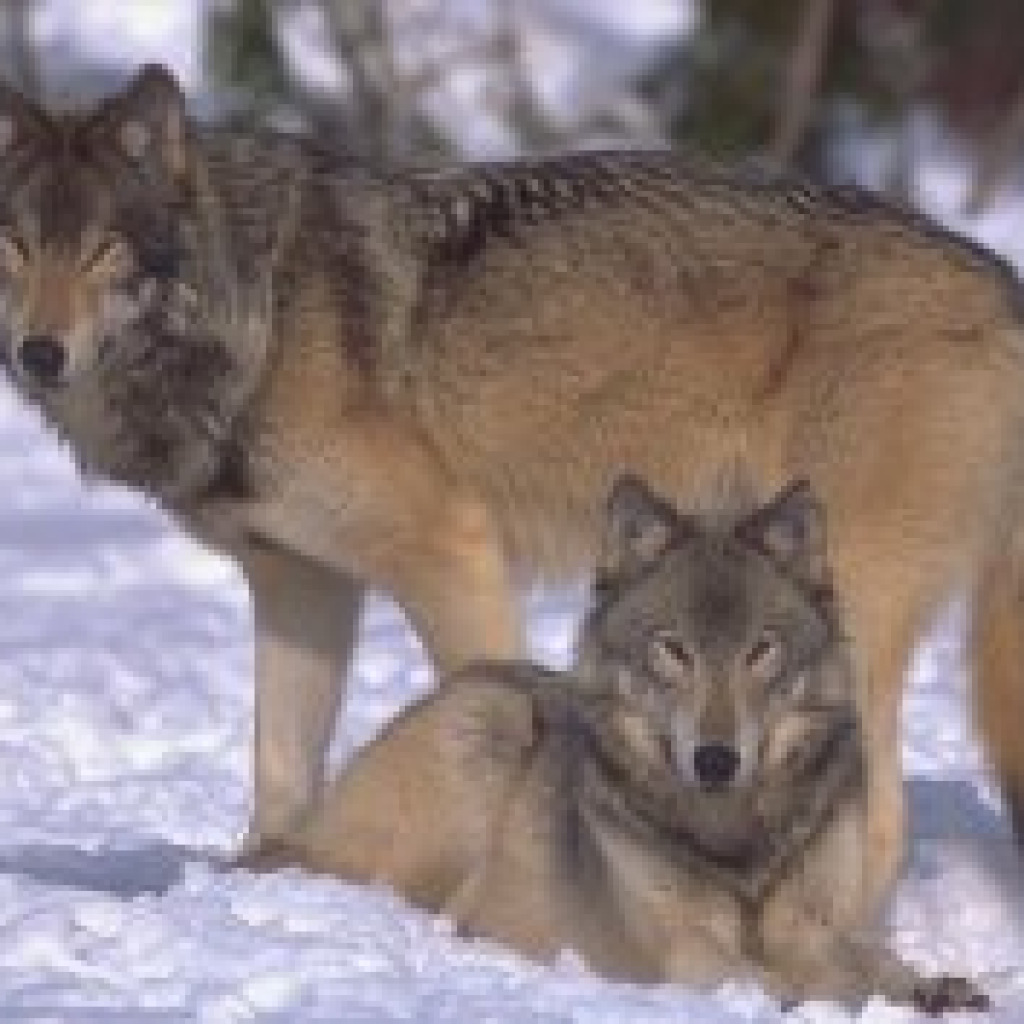}};
            \spy on \spyloc in node [left] at \spyshift;
        \end{tikzpicture} \\
   
\end{tabular}
\end{center}
\vspace{-1em}
\caption{Comparison between different denoising methods on images with noise gain of 16.} 
\label{fig:imagenet_comparison-full-page-2}
\end{figure*}

This can be further seen in \figref{fig:imagenet_comparison-full-page-1} and \figref{fig:imagenet_comparison-full-page-2}, where we showcase denoising results of these three models for several inputs with noise gain of $16$ (comparisons at other noise levels are included in the supplementary). Even at this relatively high noise level, all three models manage to remove most of the noise. However, the results of HINet suffer from considerable over-smoothing and lack high-frequency details. On the other hand, both \svdd and the baseline diffusion models manage to generate fine details. While the baseline diffusion model generally generates more details than \svdd, it eliminates less noise (top example) and furthermore, occasionally exhibits hallucinations (see the first two examples).
We hypothesize that this difference between our method and the baseline stems from fine-tuning the baseline to adapt it to our diffusion noise model, Eq.~\eqref{eq:ve_diffusion}. We conjecture that fine-tuning causes the model to lose some of its prior, instead allowing it to make more effective use of the underlying signal, by using the noisy image as the starting point.

Overall, we see that our method yields comparable performance to the state-of-the-art, while producing more realistic images. At the same time, our method retains more fidelity to the underlying signal and removes more noise than the baseline diffusion approach.

Since the diffusion baseline always starts from complete noise, its runtime is fixed ($\sim\!22$ seconds), regardless of the noise level in the input image. Starting the diffusion process from the noisy image in \svdd yields results in runtime that depends on the noise levels in the image, ranging from $\sim\!3$ seconds to less than a second for the least noisy images. 

\subsection{Ablation} \label{sec:ablation}
We validate the importance of different aspects of our approach by the ablation study in \tabref{tab:ablation_study}.
We compare the results to the baseline diffusion model that is initialized with \emph{complete noise} and conditioned on the noisy image (denoted A in the table) and to versions where diffusion is initialized with the \emph{noisy input image} (denoted by B, C).
When initializing the diffusion process with the noisy image, we consider unconditioned (B) and conditioned (C) variants.

The \emph{unconditioned} variants differ in the type of their input images: B1, where the input values are clipped to avoid negative values; and B2, a variant where input images are allowed to have negative values.
For the \emph{conditioned} setup we consider three training schemes: C1, the standard training process, and two versions that try to handle the correlation described in Section~\ref{sec:noise_correlation} -- C2, a version that enforces the starting point of the diffusion $\xt{\tmap}$ to be equal to the noisy input $\y$ in $1\%$ of training iterations; and C3, our full \svdd framework that incorporates Eq.~\eqref{eq:train_ddpm}.
All the ablation experiments are done with gain level 16, and the results are averaged over $80$ images.

The comparison to the baseline A is discussed in the previous section.
The \emph{unconditioned} version B1 fails to restore the clean signal, mainly because it is not robust to the zero clipped values. When the original noisy image is not available during the process, the prediction of $\xt{t}$ at each diffusion step is shifted and ``loses" the correct intensity levels. This is supported by the comparison with B2.

The standard \emph{conditioned} version C1 emphasizes the importance of our training scheme that takes into account the correlation between the two sources of noise.
In C2, we practically apply Eq.~\eqref{eq:train_ddpm} only for the first step of diffusion and only for $1\%$ of the training iterations (as explained in Section~\ref{sec:noise_correlation}, this is equivalent to training on samples with $\xt{\tmap}=\y$), which slightly improves the results.
However, to achieve good restoration, one must consider the correlation throughout the entire process, which is supported by the improved results achieved by our training scheme C3. 
\begin{table}[t]
\begin{minipage}{1\linewidth}
\renewcommand\footnoterule{}
\renewcommand*{\thempfootnote}{\roman{mpfootnote}}
\setlength{\tabcolsep}{2pt}
\begin{center}
\footnotesize
\begin{tabular}{l l c c c}
\hline
 & & \textbf{PSNR $\uparrow$} & \textbf{SSIM $\uparrow$} & \textbf{LPIPS $\downarrow$} \\
\hline
& \textbf{Initialized with complete noise} & \\
\hline
A & Conditioned (baseline) & 23.76 & 0.46 & 0.441 \\
\hline
& \textbf{Initialized with $\y$} &\\
\hline
B1 & Unconditioned & 15.71 & 0.41 & 0.508 \\
B2 & Unconditioned, without clipping & 22.25 & 0.36 & 0.520 \\
\hline
C1 & Conditioned, standard training & 12.59 & 0.07 & 0.759 \\
C2 & Conditioned, oversampling $\xt{\tmap}=\y$ & 16.06 & 0.16 & 0.665 \\
C3 & \svdd  & \bf{24.56} & \bf{0.54} & \bf{0.438} \\
\end{tabular}
    \caption{Ablation study (under noise gain 16), averaged over $80$ images. See Section~\ref{sec:ablation} for details.}
\label{tab:ablation_study}
\end{center}
\end{minipage}
\vspace{-1.5em}
\end{table}
\section{Conclusions}
We have presented a new diffusion-based framework for the task of single image denoising, which leverages the natural rich image prior learned by generative denoising diffusion models.
Our framework adapts denoising diffusion to utilize the noisy input image as both the condition and the starting point of the diffusion process.
To enable the integration of a realistic noisy image as a sample in the diffusion process, we have proposed a novel denoising diffusion formulation that admits a spatially-variant time embedding, with supporting training and inference schemes. 

We believe that this novel formulation can be potentially applied to any non-uniform noise distribution. 
Additionally, we have addressed a phenomenon that occurs when initializing and conditioning the diffusion process with the same noisy input image, and have mitigated it with a suitable training scheme.
Our qualitative and quantitative results show improved handling of the distortion-perception trade-off, balancing faithful image reconstruction with generation of realistic fine details and textures.
Furthermore, our formulation also significantly reduces the numer of required diffusion steps. 
In the future, we aim to further distill the rich knowledge hidden in the backbone model, and expand the scope and applicability of our approach to complex real-world scenarios.

{\small
\bibliographystyle{ieee_fullname}
\bibliography{egbib}
}

\newpage

\appendix
\thispagestyle{empty}
\onecolumn 
\section{Proofs and derivations}
\subsection{Diffusion schedule}
\begin{figure}[t]
	\centering
	\includegraphics[width=0.3\linewidth]{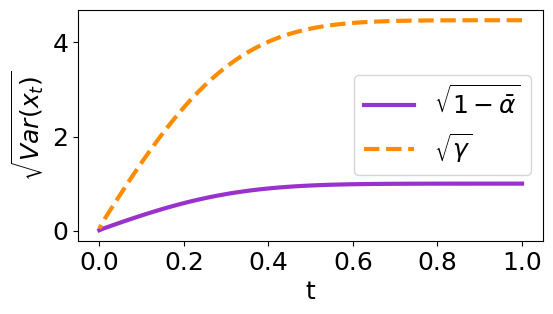}
	\caption{Schedules of standard deviation of added noise in the diffusion process~\eqref{eq:org_diffusion_noise_model},~\eqref{eq:ve_diffusion}.}
	\label{fig:diffusion_scheduler}
\end{figure}

Below we show the derivation of the diffusion schedule $\eta_t = \lambda \beta_t \Pi_{i=1}^{t-1} (1 - \beta_i)$ (Eq.~\eqref{eq:noise_schedule} in the main paper), that is used in our diffusion noise model (Eq.~\eqref{eq:ve_diffusion} in the main paper). We require that
\begin{equation}
    \gamma_t = \lambda \left( 1-\bar{\alpha}_t \right).
\end{equation}
If follows from the definition of $\gamma_t$ and $\bar{\alpha}_t$ that
\begin{equation}
    \gamma_t = \sum_{i=1}^{t} \eta_i = \lambda \left( 1 - \prod_{i=1}^t \left(1 - \beta_i \right) \right).
\end{equation}
This implies for $t=1,2$:
\begin{equation}
\begin{aligned}
    \eta_1 &= \lambda \left(1 - \left(1 - \beta_1\right) \right) = \lambda \beta_1, \\
    \eta_2 &= \lambda \left(1 - \left(1 - \beta_1\right) \left(1 - \beta_2\right) \right) - \eta_1 = \lambda \beta_2  \left(1 - \beta_1\right).
\end{aligned}
\end{equation}
For $t > 2$ we can derive the formula for $\eta_t$ by observing that $\eta_t = \gamma_t - \gamma_{t - 1}$, thus
\begin{equation}
\begin{aligned}
    \eta_t &= \gamma_t - \gamma_{t-1} =  \lambda \left( 1 - \prod_{i=1}^t \left(1 - \beta_i \right) \right) -  \lambda \left( 1 - \prod_{i=1}^{t-1} \left(1 - \beta_i \right) \right) \\
    &= \lambda \left( \prod_{i=1}^{t-1} \left(1 - \beta_i \right) - \prod_{i=1}^{t} \left(1 - \beta_i \right)  \right) =
    \lambda \cdot \left(\prod_{i=1}^{t-1} \left(1 - \beta_i \right)\right)\cdot \left(1  - \left(1-\beta_t\right)  \right) \\
    &= \lambda \beta_t \prod_{i=1}^{t-1} \left(1 - \beta_i \right).
\end{aligned}
\end{equation}
The diffusion noise schedules for both the standard diffusion and the non-stationary diffusion are depicted in Fig.~\ref{fig:diffusion_scheduler}.

\subsection{Reverse process}
In this section we show the derivation of the reverse process as appears in Eq.~\eqref{eq:ve_diffusion} in \secref{sec:diffusion_noise_model} of the main paper. 
For small enough schedule steps $\eta_t$ and given $\x$, the reverse process probability is a Gaussian of the form:
\begin{equation}
    q \left( \xt{t-1} \vert \xt{t}, \x \right) = \mathcal{N} \left( \xt{t-1}; \tilde{\boldsymbol{\mu}} \left( \xt{t}, \x \right), \tilde{\eta_t} \mathbf{I} \right).
\end{equation}
Using Bayes' rule,
\begin{equation}
\begin{aligned}
q \left( \xt{t-1} \vert \xt{t}, \x \right) 
&= q\left(\xt{t} \vert \xt{t-1}, \x\right) \frac{ q \left( \xt{t-1} \vert \x \right) }{ q \left( \xt{t} \vert \x \right) } \\
&\underset{(a)}{\propto} \exp \left(-\frac{1}{2} \left(
\frac{\left(\xt{t} - \xt{t-1}\right)^2}{\eta_t} + \frac{\left(\xt{t-1} - \x \right)^2}{\gamma_{t-1}} - \frac{\left(\xt{t} - \x \right)^2}{\gamma_t} \right) 
\right) \\
&= \exp \left(-\frac{1}{2} \left(
\frac{\xt{t}^2 - 2 \xt{t} \xt{t-1} + \xt{t-1}^2}{\eta_t} + 
\frac{\xt{t-1}^2 - 2 \x \xt{t-1} + \x^2}{\gamma_{t-1}} - 
\frac{\xt{t}^2 - 2 \x \xt{t} + \x^2}{\gamma_{t}} \right) 
\right) \\
&\underset{(b)}{=} \exp \left( -\frac{1}{2} \left( 
\left( \frac{1}{\eta_t} + \frac{1}{\gamma_{t-1}}\right) \xt{t-1}^2 - 
2\left( \frac{\xt{t}}{\eta_t} + \frac{\x}{\gamma_{t-1}} \right) \xt{t-1} + 
\mathbf{f} \left(\xt{t}, \x \right) \right) \right) \\
& \underset{(c)}{\triangleq} \exp  \left( -\frac{1}{2} \left( \frac{\left( \xt{t-1} - \tilde{\boldsymbol{\mu}}_t \right)^2}{\tilde{\eta}_t } \right) \right),
\end{aligned}
\end{equation}
where in (a) we use the forward definition in Eq.~\eqref{eq:ve_diffusion} in the main paper, in (b) we rearrange the expression as a polynomial of $\xt{t-1}$ and define the free coefficient $\mathbf{f} \left(\xt{t}, \x \right) \vcentcolon= \left( \frac{\xt{t}}{\eta_t} + \frac{\x}{\gamma_{t-1}} \right)^2 / \left( \frac{1}{\eta_t} + \frac{1}{\gamma_{t-1}}\right)$, and in (c) we denote,
\begin{equation}
\begin{aligned}
\tilde{\eta}_t 
&= \frac{1}{\left( \frac{1}{\eta_t} + \frac{1}{\gamma_{t-1}}\right)} = \frac{\gamma_{t-1} \eta_t }{\gamma_{t-1} + \eta_t} = \frac{\gamma_{t-1} \eta_t }{\gamma_{t}}, \\
\tilde{\boldsymbol{\mu}}_t \left( \xt{t}, \x \right)
&= \frac{\left( \frac{\xt{t}}{\eta_t} + \frac{\x}{\gamma_{t-1}} \right)}{\left( \frac{1}{\eta_t} + \frac{1}{\gamma_{t-1}}\right)} = \frac{\gamma_{t-1} \xt{t} + \eta_t \x}{\gamma_{t-1} + \eta_t} = \frac{\gamma_{t-1} \xt{t} + \eta_t \x}{\gamma_{t}} = \frac{\gamma_{t-1}}{\gamma_{t}} \xt{t} + \frac{\eta_t}{\gamma_{t}}  \x.
\end{aligned}
\end{equation}
The derivation of the loss function can be done by following the same approach as in~\cite{ho2020denoising}, with the only difference being our different $\mu_t$ and $\eta_t$.

\subsection{Noise correlation}
In the following section we prove the noise correlation relationship described in Eq.~\eqref{eq:train_ddpm} in \secref{sec:noise_correlation} of the main paper. 
For clarity we summarize the background for this phenomenon here: 
We consider a noisy input image generated according to the noise model in Eq.~\eqref{eq:noise_model}, and calculate the induced time map $\tmap$.
When initializing the reverse process with this noisy input, after $k$ diffusion steps the time map is given by $\tk{k} = \clipzero{\tmap - k}$. 
We wish to prove that the noise in $\xt{\tk{k}}$ can be written as a linear combination between the noise in the input image $\n{\tmap}$ and a new i.i.d. noise term $\n{\tk{k}}$. I.e.,
\begin{equation}
\begin{aligned}
\label{eq:restate_train_ddpm}
\xt{\tk{k}} &= \x + \frac{\gammamat{\tk{k}}}{\sqrt{\gammamat{\tmap}}} \n{\tmap} + \sqrt{\gammamat{\tk{k}} \left( 1 - \frac{\gammamat{\tk{k}}}{\gammamat{\tmap}} \right)} \n{\tk{k}}. \\
\end{aligned}
\end{equation}
We show this by induction. For $k=0$, we ought to prove
\begin{equation}
\begin{aligned}
\xt{\tk{0}} &= \x + \frac{\gammamat{\tk{0}}}{\sqrt{\gammamat{\tmap}}} \n{\tmap} + \sqrt{\gammamat{\tk{0}} \left( 1 - \frac{\gammamat{\tk{0}}}{\gammamat{\tmap}} \right)} \n{\tk{0}}. 
\end{aligned}
\end{equation}
Since $\tk{0}=\tmap$, this reduces to showing
\begin{equation}
\begin{aligned}
\xt{\tmap} &= \x + \frac{\gammamat{\tmap}}{\sqrt{\gammamat{\tmap}}} \n{\tmap} + \sqrt{\gammamat{\tmap} \left( 1 - \frac{\gammamat{\tmap}}{\gammamat{\tmap}} \right)} \n{\tk{0}} = \x + \sqrt{\gammamat{\tmap}}\n{\tmap}
\end{aligned}
\end{equation}
which is true by definition.

Now, suppose that Eq.~\eqref{eq:restate_train_ddpm} holds for $\tk{k} = \clipzero{\tmap - k}$ with a new i.i.d. noise term $\n{\tk{k}}$.
The next reverse step is $\clipzero{\tk{k} - 1}$. For simplicity we omit the clipping notation.
By the reverse process equation (Eq.~\eqref{eq:ve_reverse} in the main paper) we can express $\xt{\tk{k}-1}$ as a function of $\xt{\tk{k}}, \x$ and a new i.i.d. noise term $\nbar{\tk{k}-1}$. We use an equivalent reformulation for equation Eq.~\eqref{eq:ve_reverse}, noting that 
\begin{equation}
    \tilde{\eta}_t = \frac{\gamma_{t-1} \eta_{t}}{\gamma_{t}} = \frac{\gamma_{t-1} \left(\gamma_{t} - \gamma_{t-1} \right)}{\gamma_{t}} = \gamma_{t-1} \left( 1 - \frac{\gamma_{t-1}}{\gamma_{t}} \right).
\end{equation}
Hence, we have
\begin{equation}
\label{eq:reverse_k-1}
\xt{\tk{k}\!-\!1} \! =\! \frac{\gammamat{\tk{k}\!-\!1}}{\gammamat{\tk{k}}} \xt{\tk{k}} + \frac{\etamat{\tk{k}}}{\gammamat{\tk{k}}} \x + \sqrt{\gammamat{\tk{k}\!-\!1} \left(1 - \frac{\gammamat{\tk{k}\!-\!1}}{\gammamat{\tk{k}}} \right) } \nbar{\tk{k}\!-\!1}.
\end{equation}
Plugging Eq.~\eqref{eq:restate_train_ddpm} into Eq.~\eqref{eq:reverse_k-1},
\begin{equation}
\begin{aligned}
    \xt{\tk{k}\!-\!1} \! &=\! \frac{\gammamat{\tk{k}\!-\!1}}{\gammamat{\tk{k}}} \left[ \x + \frac{\gammamat{\tk{k}}}{\sqrt{\gammamat{\tmap}}} \n{\tmap} + \sqrt{\gammamat{\tk{k}} \!\! \left(\! 1 \!-\! \frac{\gammamat{\tk{k}}}{\gammamat{\tmap}} \! \right)} \n{\tk{k}} \right] + \frac{\etamat{\tk{k}}}{\gammamat{\tk{k}}} \x + \sqrt{\gammamat{\tk{k}\!-\!1} \left(1 - \frac{\gammamat{\tk{k}\!-\!1}}{\gammamat{\tk{k}}} \right) } \nbar{\tk{k}\!-\!1} \\
    &\underset{(a)}{=} \! \x + \underset{\text{``Source noise"}}{\underbrace{\frac{\gammamat{\tk{k}-1}}{\sqrt{\gammamat{\tmap}}}\n{\tmap}}} + \underset{\text{``Uncorrelated noise"}}
    {\underbrace{\frac{\gammamat{\tk{k}\!-\!1}}{\gammamat{\tk{k}}} \sqrt{\gammamat{\tk{k}} \!\! \left(\! 1 \!-\! \frac{\gammamat{\tk{k}}}{\gammamat{\tmap}} \! \right)} \n{\tk{k}} + \sqrt{\gammamat{\tk{k}\!-\!1} \left(1 - \frac{\gammamat{\tk{k}\!-\!1}}{\gammamat{\tk{k}}} \right) } \nbar{\tk{k}\!-\!1}}} \\
    &\underset{(b)}{=} \! \x + \underset{\text{``Source noise"}}{\underbrace{\frac{\gammamat{\tk{k}-1}}{\sqrt{\gammamat{\tmap}}}\n{\tmap}}} + 
    \underset{\text{``Uncorrelated noise"}}{\underbrace{\sqrt{\left( \frac{\gammamat{\tk{k}\!-\!1}}{\gammamat{\tk{k}}} \right)^2 \gammamat{\tk{k}} \!\! \left(\! 1 \!-\! \frac{\gammamat{\tk{k}}}{\gammamat{\tmap}} \! \right) + \gammamat{\tk{k}\!-\!1} \left(1 - \frac{\gammamat{\tk{k}\!-\!1}}{\gammamat{\tk{k}}} \right)}}} \n{\tk{k}\!-\!1} \\
    &= \! \x + \underset{\text{``Source noise"}}{\underbrace{\frac{\gammamat{\tk{k}-1}}{\sqrt{\gammamat{\tmap}}}\n{\tmap}}} + 
    \underset{\text{``Uncorrelated noise"}}{\underbrace{\sqrt{\gammamat{\tk{k}\!-\!1} \left[  \left( \frac{\gammamat{\tk{k}\!-\!1}}{\gammamat{\tk{k}}} - \frac{\gammamat{\tk{k}\!-\!1}}{\gammamat{\tmap}}\right) + \left(1 - \frac{\gammamat{\tk{k}\!-\!1}}{\gammamat{\tk{k}}} \right) \right]}}} \n{\tk{k}\!-\!1} \\
    &= \! \x + \underset{\text{``Source noise"}}{\underbrace{\frac{\gammamat{\tk{k}-1}}{\sqrt{\gammamat{\tmap}}}\n{\tmap}}} + 
    \underset{\text{``Uncorrelated noise"}}{\underbrace{\sqrt{\gammamat{\tk{k}\!-\!1} \left( 1 - \frac{\gammamat{\tk{k}\!-\!1}}{\gammamat{\tmap}}\right)}}} \n{\tk{k}\!-\!1} \\
\end{aligned}
\end{equation}
where in (a) we rearrange the noise term to seperate between the noise of the input image and the noise terms that are uncorrelated to it and in (b) we use the property of summation of independent Gaussian variables and introduced a new i.i.d. noise term $\n{\tk{k}\!-\!1}$. 

Finally, we wish to express $\xt{\tk{k}\!-\!1}$ with a single noise term, which will be used in the loss function. 
This is done again with summation of two independent variables,
\begin{equation}
\begin{aligned}
    \xt{\tk{k}\!-\!1} &= \! \x + \underset{\text{``Source noise"}}{\underbrace{\frac{\gammamat{\tk{k}-1}}{\sqrt{\gammamat{\tmap}}}\n{\tmap}}} + 
    \underset{\text{``Uncorrelated noise"}}{\underbrace{\sqrt{\gammamat{\tk{k}\!-\!1} \left( 1 - \frac{\gammamat{\tk{k}\!-\!1}}{\gammamat{\tmap}}\right)}}} \n{\tk{k}\!-\!1} \\
     &= \! \x + \sqrt{\frac{\gammamat{\tk{k}-1}^2}{\gammamat{\tmap}} + \gammamat{\tk{k}\!-\!1} \left( 1 - \frac{\gammamat{\tk{k}\!-\!1}}{\gammamat{\tmap}}\right)} \ntilde{\tk{k}\!-\!1} \\
     &= \! \x + \sqrt{\gammamat{\tk{k}\!-\!1}} \ntilde{\tk{k}\!-\!1}, \\
\end{aligned}
\end{equation}
and the expression for $\ntilde{\tk{k}\!-\!1}$ is given by division of the noise terms by $\sqrt{\gammamat{\tk{k}\!-\!1}}$,
\begin{equation}
    \ntilde{\tk{k}\!-\!1} = \underset{\text{``Source noise"}}{\underbrace{\sqrt{\frac{\gammamat{\tk{k}-1}}{\gammamat{\tmap}}}\n{\tmap}}} + 
    \underset{\text{``Uncorrelated noise"}}{\underbrace{\sqrt{ 1 - \frac{\gammamat{\tk{k}\!-\!1}}{\gammamat{\tmap}}}\n{\tk{k}\!-\!1}}}.
\end{equation}

\section{Results}
We show a comparison of our method with the baseline diffusion model and a state-of-the-art denoising network~\cite{chen2021hinet}, on images from ImageNet deteriorated by our noise model~\eqref{eq:noise_model}, under various noise levels. We show results from three noise levels, corresponding to camera gain levels of 1, 4, and 20 (recall that the results from gain level 16 were presented in \figref{fig:imagenet_comparison-full-page-1} and \figref{fig:imagenet_comparison-full-page-2} in the main paper).

\begin{figure*}[htbp]
\setlength{\ww}{0.192\textwidth}
\begin{center}
\newcommand{\magn}{5.0}
\newcommand{\spyloc}{(0.28,0.15)}    
\newcommand{\spyshift}{(1.675,-3.4)}
\newcommand{\bright}{0.1 1 0.1 1 0.1 1}
\small\addtolength{\tabcolsep}{-8.5pt}
\begin{tabular}{ccccc}
    Noisy & HINet~\cite{chen2021hinet} & Baseline & Ours & Clean GT\\
        %% socks 990
        \renewcommand{\spyloc}{(0.2,1)}
        \begin{tikzpicture}[spy using outlines={red,magnification=\magn,size=\ww}]
            \node {\includegraphics[width=\ww]{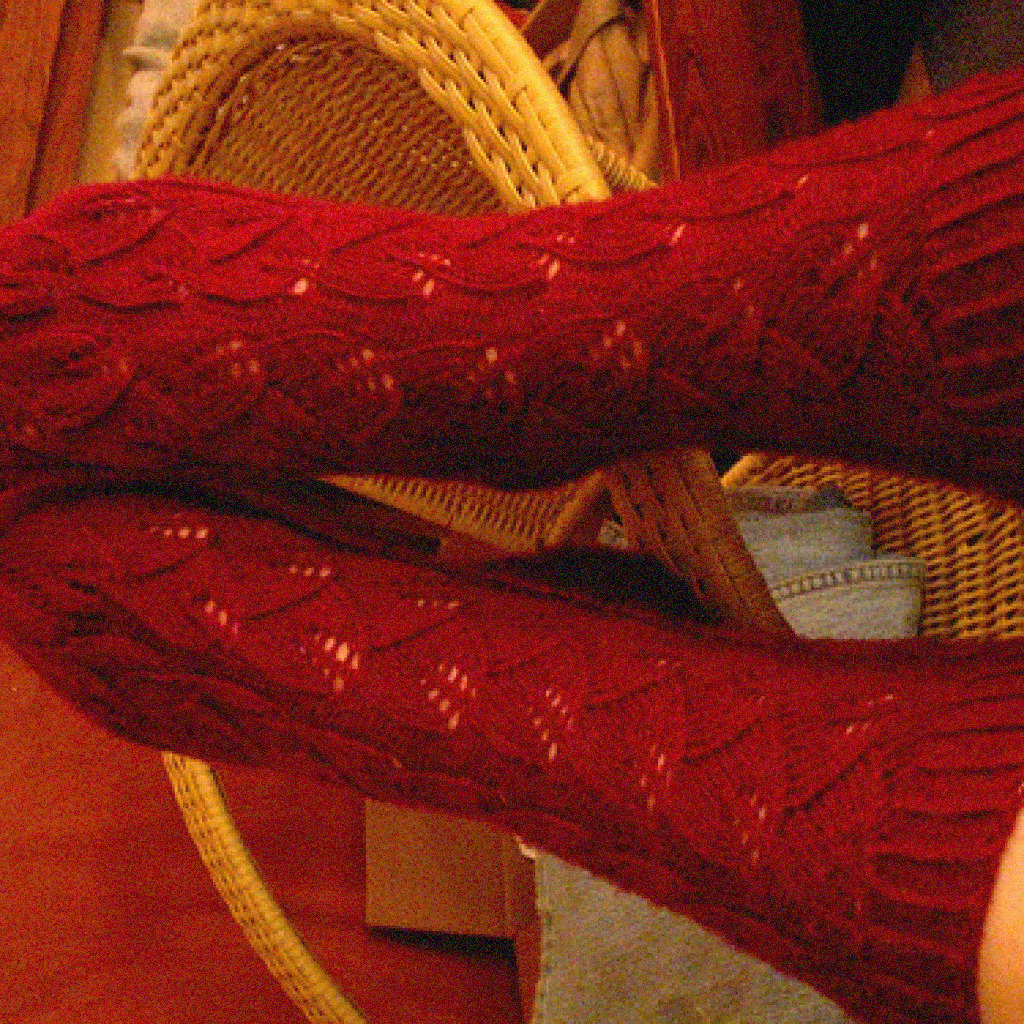}};
            \spy on \spyloc in node [left] at \spyshift;
        \end{tikzpicture} &
        \renewcommand{\spyloc}{(0.2,1)}
        \begin{tikzpicture}[spy using outlines={red,magnification=\magn,size=\ww}]
            \node {\includegraphics[width=\ww]{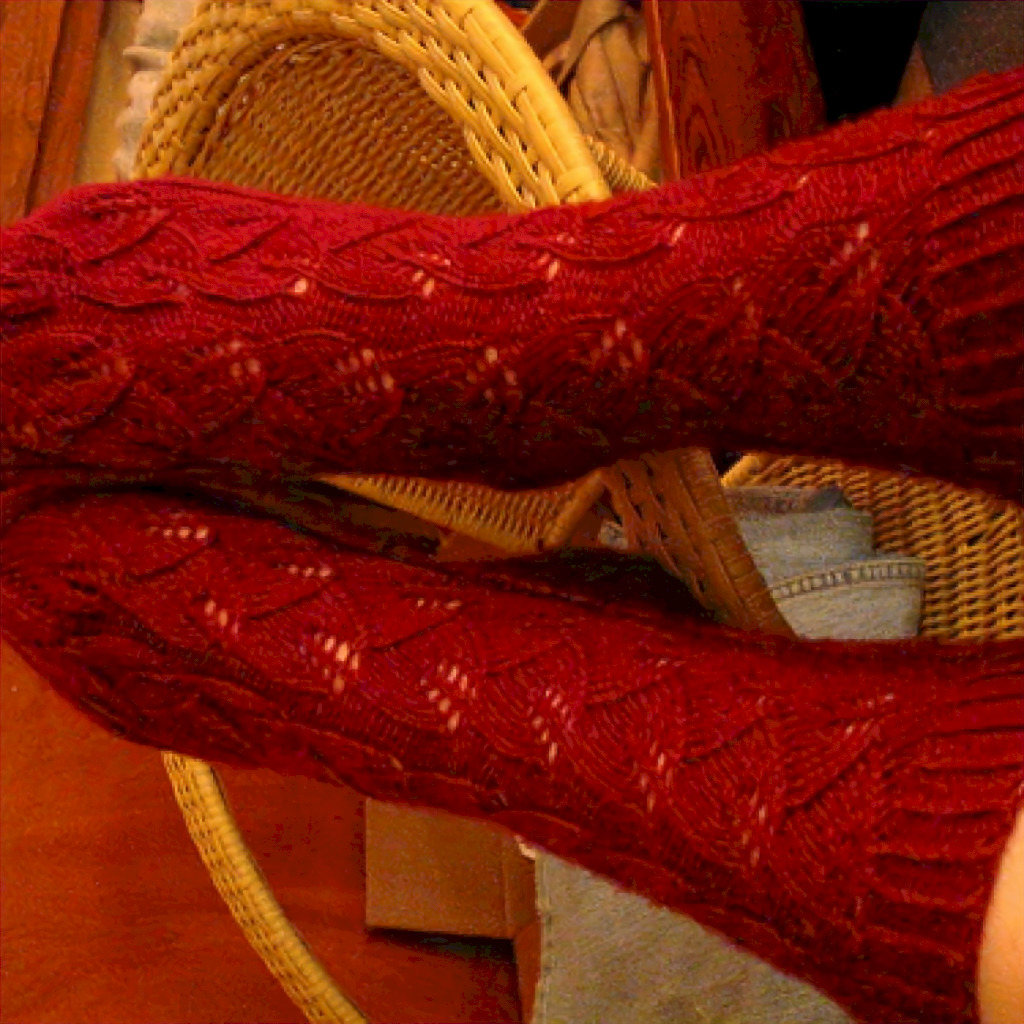}};
            \spy on \spyloc in node [left] at \spyshift;
        \end{tikzpicture} &
        \renewcommand{\spyloc}{(0.2,1)}
        \begin{tikzpicture}[spy using outlines={red,magnification=\magn,size=\ww}]
            \node {\includegraphics[width=\ww]{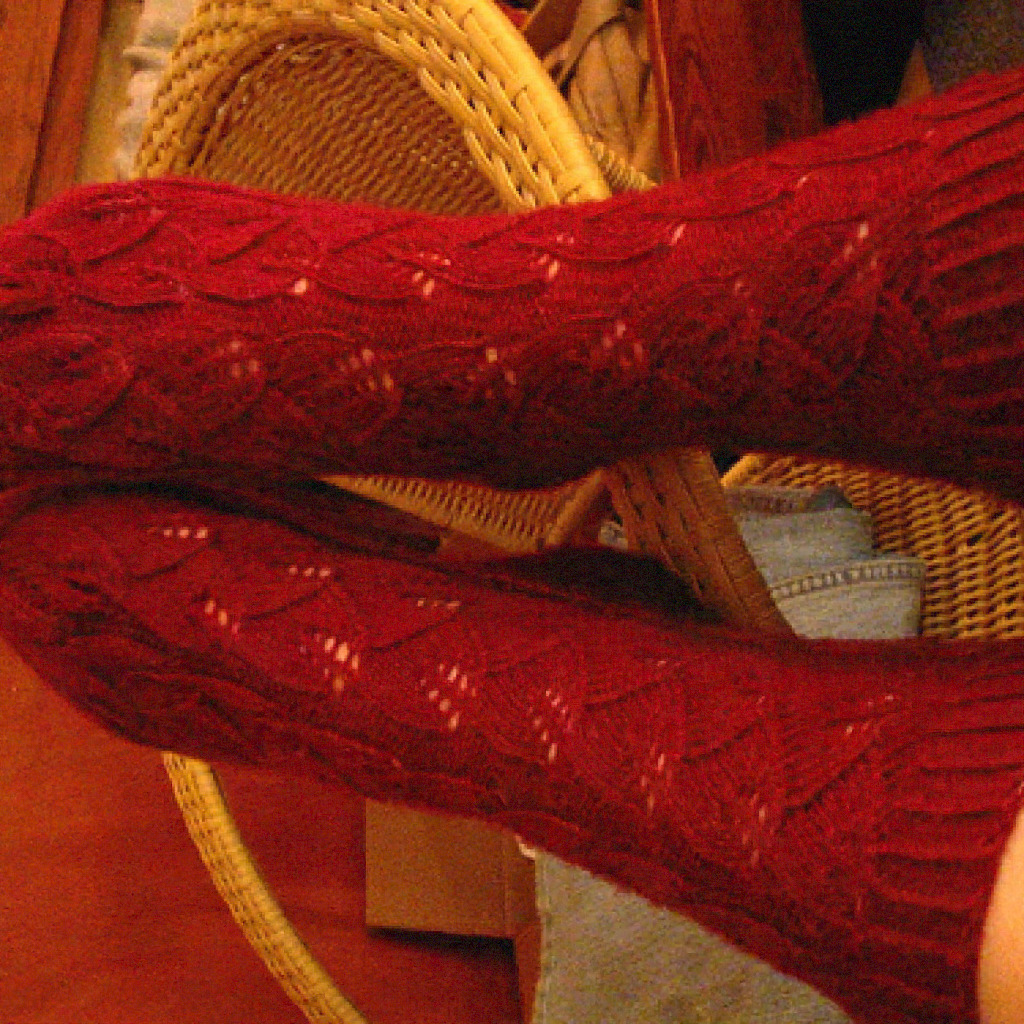}};
            \spy on \spyloc in node [left] at \spyshift;
        \end{tikzpicture} &
        \renewcommand{\spyloc}{(0.2,1)}
        \begin{tikzpicture}[spy using outlines={red,magnification=\magn,size=\ww}]
            \node {\includegraphics[width=\ww]{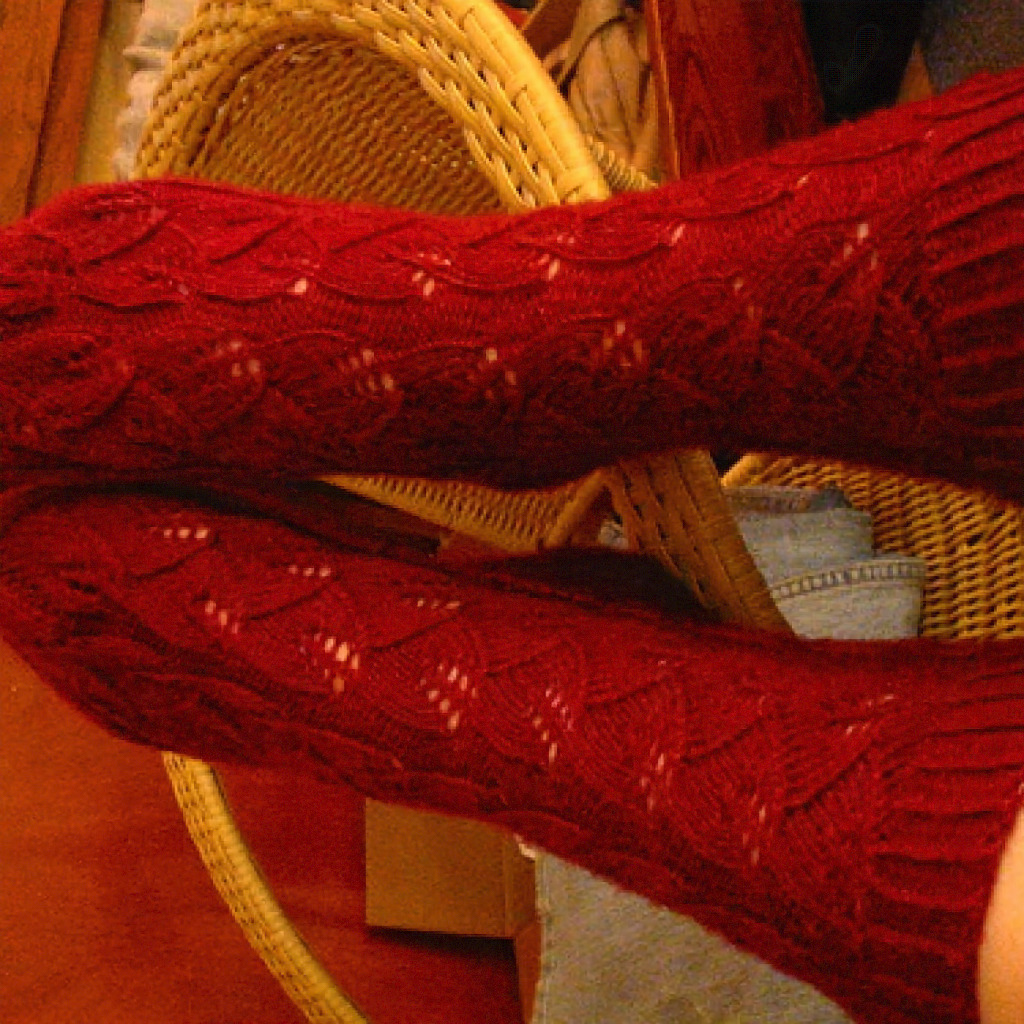}};
            \spy on \spyloc in node [left] at \spyshift;
        \end{tikzpicture} &
        \renewcommand{\spyloc}{(0.2,1)}
        \begin{tikzpicture}[spy using outlines={red,magnification=\magn,size=\ww}]
            \node {\includegraphics[width=\ww]{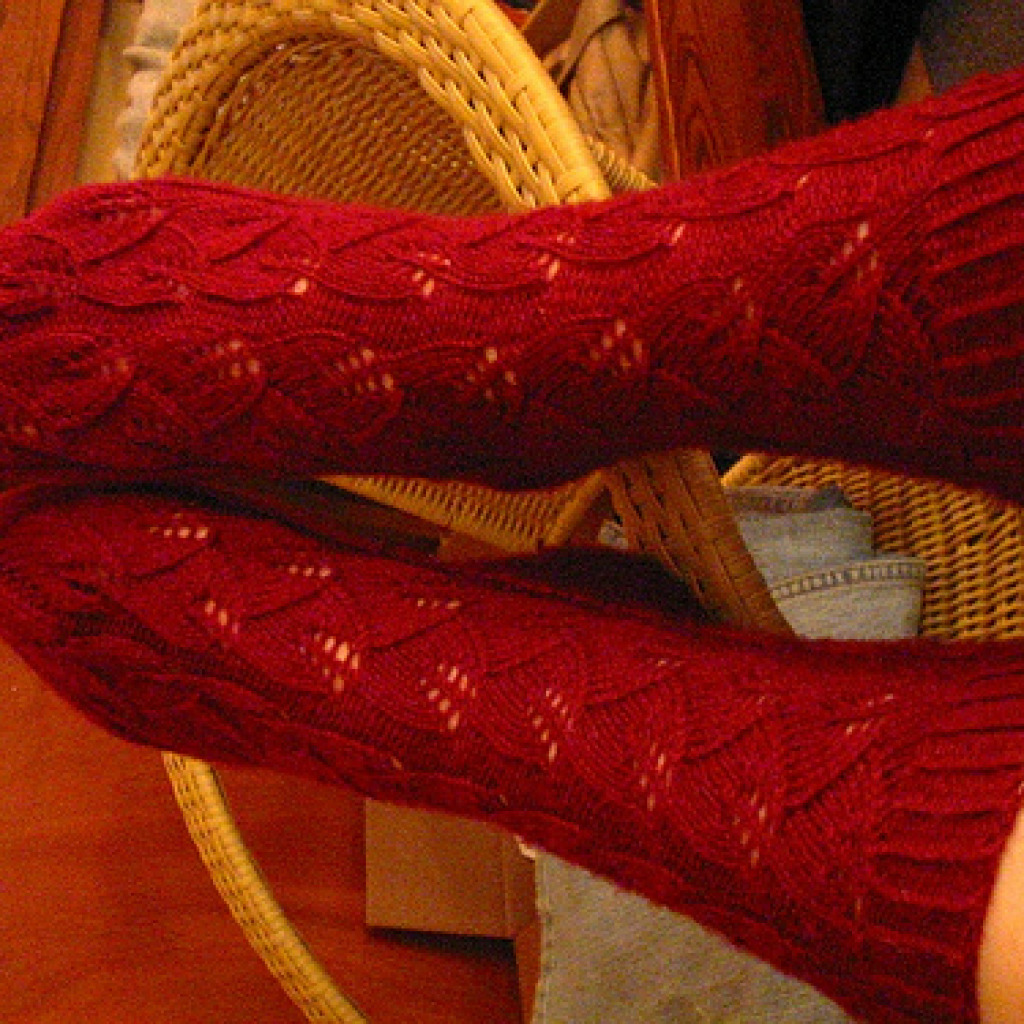}};
            \spy on \spyloc in node [left] at \spyshift;
        \end{tikzpicture} \\

        %% owl 957
        \renewcommand{\spyloc}{(-0.2,-1)}
        \begin{tikzpicture}[spy using outlines={red,magnification=\magn,size=\ww}]
            \node {\includegraphics[width=\ww, decodearray=\bright]{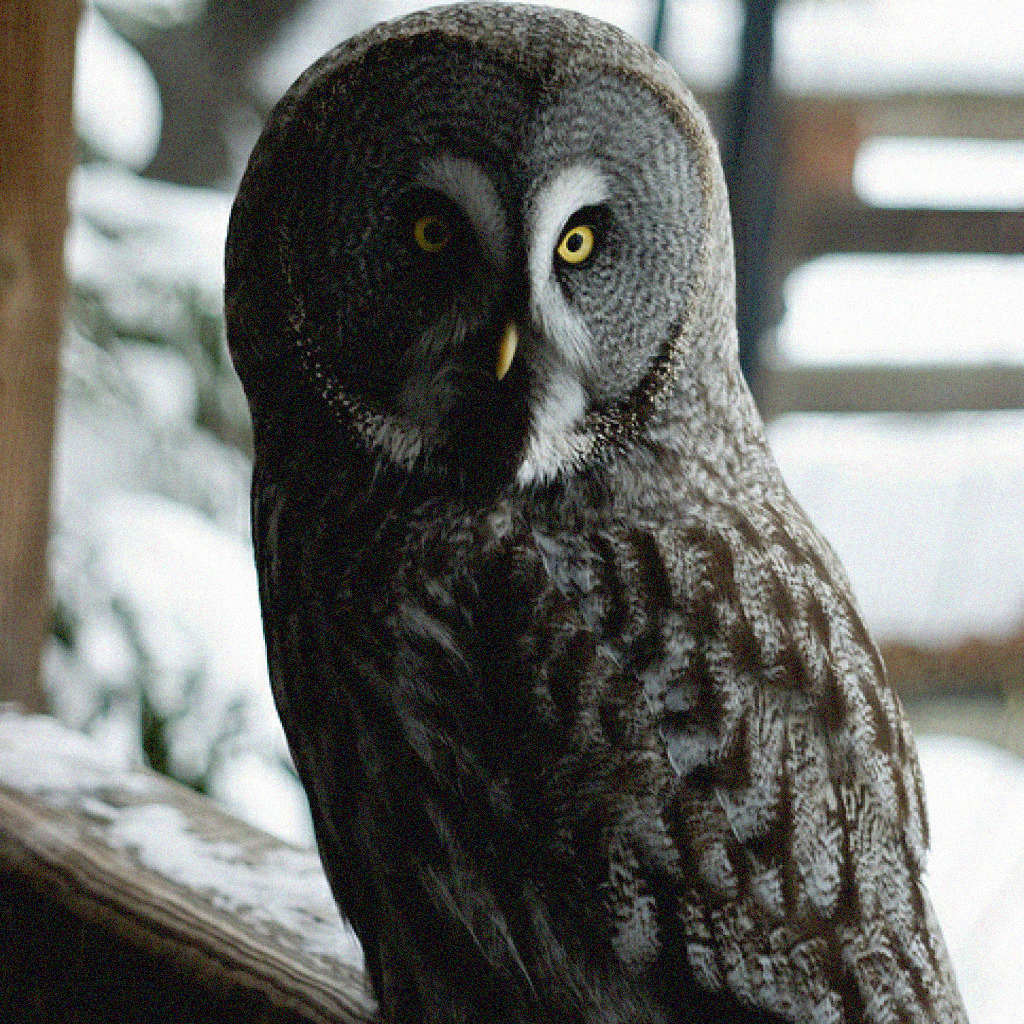}};
            \spy on \spyloc in node [left] at \spyshift;
        \end{tikzpicture} &
        \renewcommand{\spyloc}{(-0.2,-1)}
        \begin{tikzpicture}[spy using outlines={red,magnification=\magn,size=\ww}]
            \node {\includegraphics[width=\ww, decodearray=\bright]{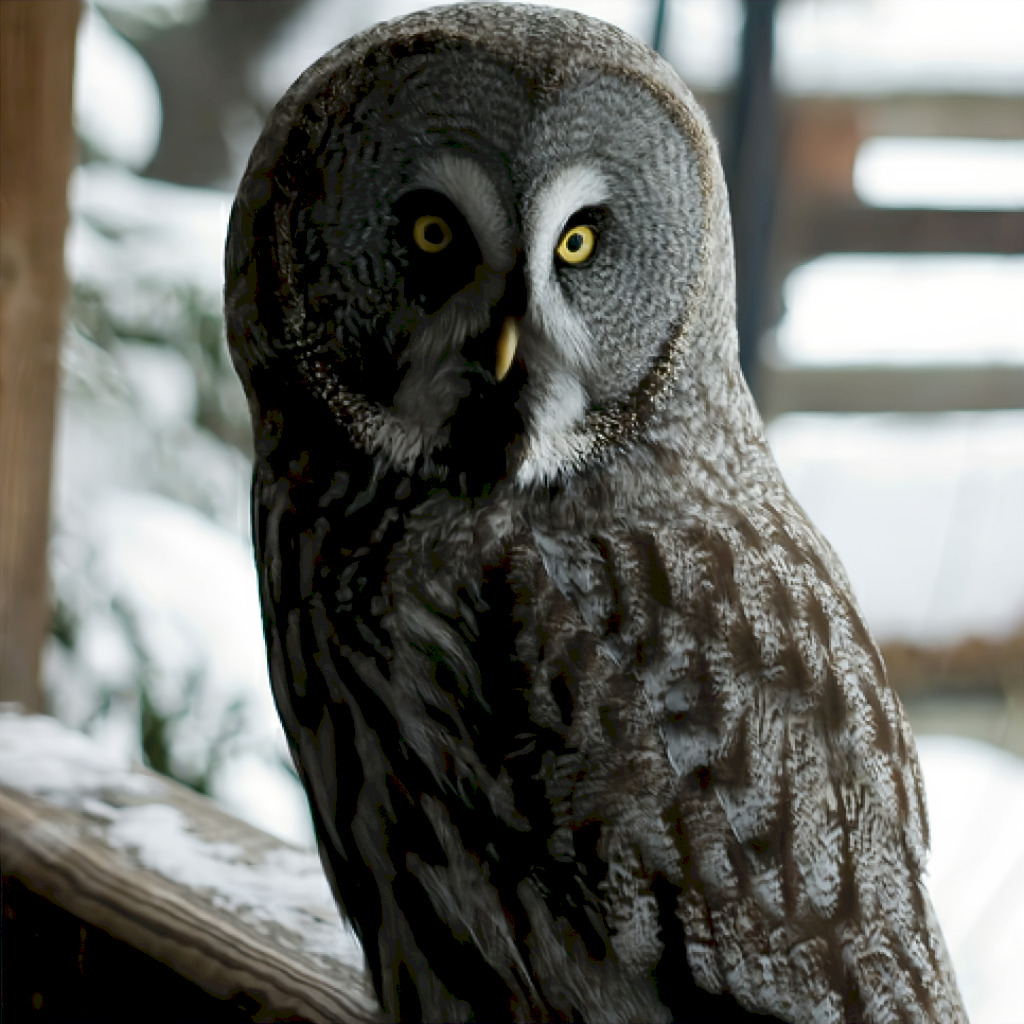}};
            \spy on \spyloc in node [left] at \spyshift;
        \end{tikzpicture} &
        \renewcommand{\spyloc}{(-0.2,-1)}
        \begin{tikzpicture}[spy using outlines={red,magnification=\magn,size=\ww}]
            \node {\includegraphics[width=\ww, decodearray=\bright]{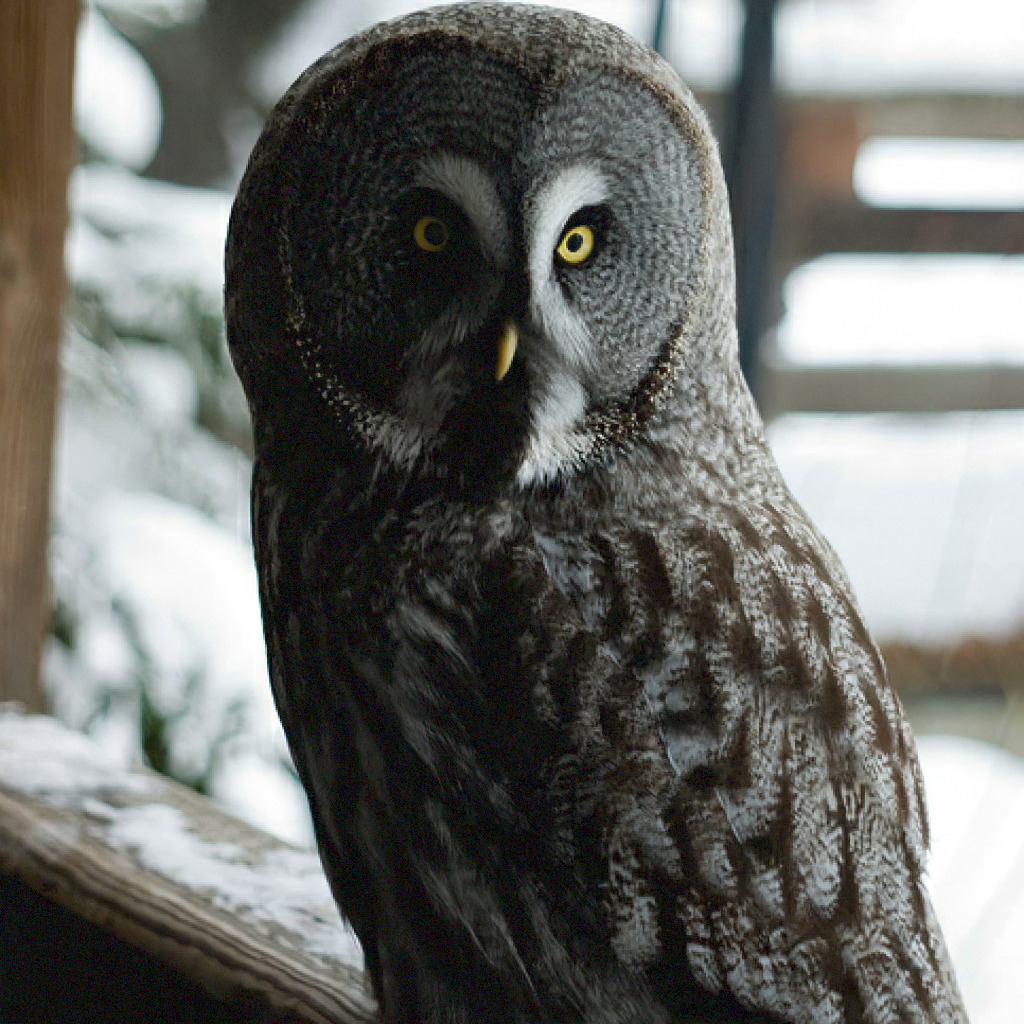}};
            \spy on \spyloc in node [left] at \spyshift;
        \end{tikzpicture} &
        \renewcommand{\spyloc}{(-0.2,-1)}
        \begin{tikzpicture}[spy using outlines={red,magnification=\magn,size=\ww}]
            \node {\includegraphics[width=\ww, decodearray=\bright]{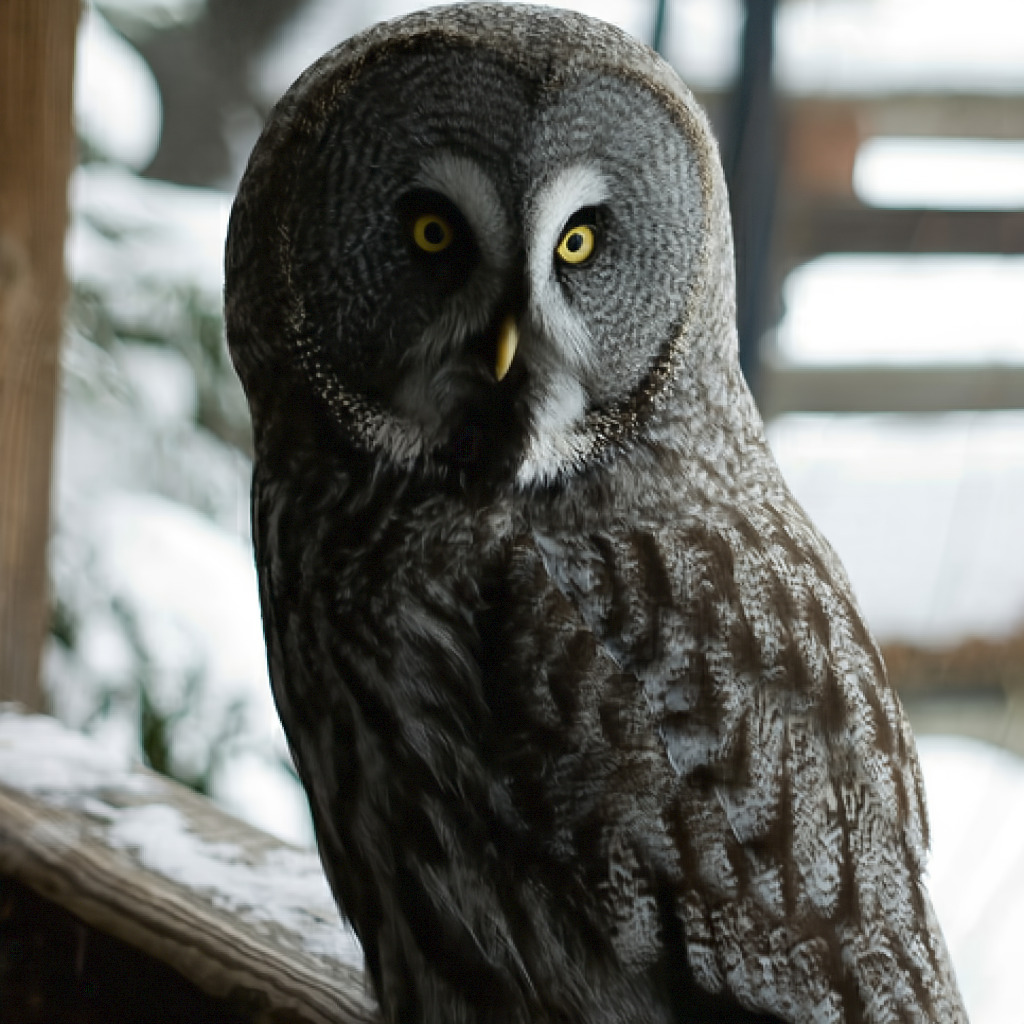}};
            \spy on \spyloc in node [left] at \spyshift;
        \end{tikzpicture} &
        \renewcommand{\spyloc}{(-0.2,-1)}
        \begin{tikzpicture}[spy using outlines={red,magnification=\magn,size=\ww}]
            \node {\includegraphics[width=\ww, decodearray=\bright]{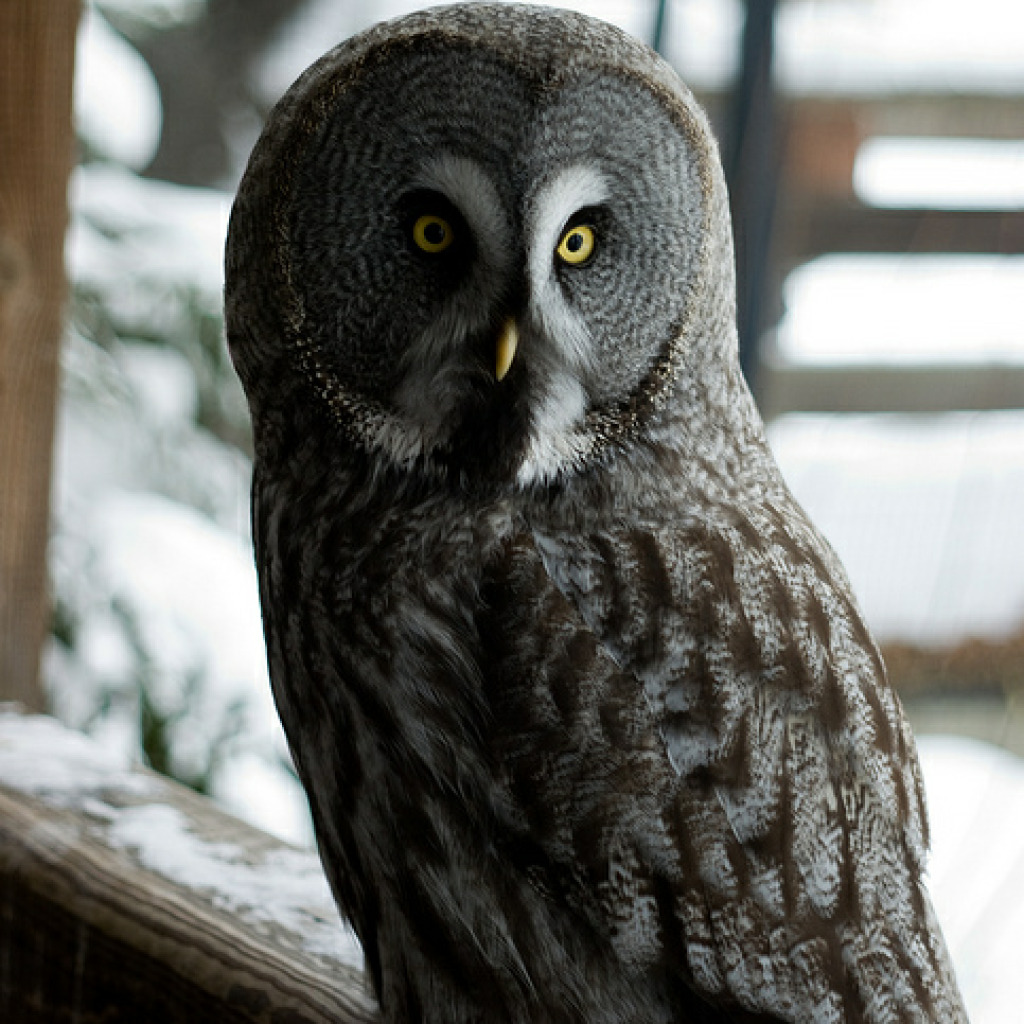}};
            \spy on \spyloc in node [left] at \spyshift;
        \end{tikzpicture} \\

        %% duck 949
        \renewcommand{\spyloc}{(-0.8, 0.6)}
        \begin{tikzpicture}[spy using outlines={red,magnification=\magn,size=\ww}]
            \node {\includegraphics[width=\ww]{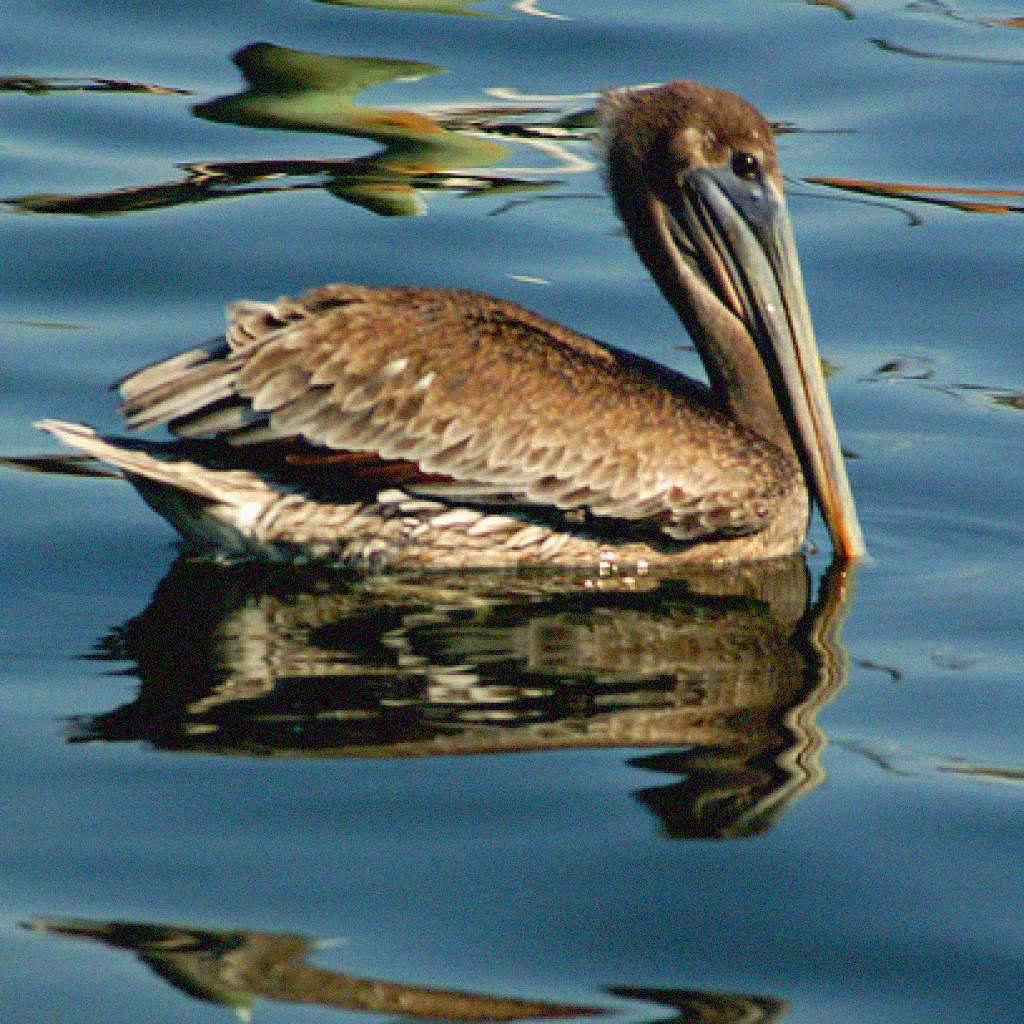}};
            \spy on \spyloc in node [left] at \spyshift;
        \end{tikzpicture} &
        \renewcommand{\spyloc}{(-0.8, 0.6)}
        \begin{tikzpicture}[spy using outlines={red,magnification=\magn,size=\ww}]
            \node {\includegraphics[width=\ww]{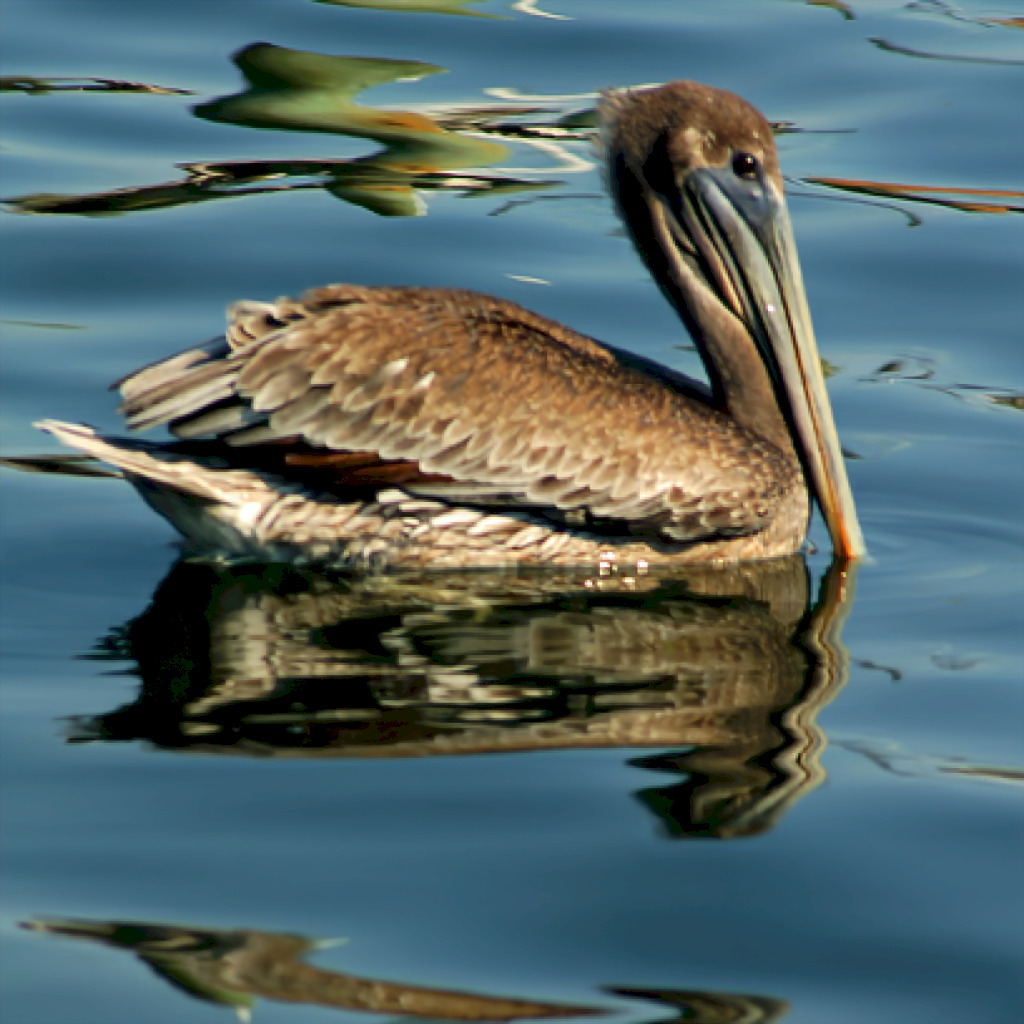}};
            \spy on \spyloc in node [left] at \spyshift;
        \end{tikzpicture} &
        \renewcommand{\spyloc}{(-0.8, 0.6)}
        \begin{tikzpicture}[spy using outlines={red,magnification=\magn,size=\ww}]
            \node {\includegraphics[width=\ww]{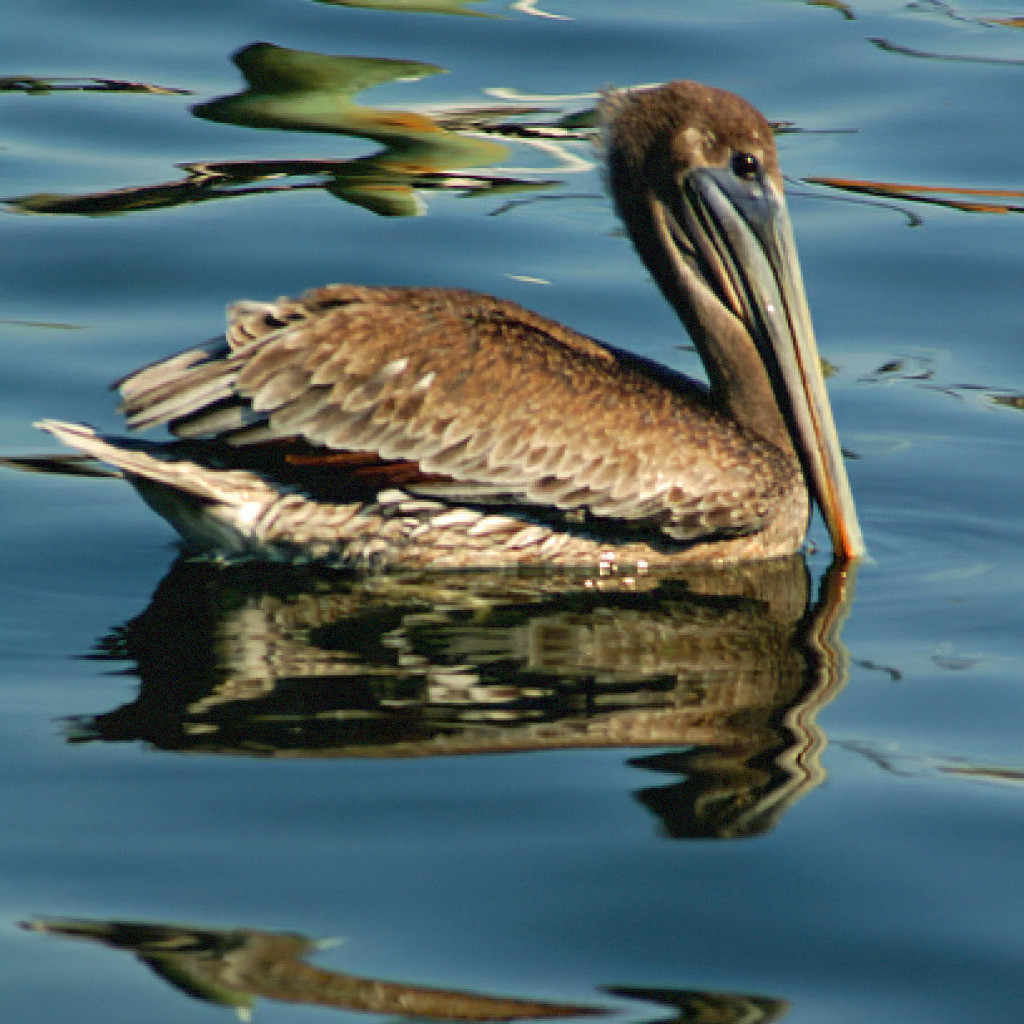}};
            \spy on \spyloc in node [left] at \spyshift;
        \end{tikzpicture} &
        \renewcommand{\spyloc}{(-0.8, 0.6)}
        \begin{tikzpicture}[spy using outlines={red,magnification=\magn,size=\ww}]
            \node {\includegraphics[width=\ww]{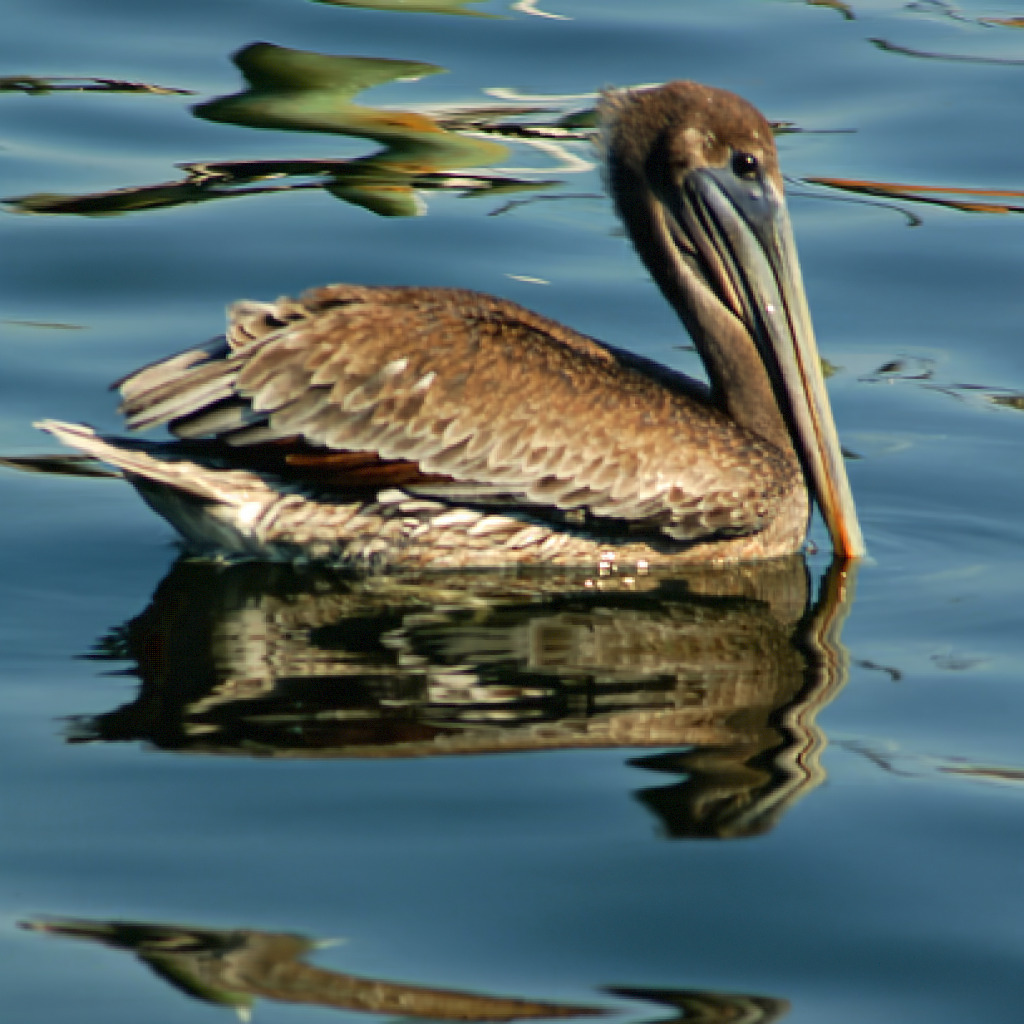}};
            \spy on \spyloc in node [left] at \spyshift;
        \end{tikzpicture} &
        \renewcommand{\spyloc}{(-0.8, 0.6)}
        \begin{tikzpicture}[spy using outlines={red,magnification=\magn,size=\ww}]
            \node {\includegraphics[width=\ww]{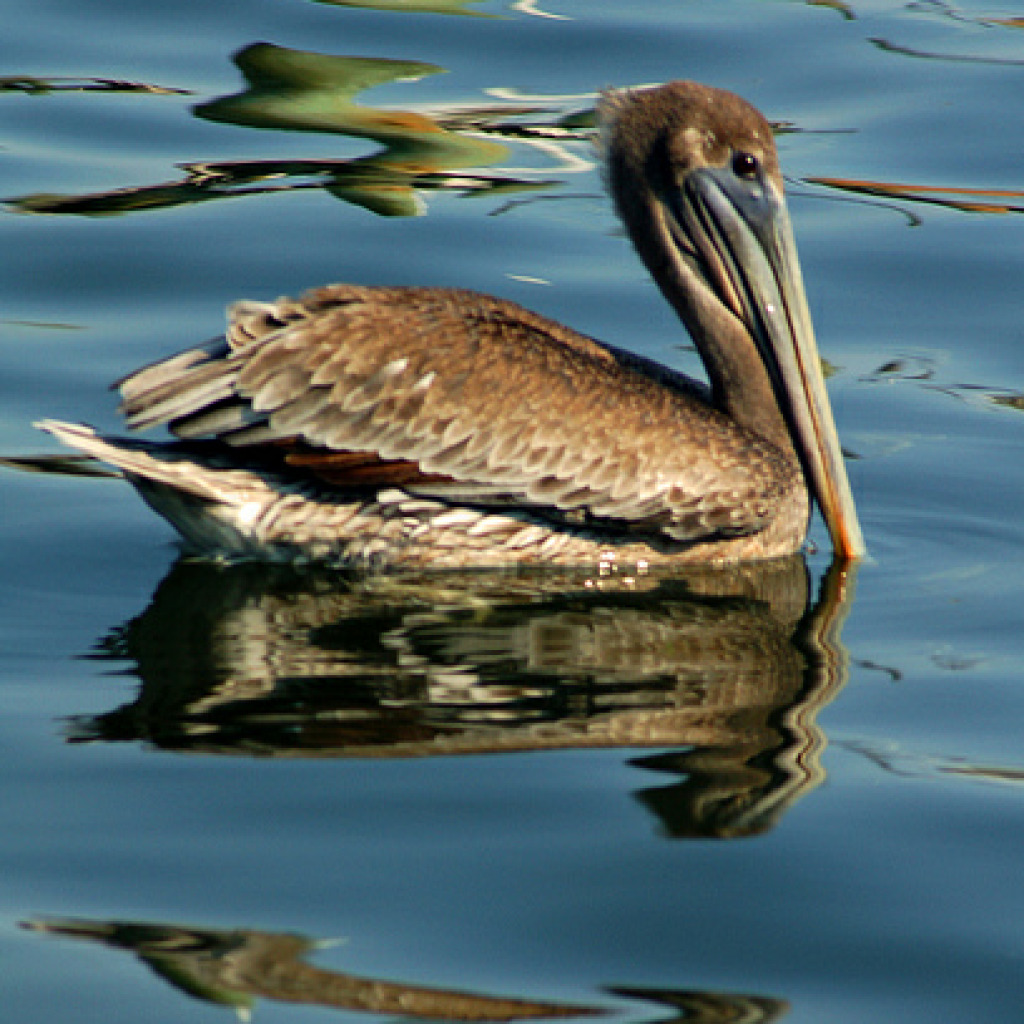}};
            \spy on \spyloc in node [left] at \spyshift;
        \end{tikzpicture} \\

\end{tabular}
\end{center}
\vspace{-1em}
\caption{Comparison between different denoising methods on images with noise gain of 1. Some images are brightened for visualization.}
%\vspace{-1em}
\label{fig:imagenet_comparison_1}
\end{figure*}

\begin{figure}[htbp]
\setlength{\ww}{0.192\textwidth}
\begin{center}
\newcommand{\magn}{5.0}
\newcommand{\spyloc}{(0.28,0.15)}    
\newcommand{\spyshift}{(1.675,-3.4)}
\newcommand{\bright}{0.1 1 0.1 1 0.1 1}
\small\addtolength{\tabcolsep}{-8.5pt}
\begin{tabular}{ccccc}
    Noisy & HINet~\cite{chen2021hinet} & Baseline & Ours & Clean GT\\
        %% bird 991
        \renewcommand{\spyloc}{(0.8,0.8)}
        \begin{tikzpicture}[spy using outlines={red,magnification=\magn,size=\ww}]
            \node {\includegraphics[width=\ww]{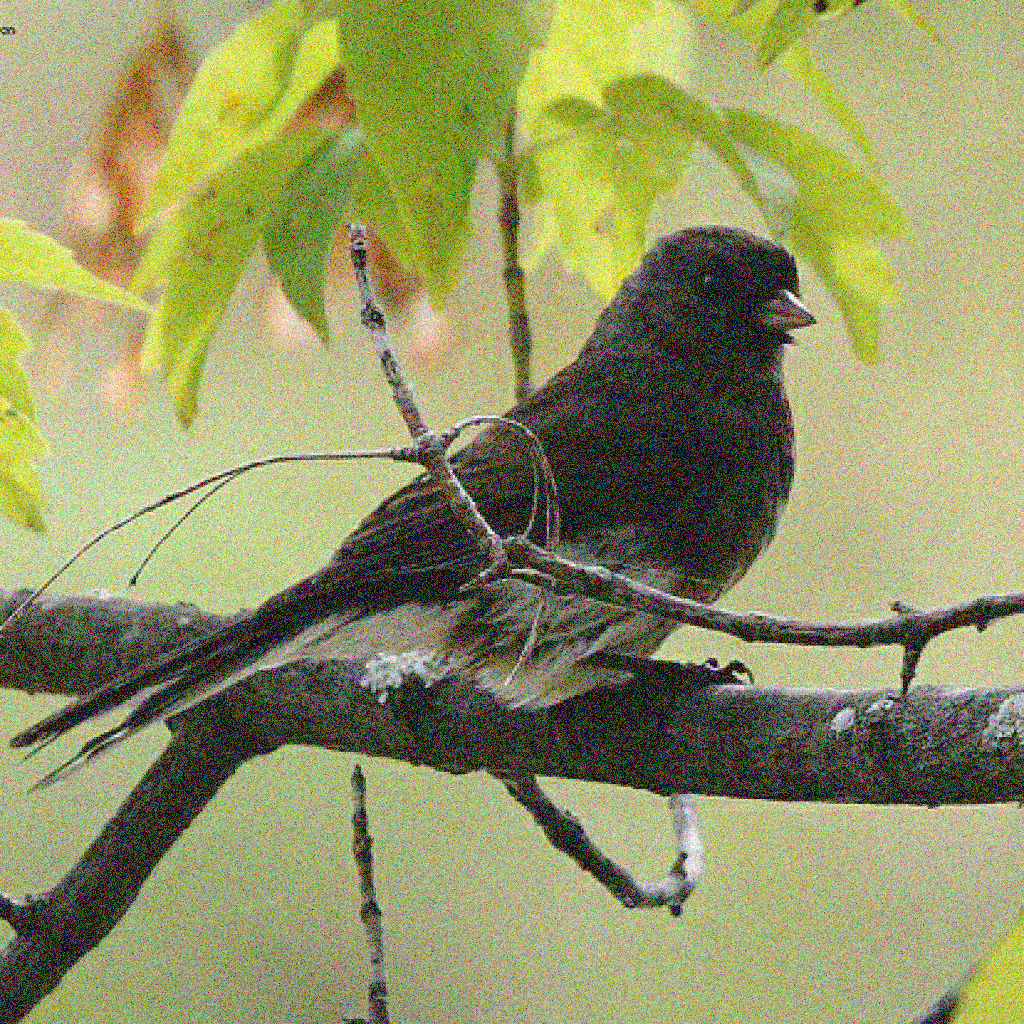}};
            \spy on \spyloc in node [left] at \spyshift;
        \end{tikzpicture} &
        \renewcommand{\spyloc}{(0.8,0.8)}
        \begin{tikzpicture}[spy using outlines={red,magnification=\magn,size=\ww}]
            \node {\includegraphics[width=\ww]{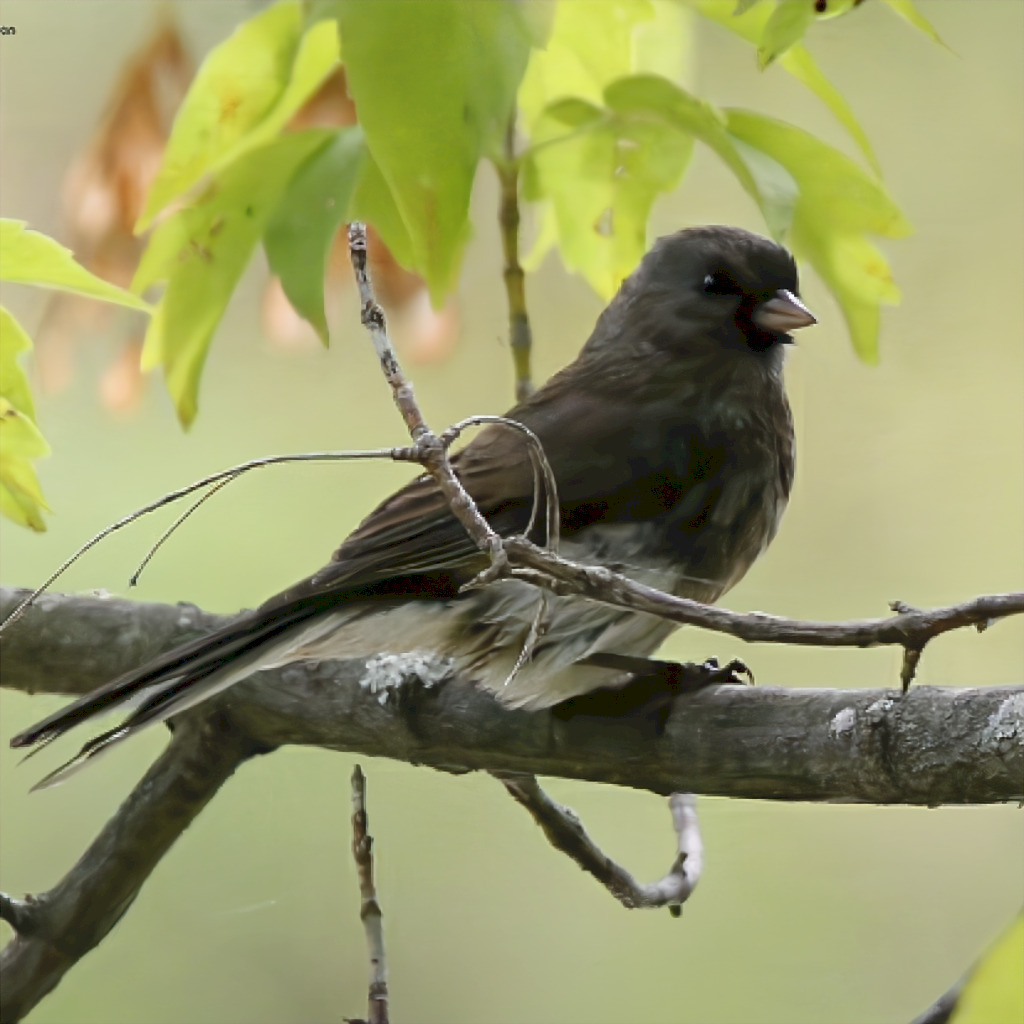}};
            \spy on \spyloc in node [left] at \spyshift;
        \end{tikzpicture} &
        \renewcommand{\spyloc}{(0.8,0.8)}
        \begin{tikzpicture}[spy using outlines={red,magnification=\magn,size=\ww}]
            \node {\includegraphics[width=\ww]{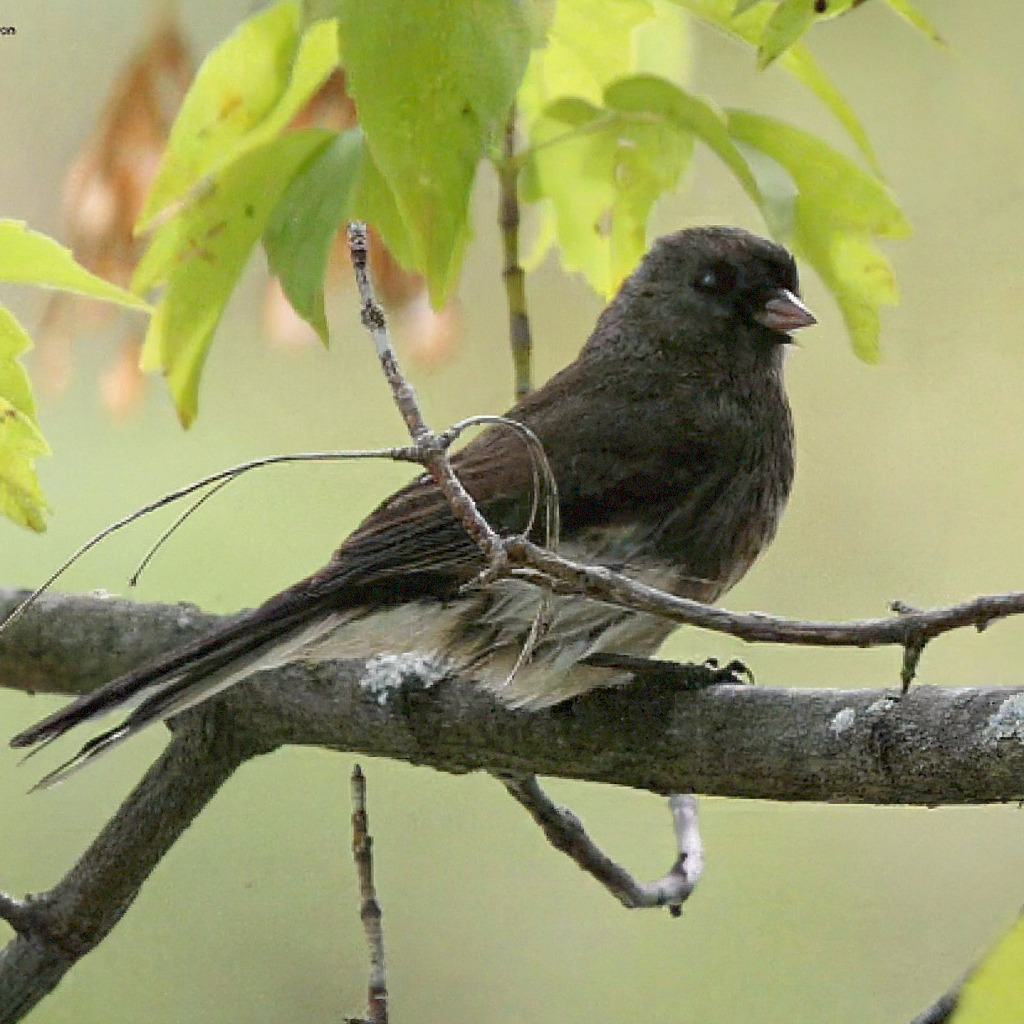}};
            \spy on \spyloc in node [left] at \spyshift;
        \end{tikzpicture} &
        \renewcommand{\spyloc}{(0.8,0.8)}
        \begin{tikzpicture}[spy using outlines={red,magnification=\magn,size=\ww}]
            \node {\includegraphics[width=\ww]{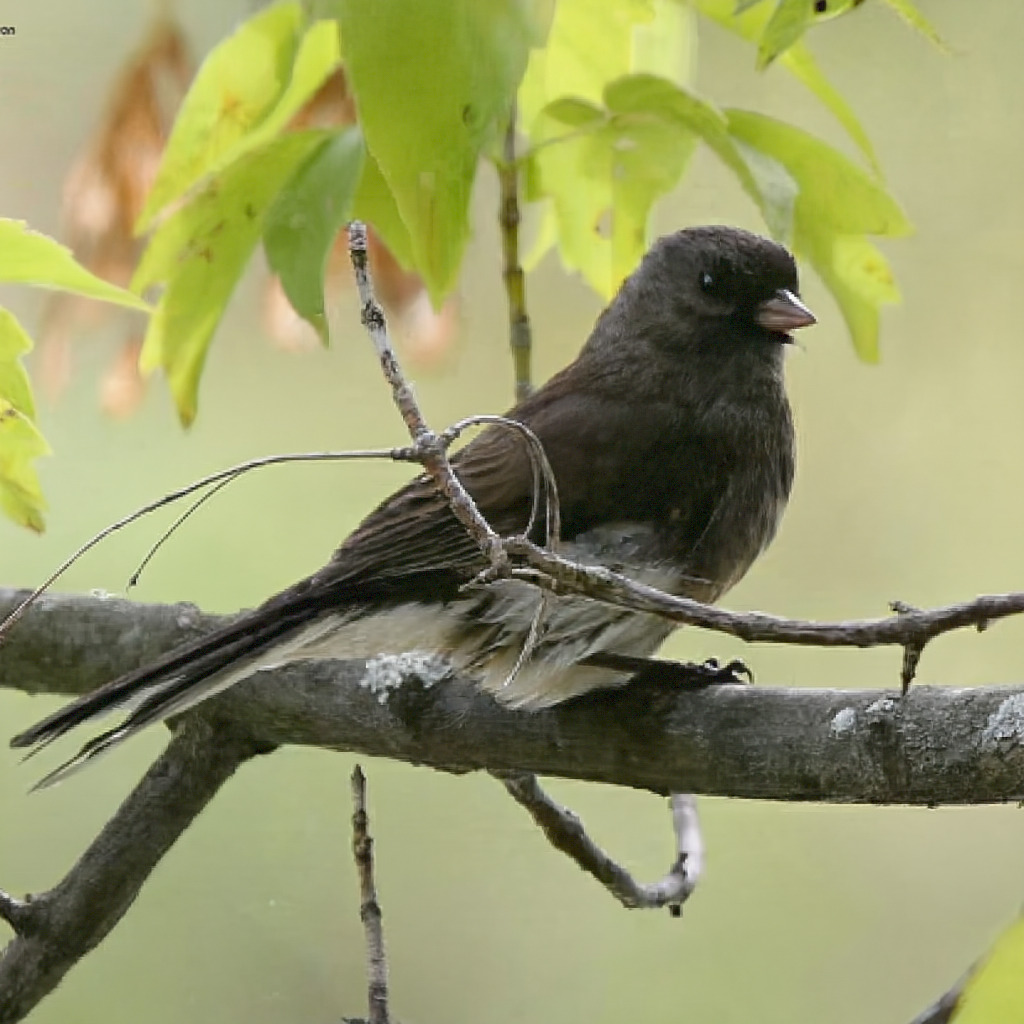}};
            \spy on \spyloc in node [left] at \spyshift;
        \end{tikzpicture} &
        \renewcommand{\spyloc}{(0.8,0.8)}
        \begin{tikzpicture}[spy using outlines={red,magnification=\magn,size=\ww}]
            \node {\includegraphics[width=\ww]{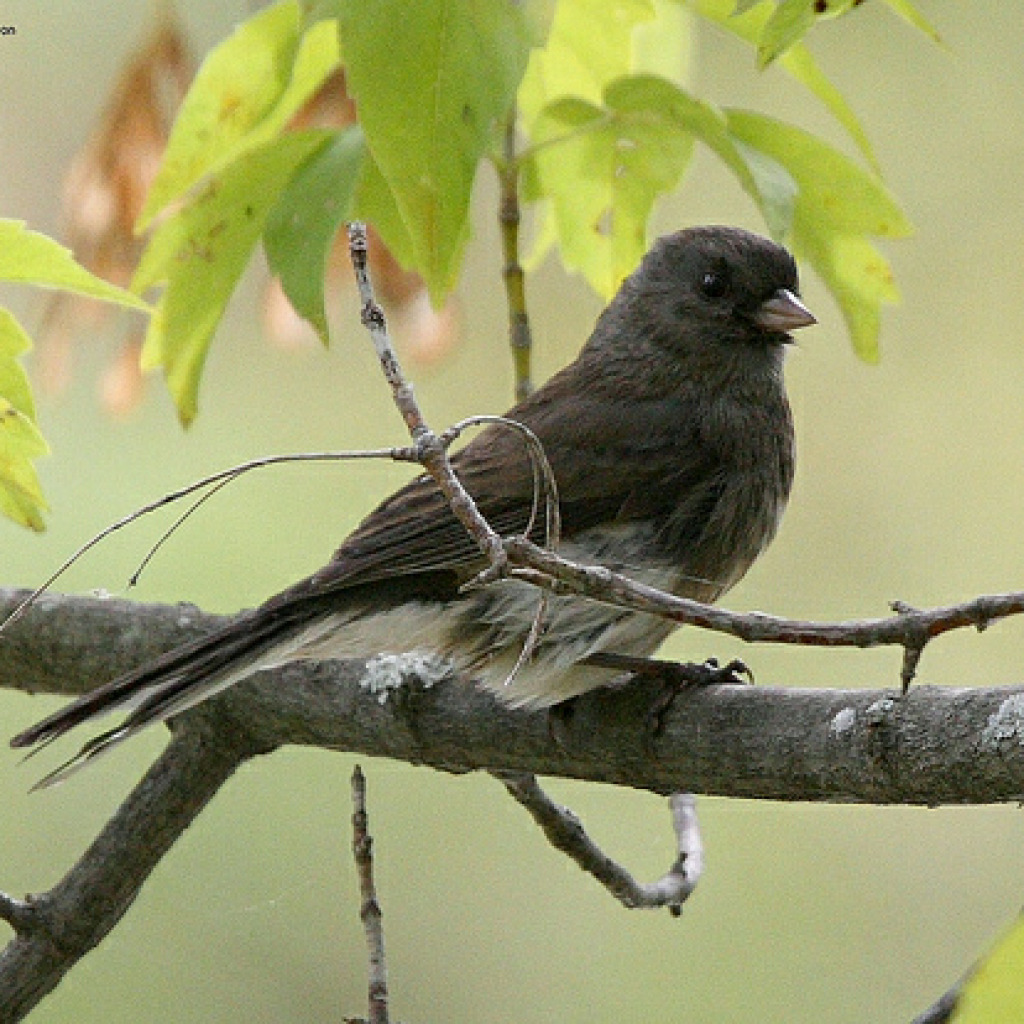}};
            \spy on \spyloc in node [left] at \spyshift;
        \end{tikzpicture} \\

        %% oak 942
        \renewcommand{\spyloc}{(-0.2,-1.2)}
        \begin{tikzpicture}[spy using outlines={red,magnification=\magn,size=\ww}]
            \node {\includegraphics[width=\ww]{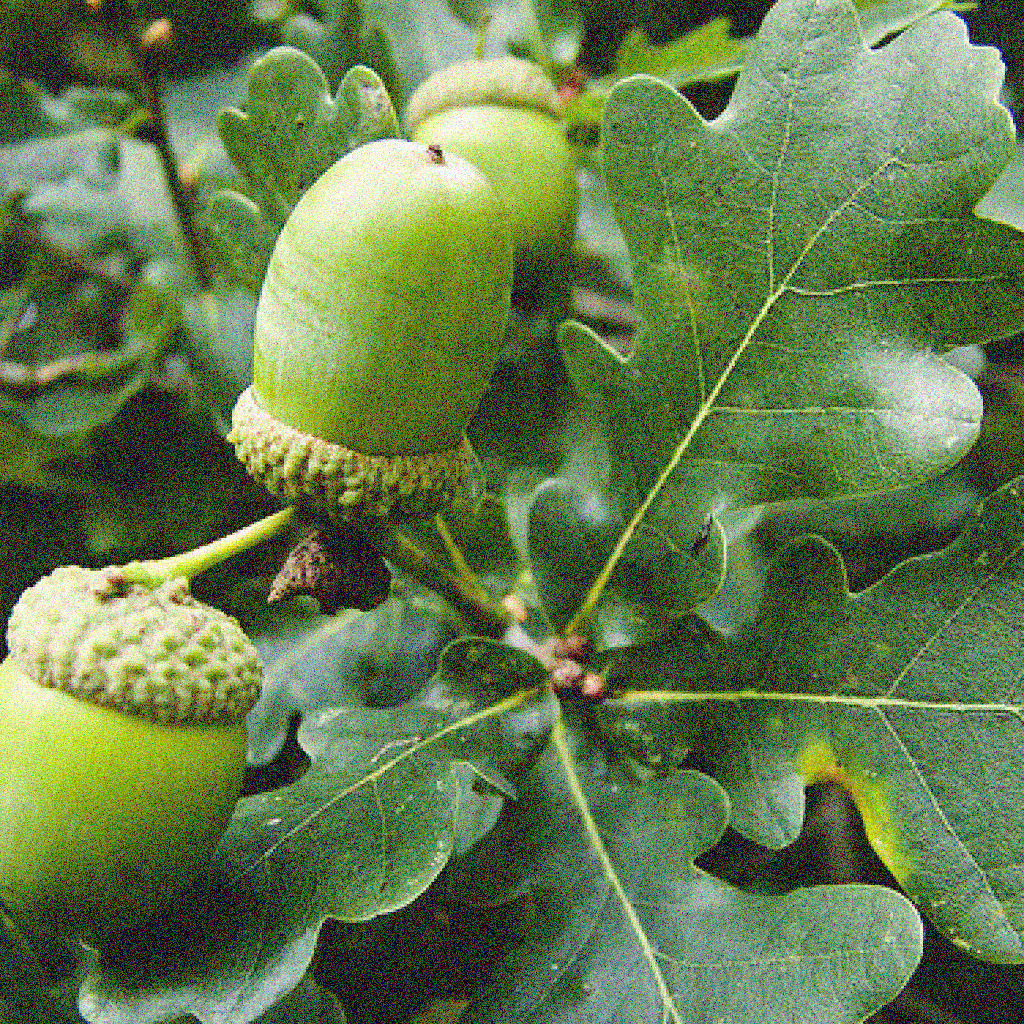}};
            \spy on \spyloc in node [left] at \spyshift;
        \end{tikzpicture} &
        \renewcommand{\spyloc}{(-0.2,-1.2)}
        \begin{tikzpicture}[spy using outlines={red,magnification=\magn,size=\ww}]
            \node {\includegraphics[width=\ww]{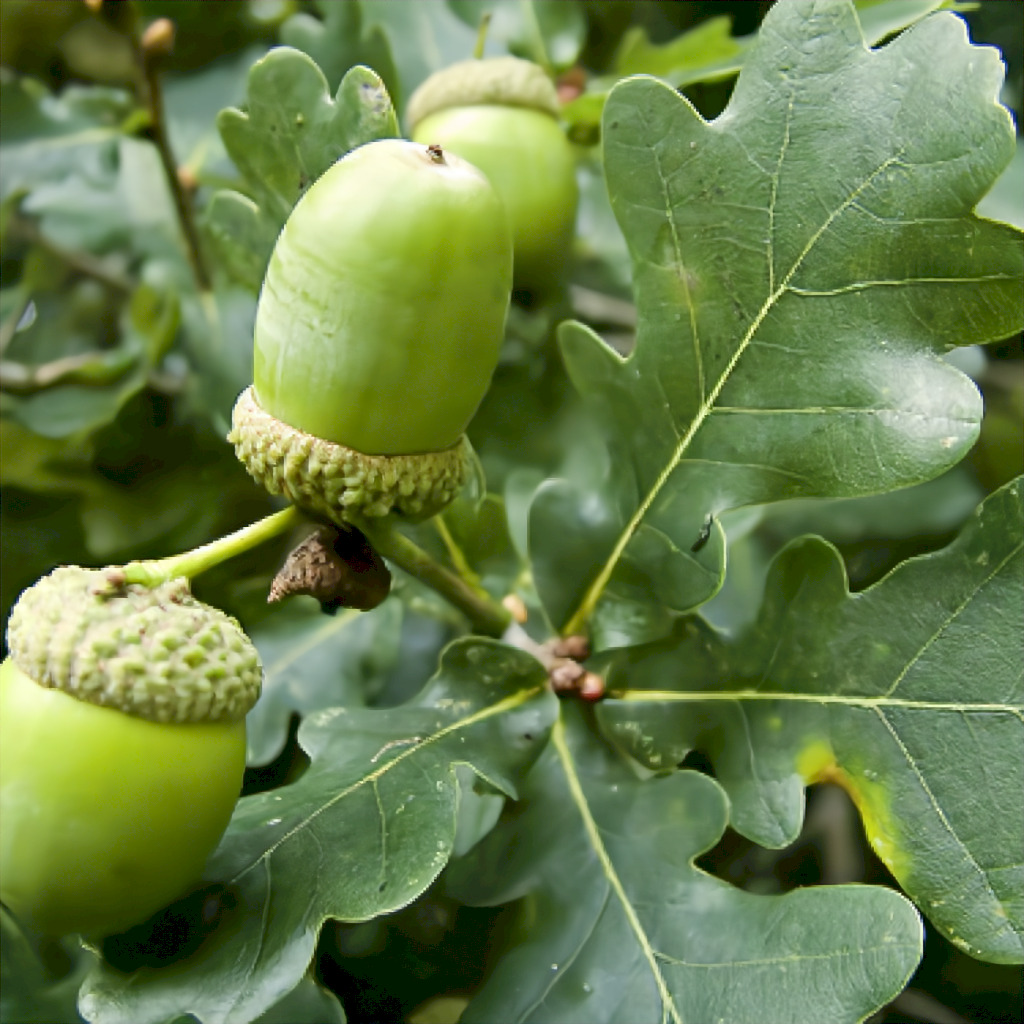}};
            \spy on \spyloc in node [left] at \spyshift;
        \end{tikzpicture} &
        \renewcommand{\spyloc}{(-0.2,-1.2)}
        \begin{tikzpicture}[spy using outlines={red,magnification=\magn,size=\ww}]
            \node {\includegraphics[width=\ww]{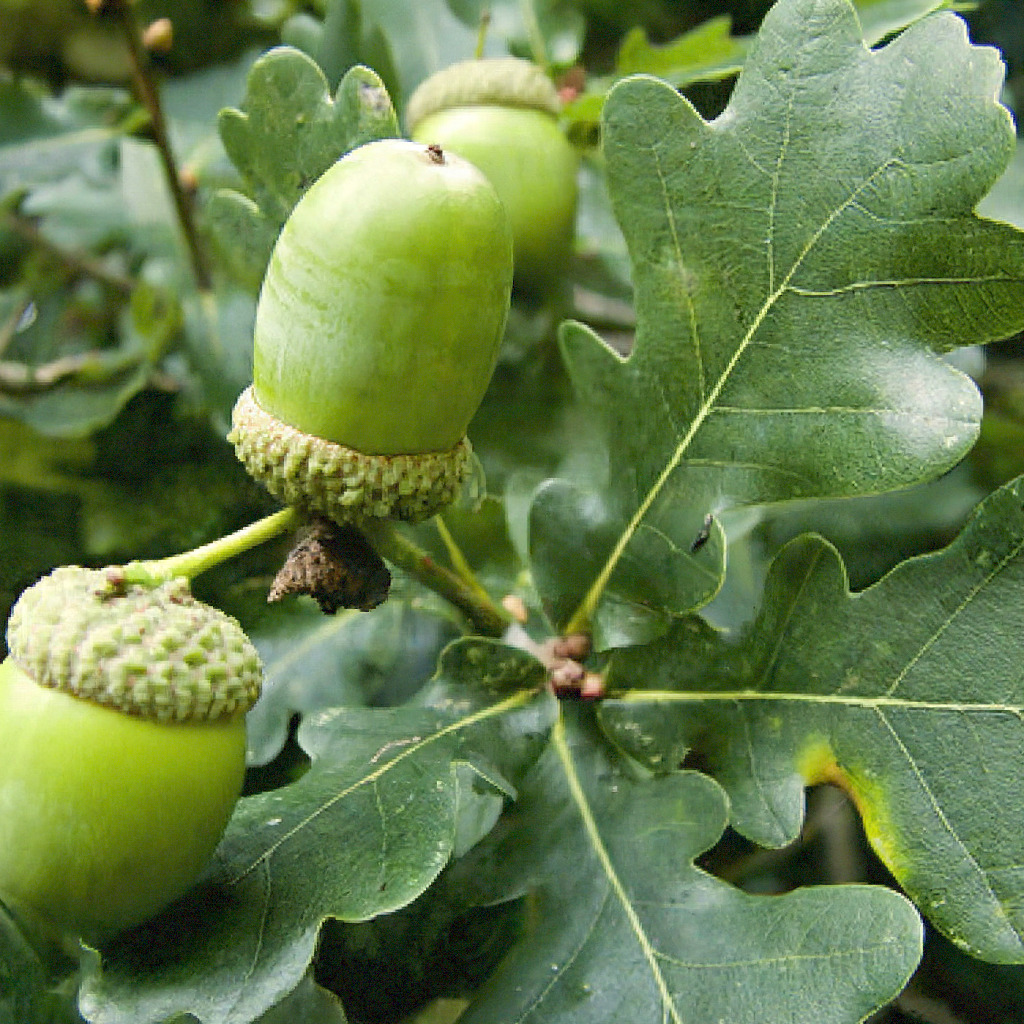}};
            \spy on \spyloc in node [left] at \spyshift;
        \end{tikzpicture} &
        \renewcommand{\spyloc}{(-0.2,-1.2)}
        \begin{tikzpicture}[spy using outlines={red,magnification=\magn,size=\ww}]
            \node {\includegraphics[width=\ww]{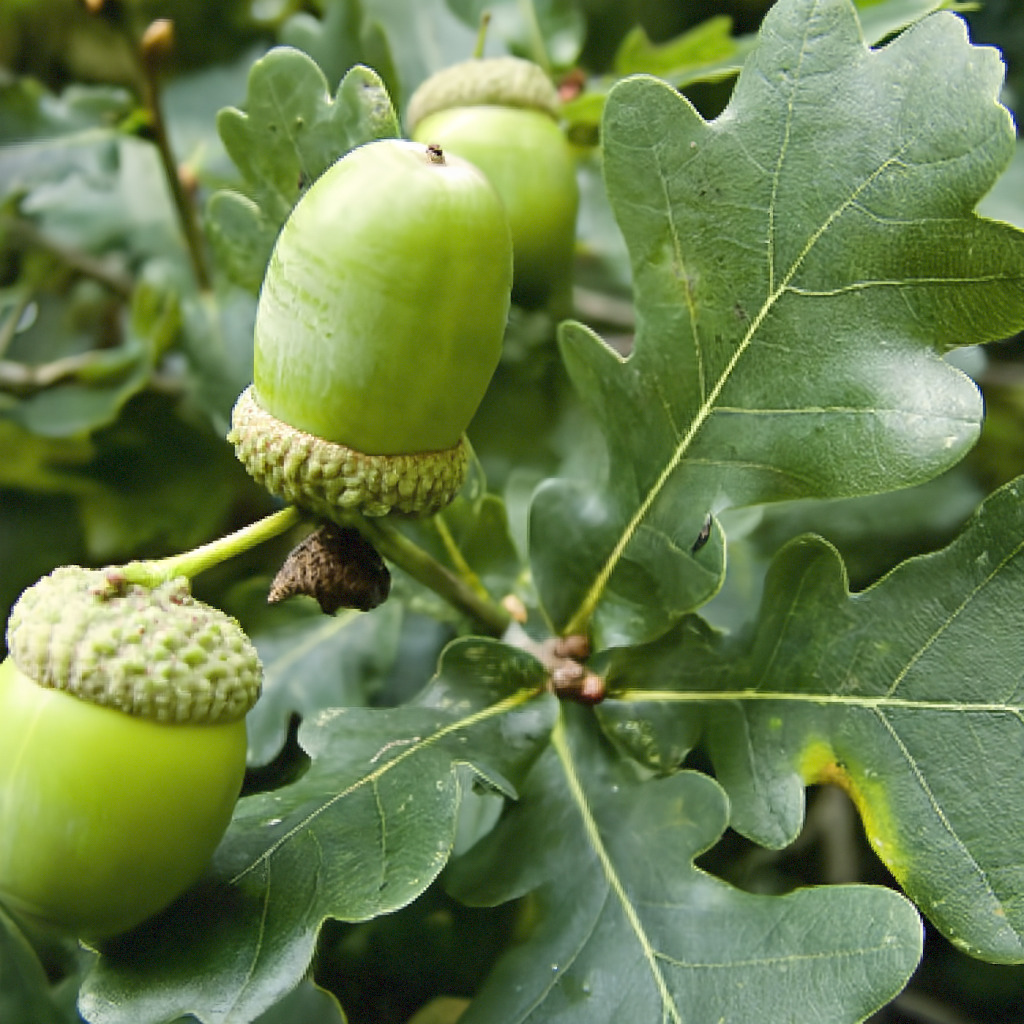}};
            \spy on \spyloc in node [left] at \spyshift;
        \end{tikzpicture} &
        \renewcommand{\spyloc}{(-0.2,-1.2)}
        \begin{tikzpicture}[spy using outlines={red,magnification=\magn,size=\ww}]
            \node {\includegraphics[width=\ww]{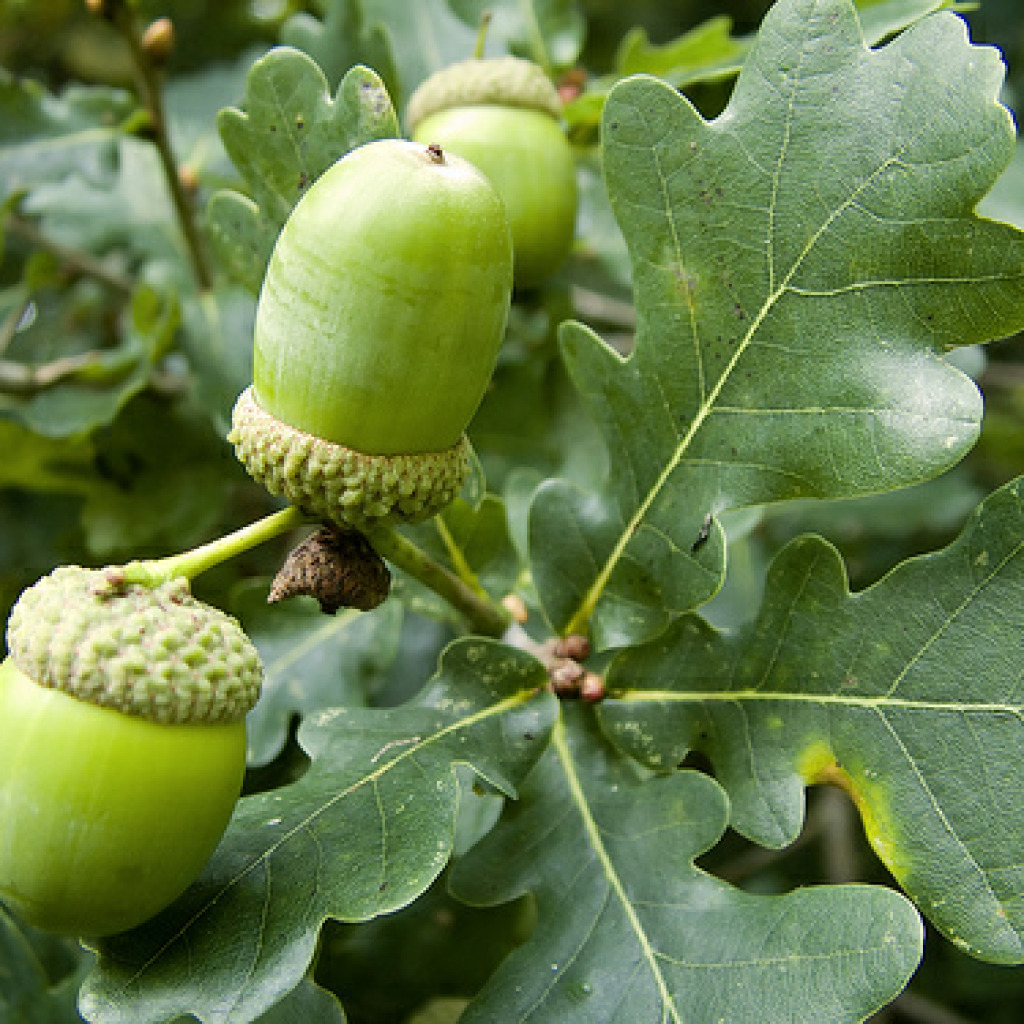}};
            \spy on \spyloc in node [left] at \spyshift;
        \end{tikzpicture} \\

                %% pumpkin 866
        \renewcommand{\spyloc}{(-0.4,0.8)}
        \begin{tikzpicture}[spy using outlines={red,magnification=\magn,size=\ww}]
            \node {\includegraphics[width=\ww]{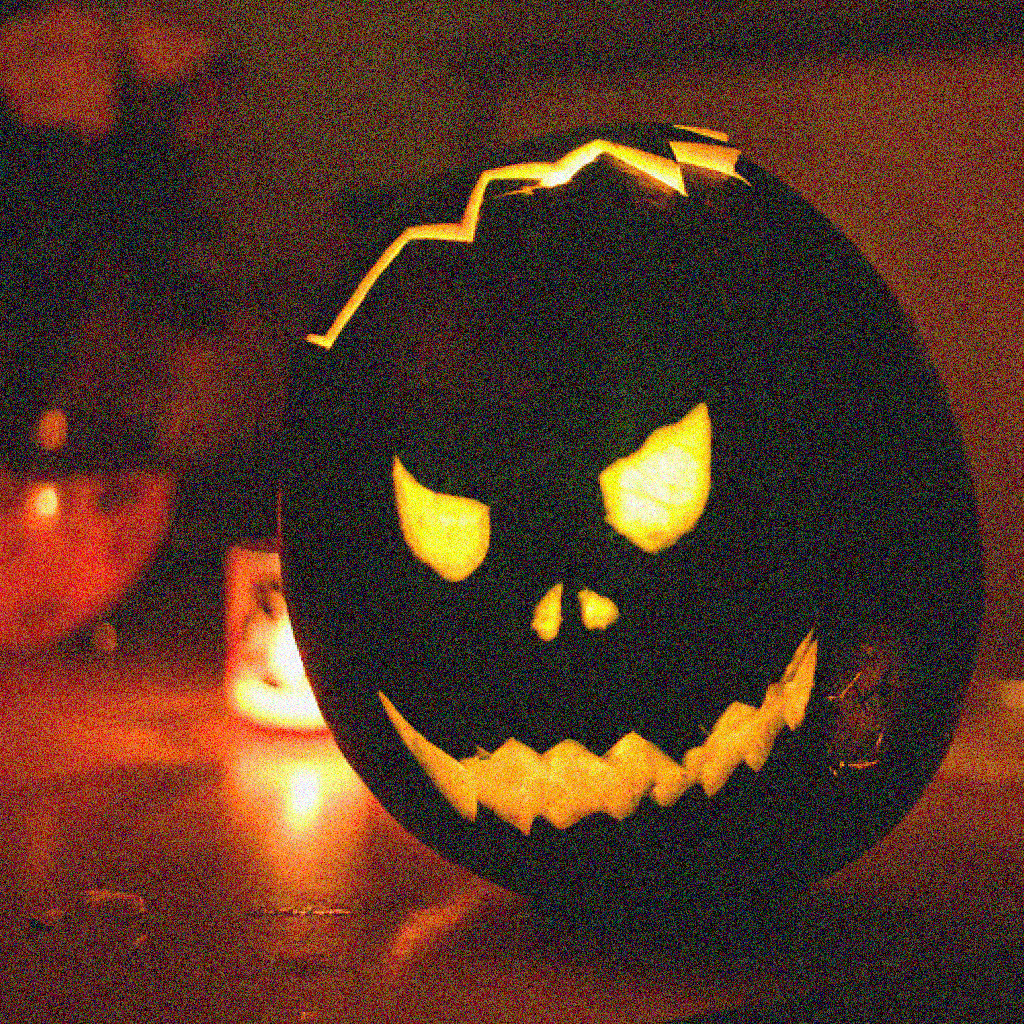}};
            \spy on \spyloc in node [left] at \spyshift;
        \end{tikzpicture} &
        \renewcommand{\spyloc}{(-0.4,0.8)}
        \begin{tikzpicture}[spy using outlines={red,magnification=\magn,size=\ww}]
            \node {\includegraphics[width=\ww]{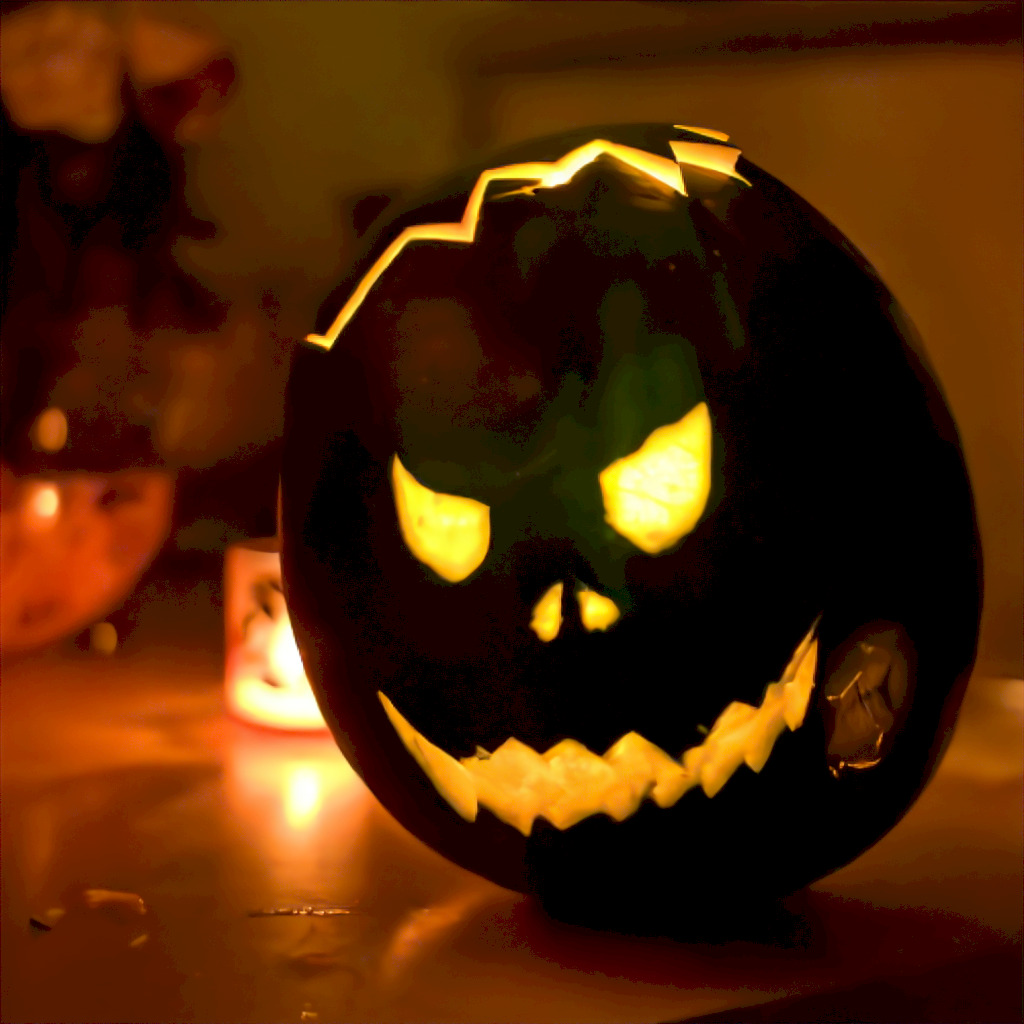}};
            \spy on \spyloc in node [left] at \spyshift;
        \end{tikzpicture} &
        \renewcommand{\spyloc}{(-0.4,0.8)}
        \begin{tikzpicture}[spy using outlines={red,magnification=\magn,size=\ww}]
            \node {\includegraphics[width=\ww]{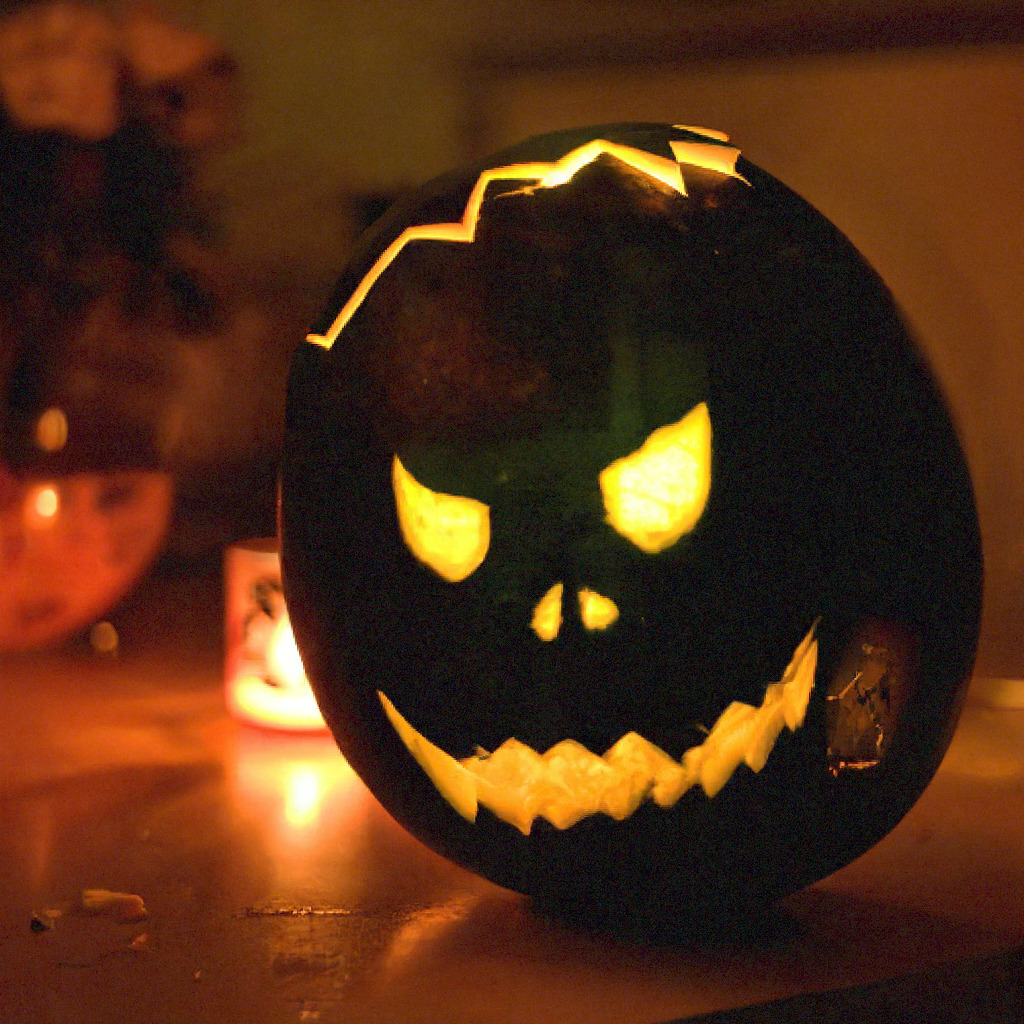}};
            \spy on \spyloc in node [left] at \spyshift;
        \end{tikzpicture} &
        \renewcommand{\spyloc}{(-0.4,0.8)}
        \begin{tikzpicture}[spy using outlines={red,magnification=\magn,size=\ww}]
            \node {\includegraphics[width=\ww]{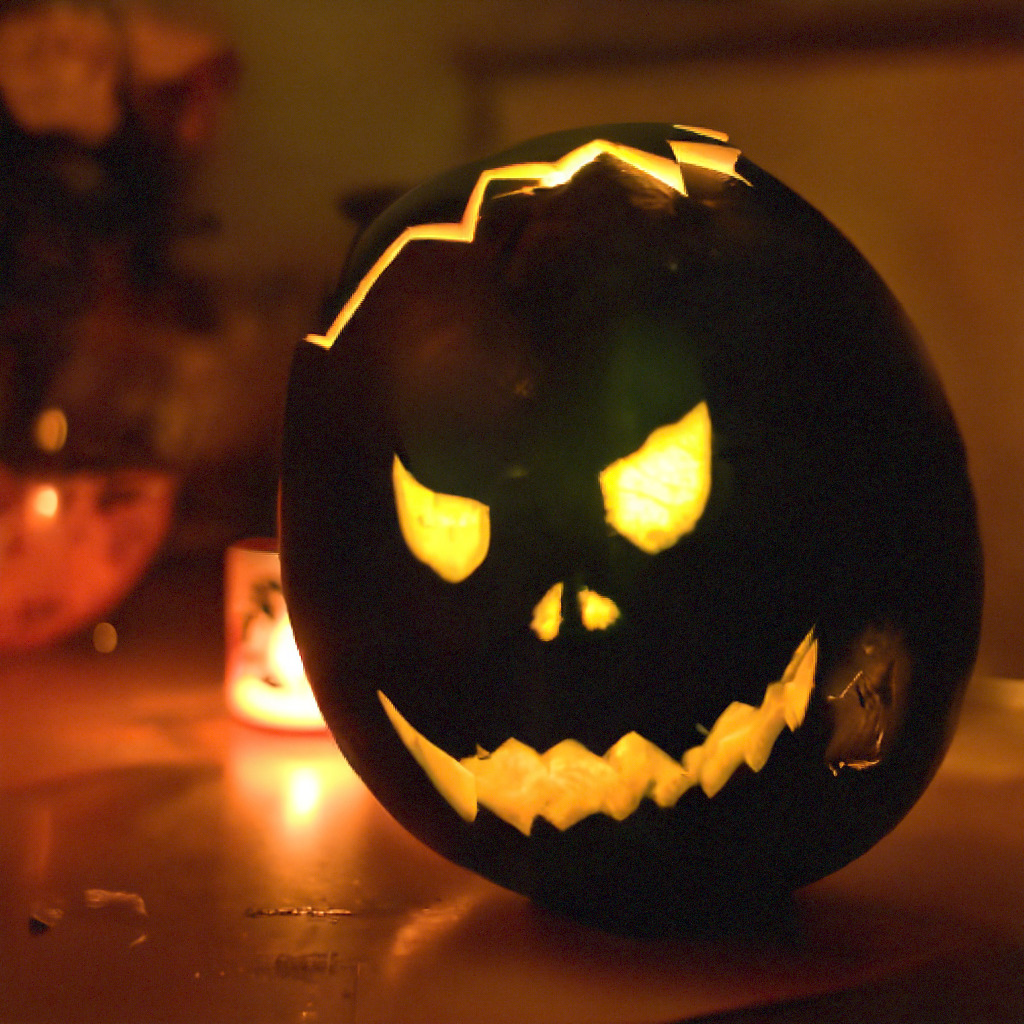}};
            \spy on \spyloc in node [left] at \spyshift;
        \end{tikzpicture} &
        \renewcommand{\spyloc}{(-0.4,0.8)}
        \begin{tikzpicture}[spy using outlines={red,magnification=\magn,size=\ww}]
            \node {\includegraphics[width=\ww]{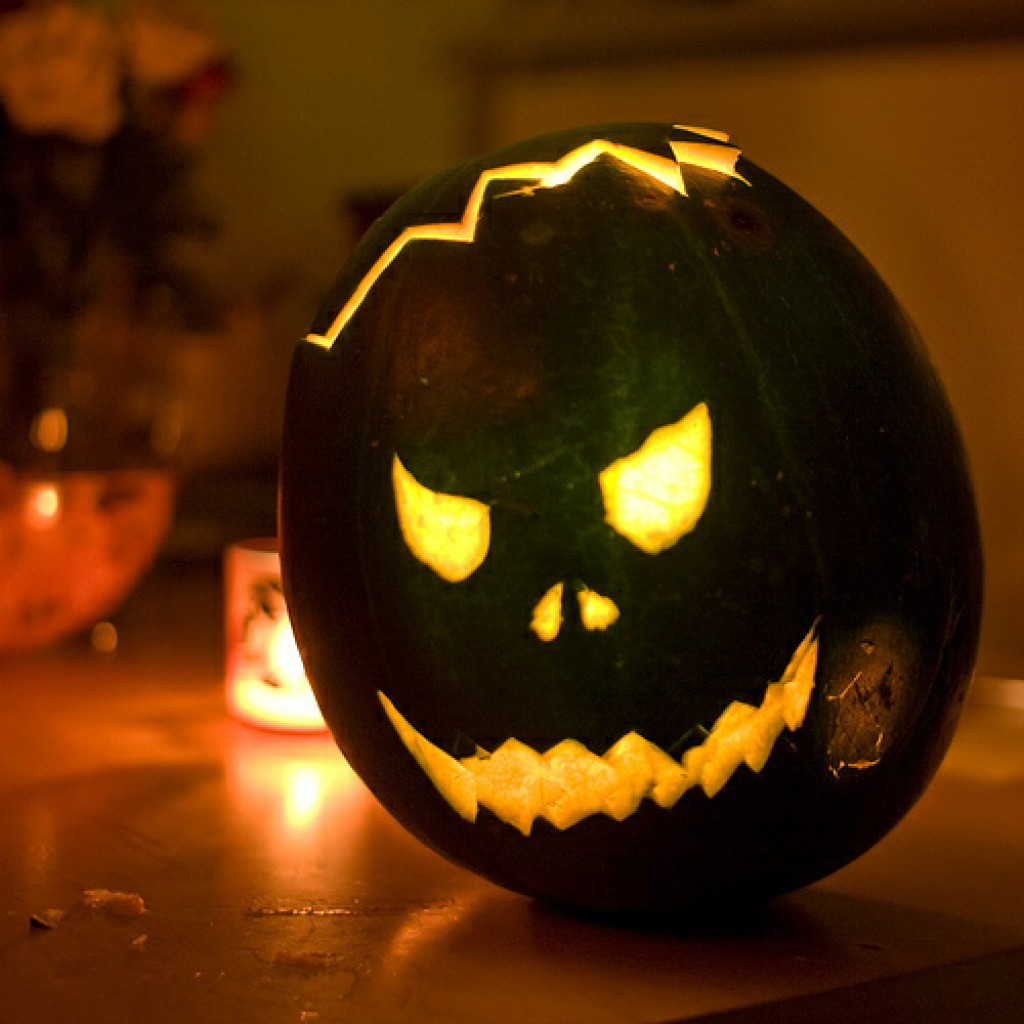}};
            \spy on \spyloc in node [left] at \spyshift;
        \end{tikzpicture} \\
\end{tabular}
\end{center}
\vspace{-1em}
\caption{Comparison between different denoising methods on images with noise gain of 4.} 
\label{fig:imagenet_comparison_4_1}
\end{figure}

\begin{figure}[htbp]
\setlength{\ww}{0.192\textwidth}
\begin{center}
\newcommand{\magn}{5.0}
\newcommand{\spyloc}{(0.28,0.15)}    
\newcommand{\spyshift}{(1.675,-3.4)}
\newcommand{\bright}{0.1 1 0.1 1 0.1 1}
\small\addtolength{\tabcolsep}{-8.5pt}
\begin{tabular}{ccccc}
    Noisy & HINet~\cite{chen2021hinet} & Baseline & Ours & Clean GT\\
        %% pepper 1869
        \renewcommand{\spyloc}{(-0.4,-0.8)}
        \begin{tikzpicture}[spy using outlines={red,magnification=\magn,size=\ww}]
            \node {\includegraphics[width=\ww]{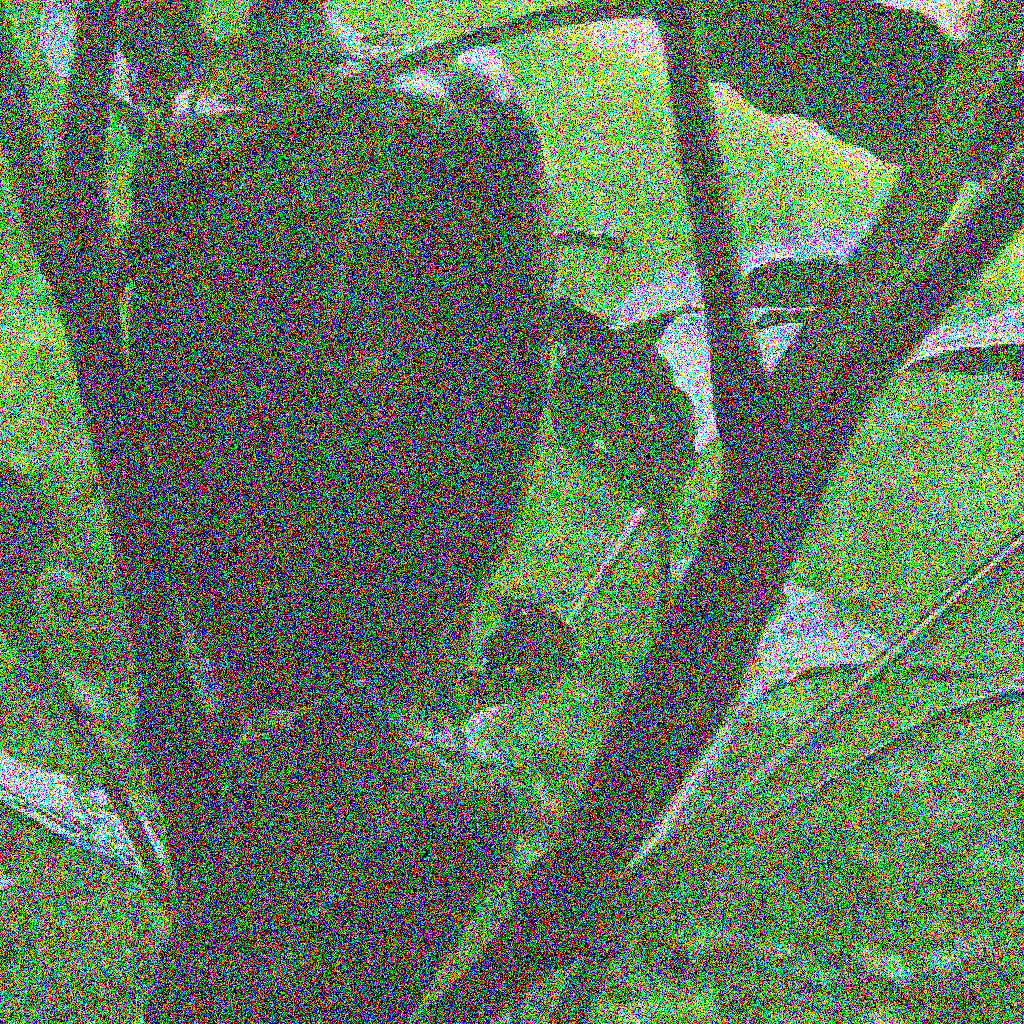}};
            \spy on \spyloc in node [left] at \spyshift;
        \end{tikzpicture} &
        \renewcommand{\spyloc}{(-0.4,-0.8)}
        \begin{tikzpicture}[spy using outlines={red,magnification=\magn,size=\ww}]
            \node {\includegraphics[width=\ww]{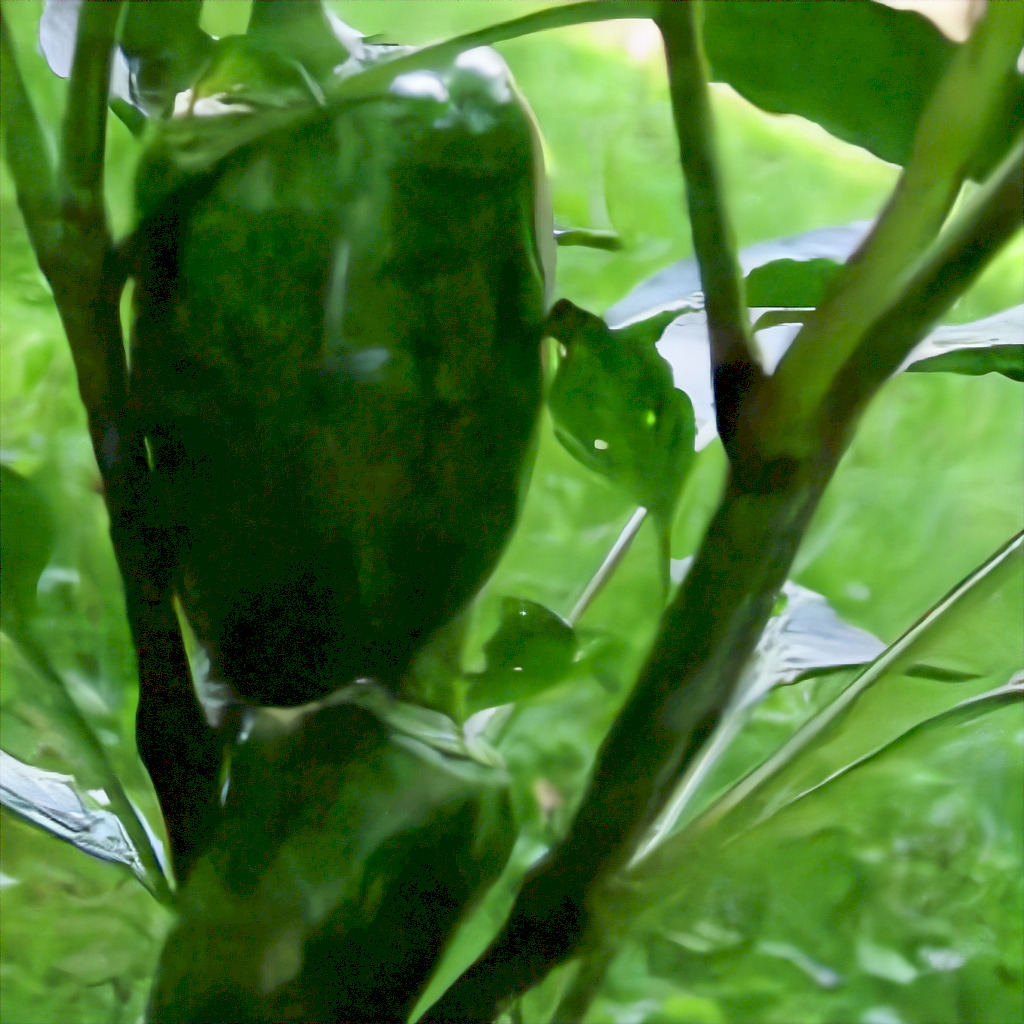}};
            \spy on \spyloc in node [left] at \spyshift;
        \end{tikzpicture} &
        \renewcommand{\spyloc}{(-0.4,-0.8)}
        \begin{tikzpicture}[spy using outlines={red,magnification=\magn,size=\ww}]
            \node {\includegraphics[width=\ww]{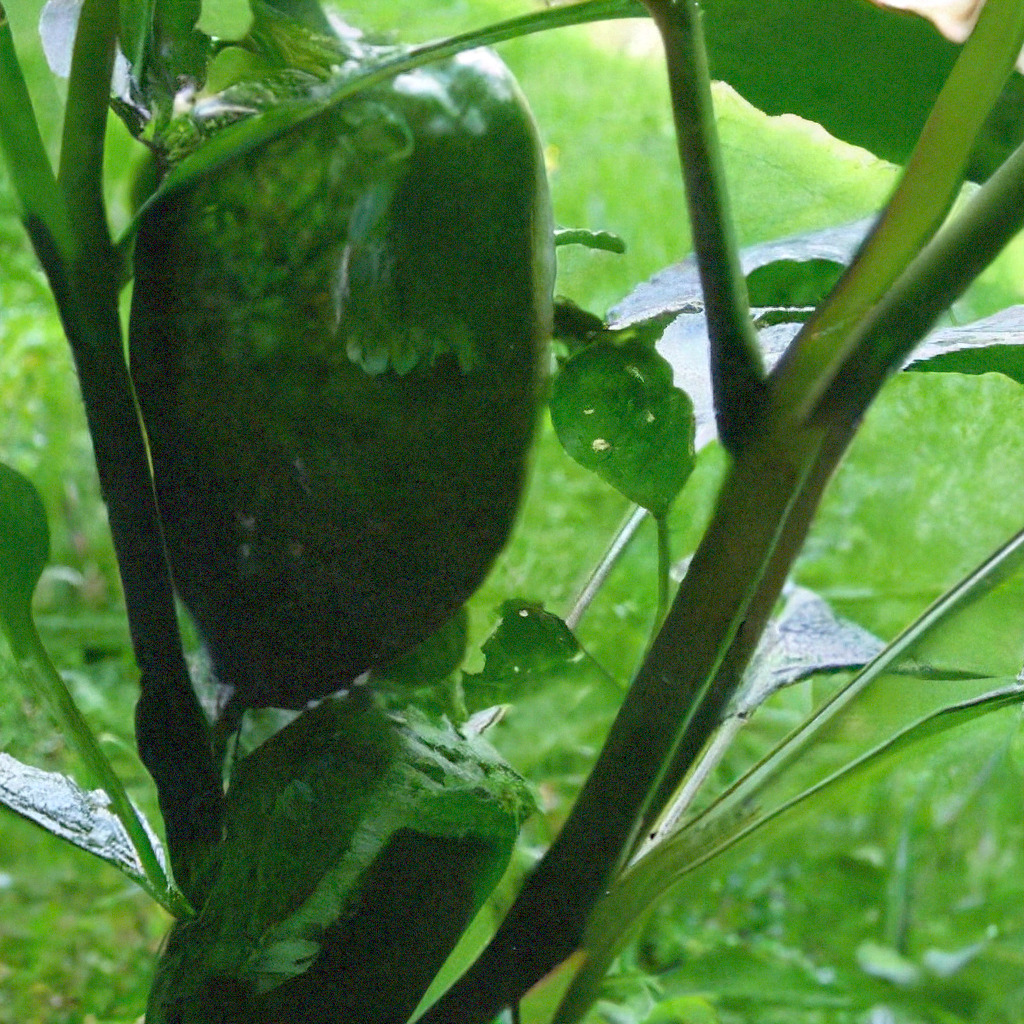}};
            \spy on \spyloc in node [left] at \spyshift;
        \end{tikzpicture} &
        \renewcommand{\spyloc}{(-0.4,-0.8)}
        \begin{tikzpicture}[spy using outlines={red,magnification=\magn,size=\ww}]
            \node {\includegraphics[width=\ww]{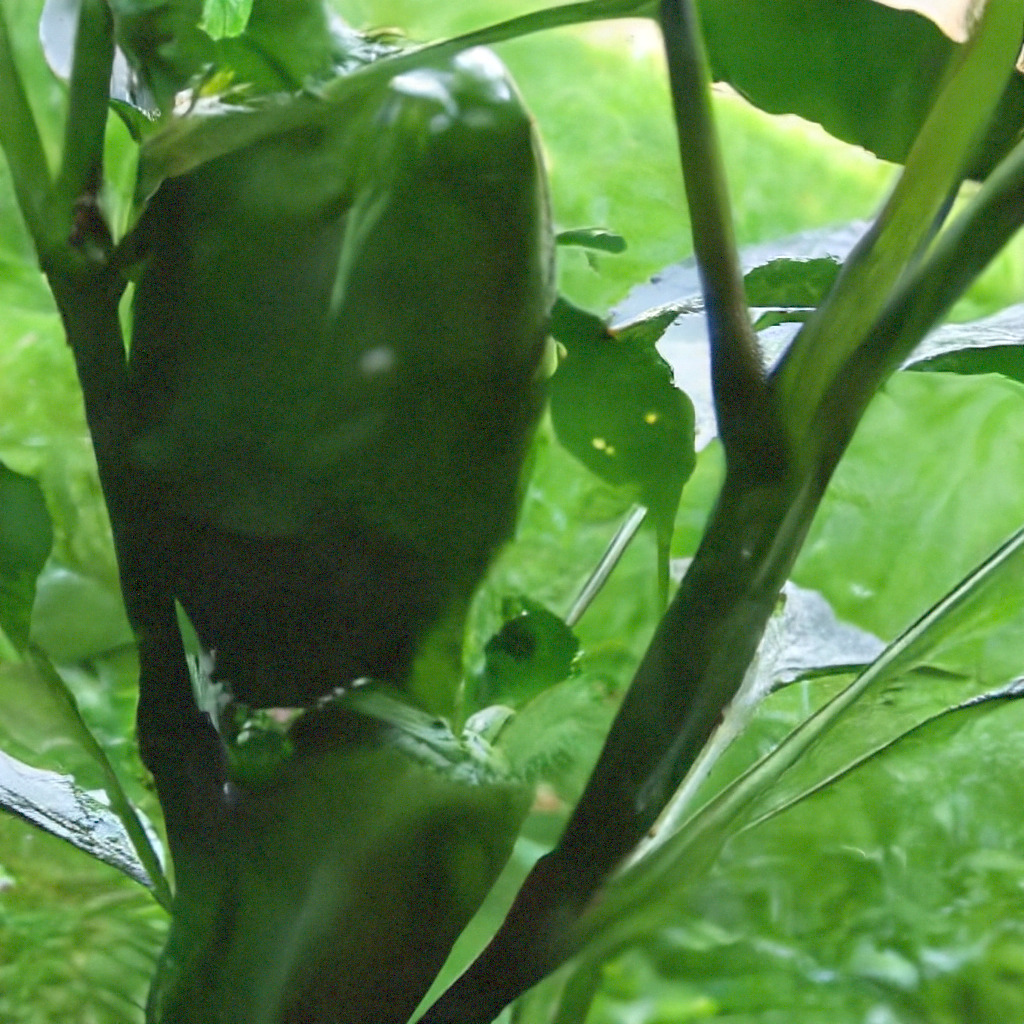}};
            \spy on \spyloc in node [left] at \spyshift;
        \end{tikzpicture} &
        \renewcommand{\spyloc}{(-0.4,-0.8)}
        \begin{tikzpicture}[spy using outlines={red,magnification=\magn,size=\ww}]
            \node {\includegraphics[width=\ww]{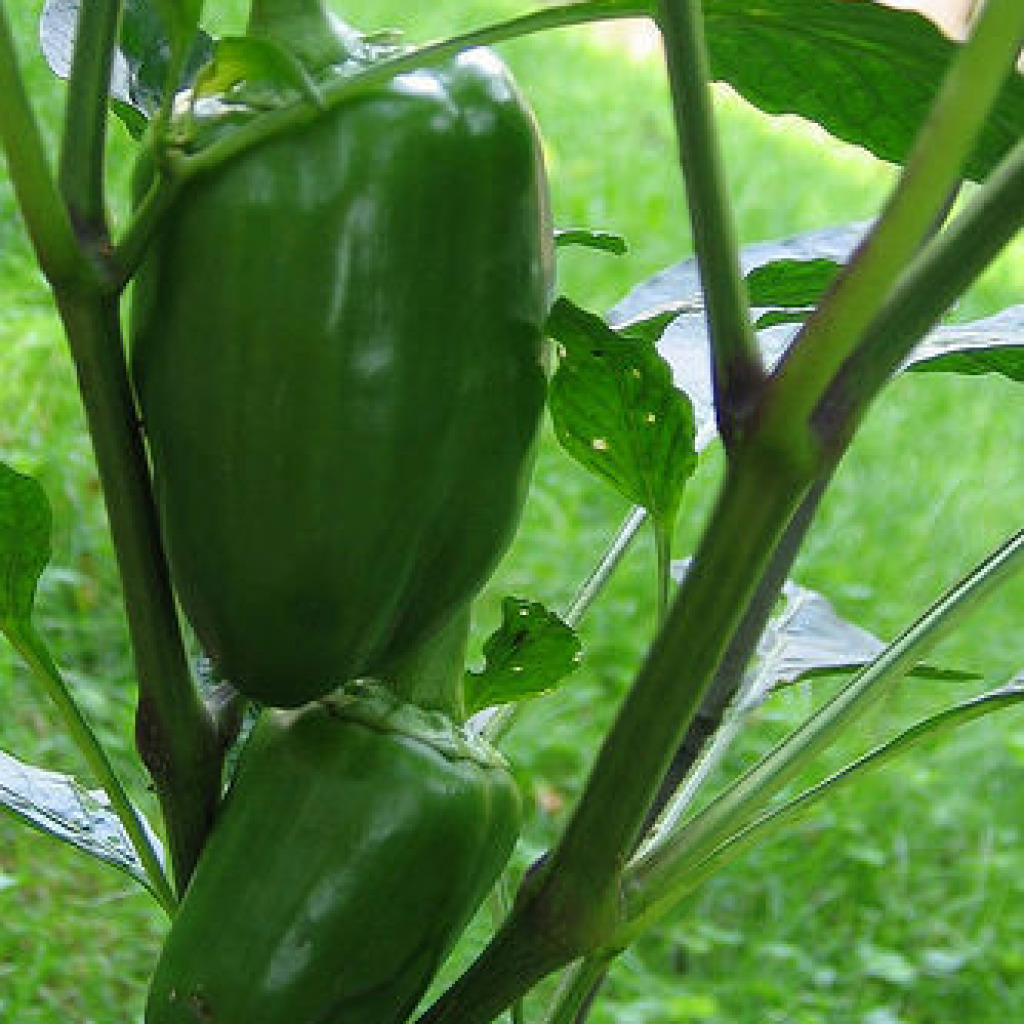}};
            \spy on \spyloc in node [left] at \spyshift;
        \end{tikzpicture} \\

        %% dog 1895
        \renewcommand{\spyloc}{(-0.7,0.6)}
        \begin{tikzpicture}[spy using outlines={red,magnification=\magn,size=\ww}]
            \node {\includegraphics[width=\ww]{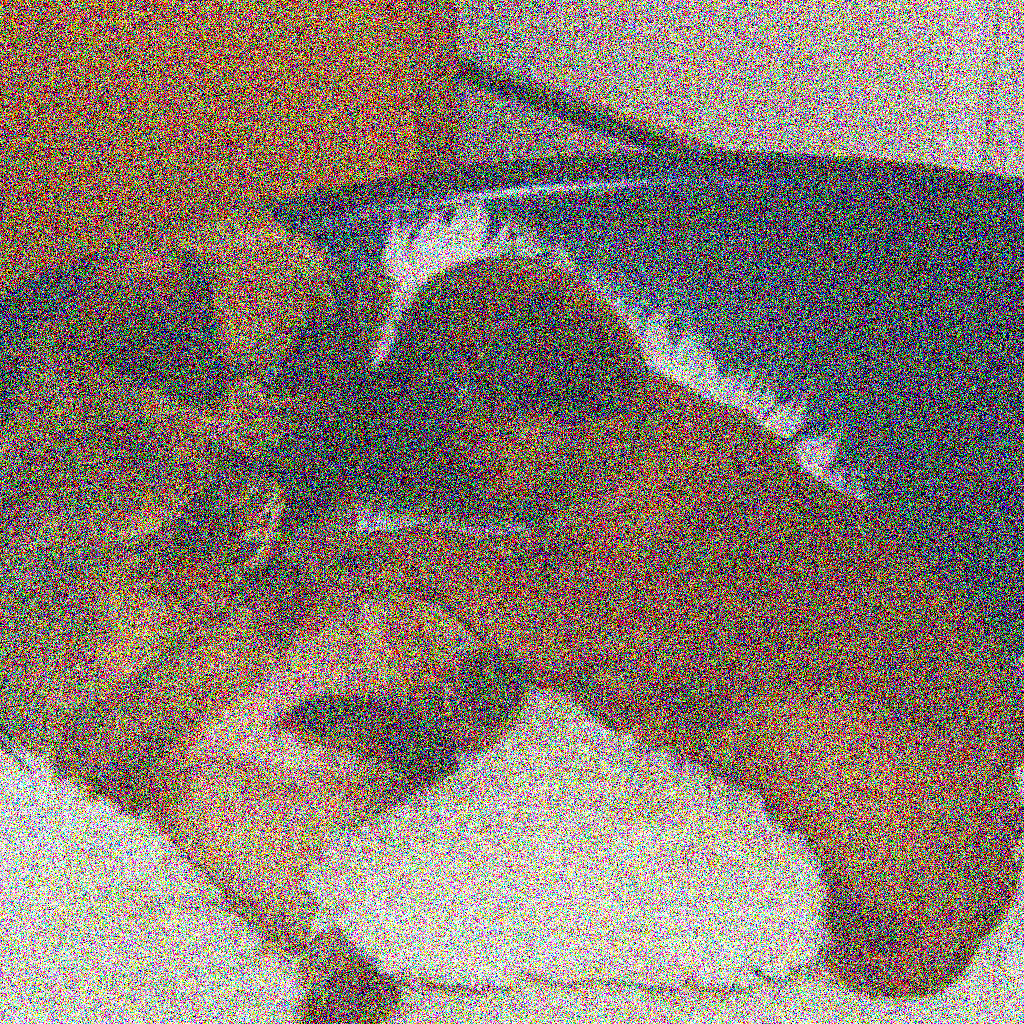}};
            \spy on \spyloc in node [left] at \spyshift;
        \end{tikzpicture} &
        \renewcommand{\spyloc}{(-0.7,0.6)}
        \begin{tikzpicture}[spy using outlines={red,magnification=\magn,size=\ww}]
            \node {\includegraphics[width=\ww]{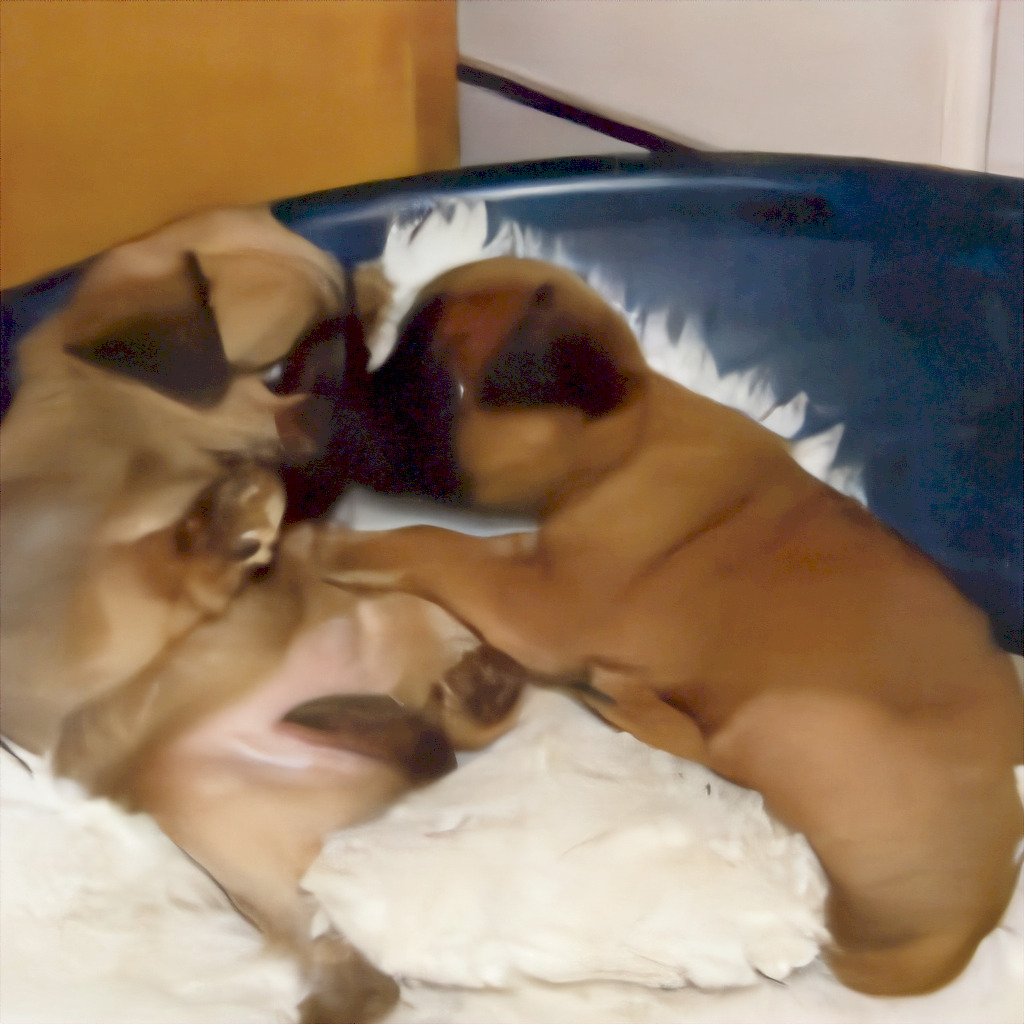}};
            \spy on \spyloc in node [left] at \spyshift;
        \end{tikzpicture} &
        \renewcommand{\spyloc}{(-0.7,0.6)}
        \begin{tikzpicture}[spy using outlines={red,magnification=\magn,size=\ww}]
            \node {\includegraphics[width=\ww]{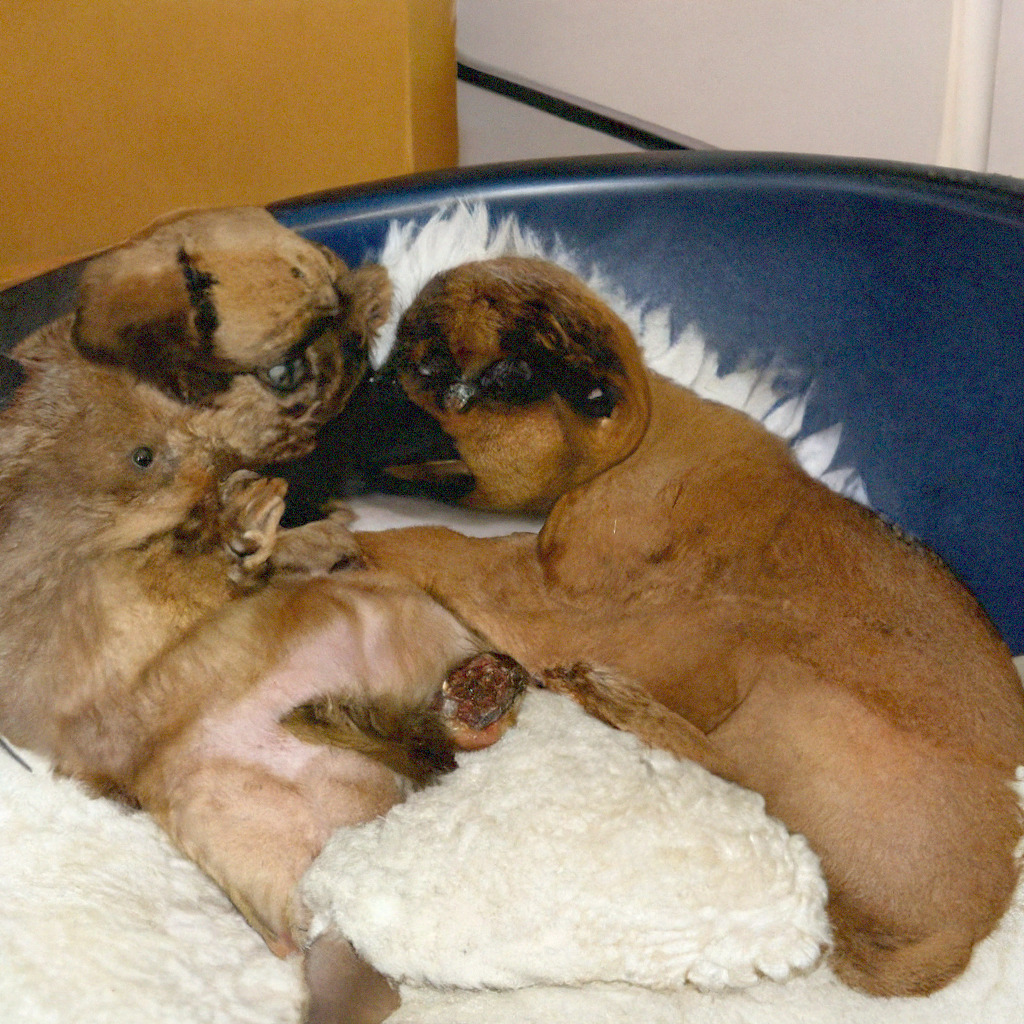}};
            \spy on \spyloc in node [left] at \spyshift;
        \end{tikzpicture} &
        \renewcommand{\spyloc}{(-0.7,0.6)}
        \begin{tikzpicture}[spy using outlines={red,magnification=\magn,size=\ww}]
            \node {\includegraphics[width=\ww]{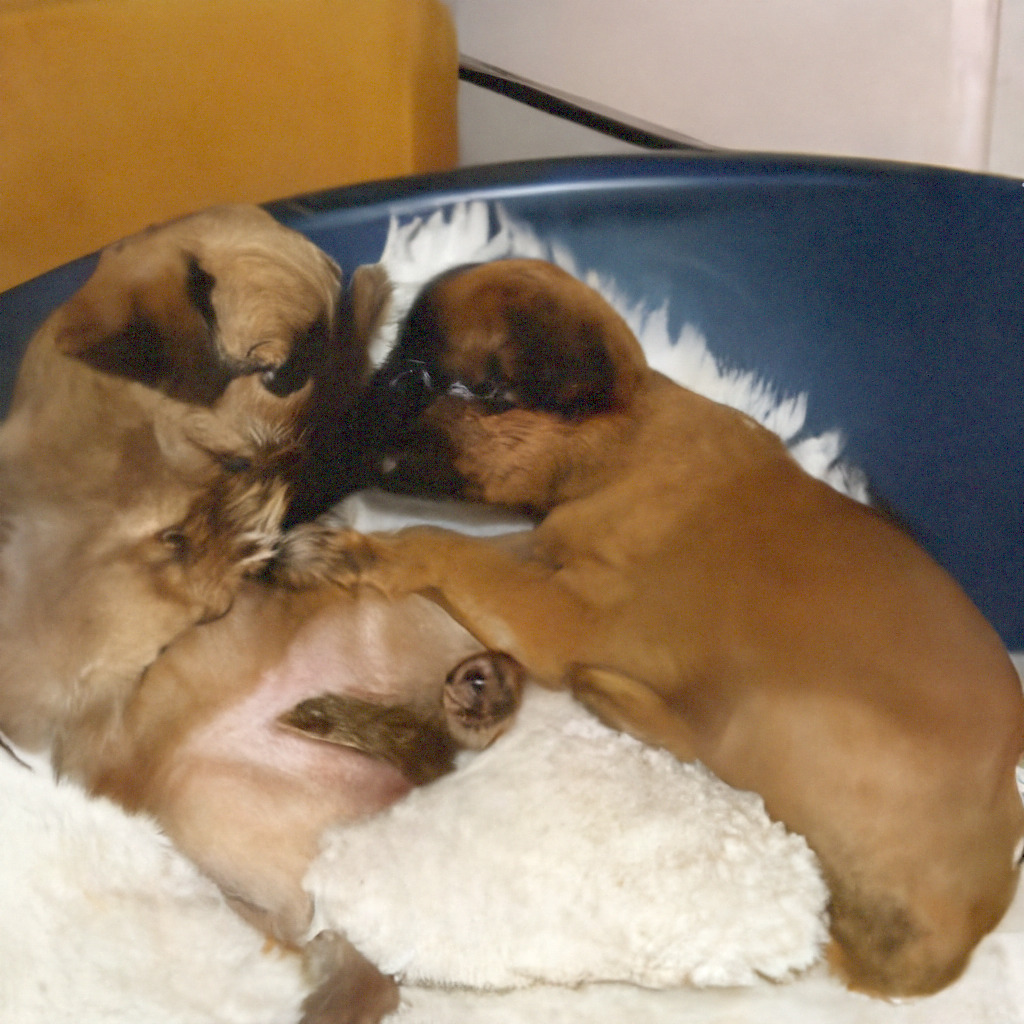}};
            \spy on \spyloc in node [left] at \spyshift;
        \end{tikzpicture} &
        \renewcommand{\spyloc}{(-0.7,0.6)}
        \begin{tikzpicture}[spy using outlines={red,magnification=\magn,size=\ww}]
            \node {\includegraphics[width=\ww]{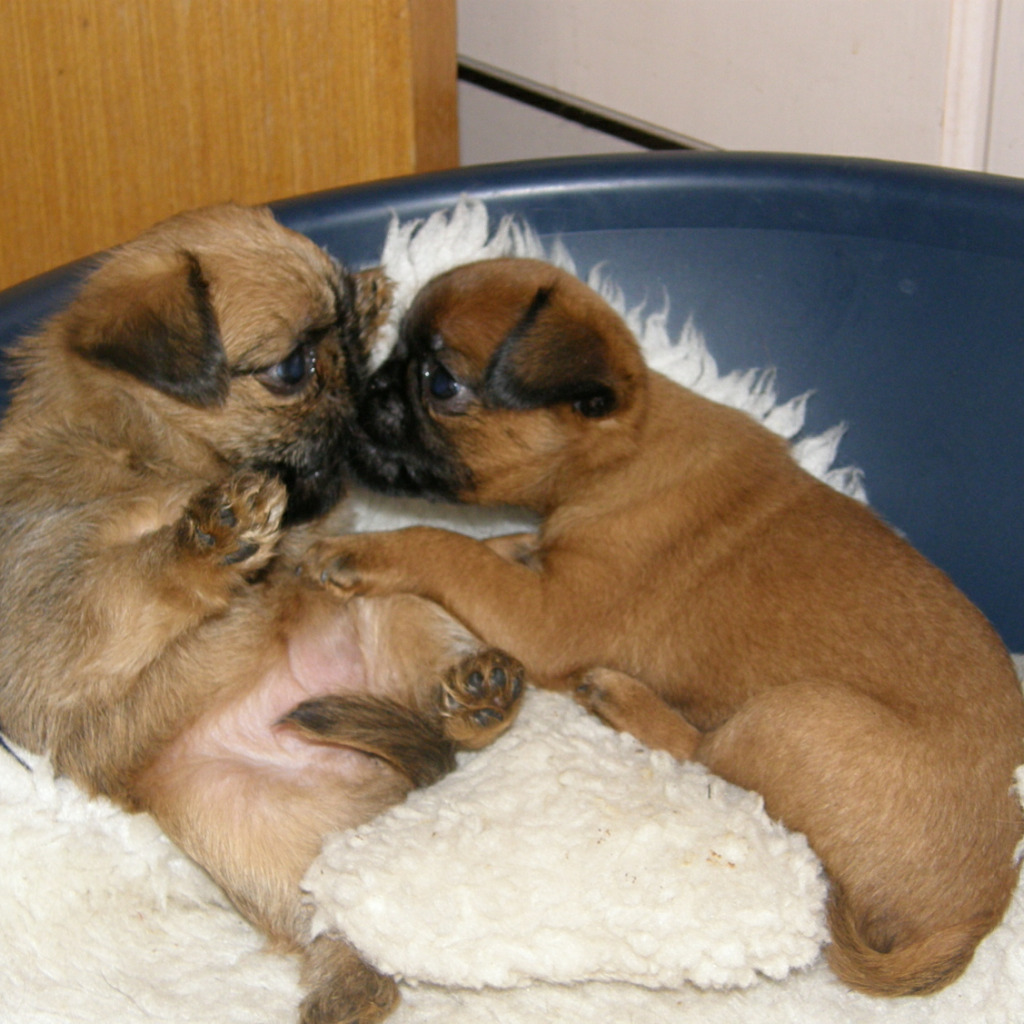}};
            \spy on \spyloc in node [left] at \spyshift;
        \end{tikzpicture} \\

        %% chamilion 80
        \renewcommand{\spyloc}{(0,-1.2)}
        \begin{tikzpicture}[spy using outlines={red,magnification=\magn,size=\ww}]
            \node {\includegraphics[width=\ww]{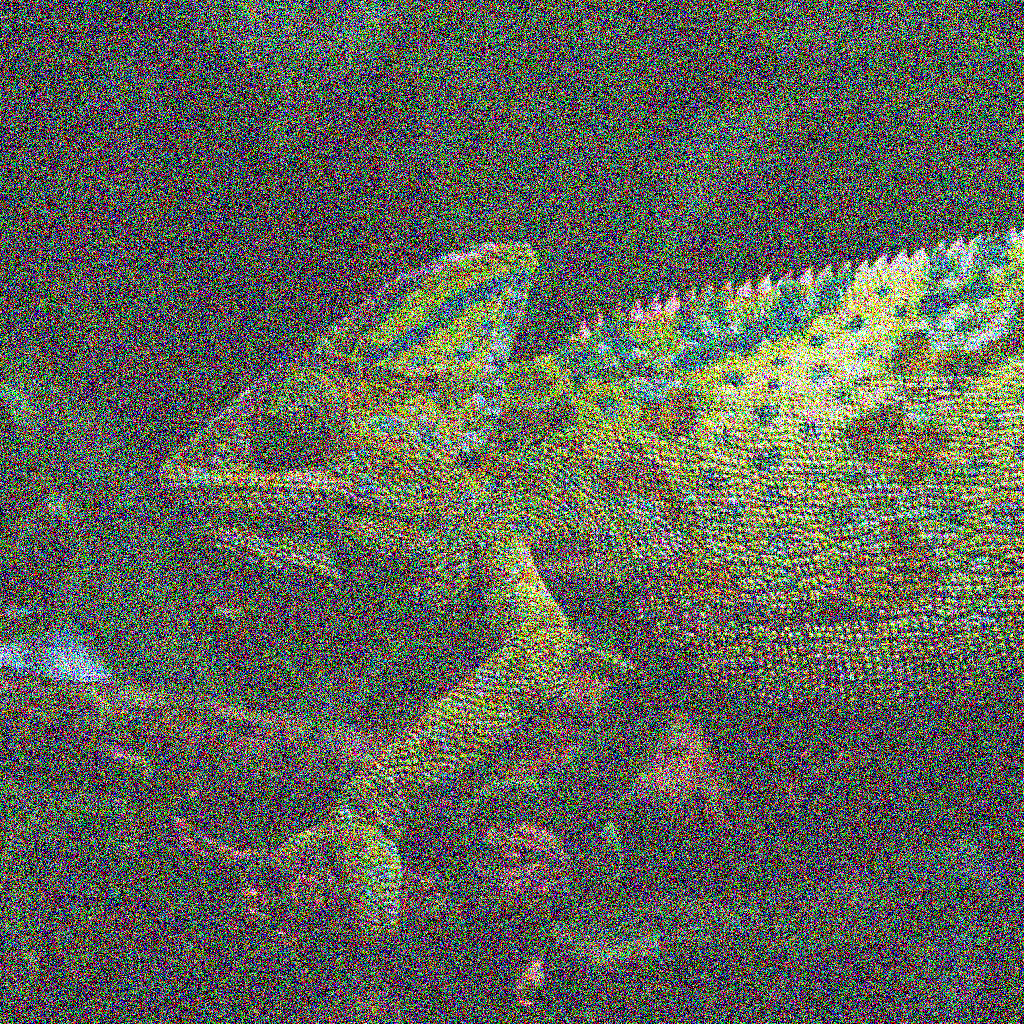}};
            \spy on \spyloc in node [left] at \spyshift;
        \end{tikzpicture} &
        \renewcommand{\spyloc}{(0,-1.2)}
        \begin{tikzpicture}[spy using outlines={red,magnification=\magn,size=\ww}]
            \node {\includegraphics[width=\ww]{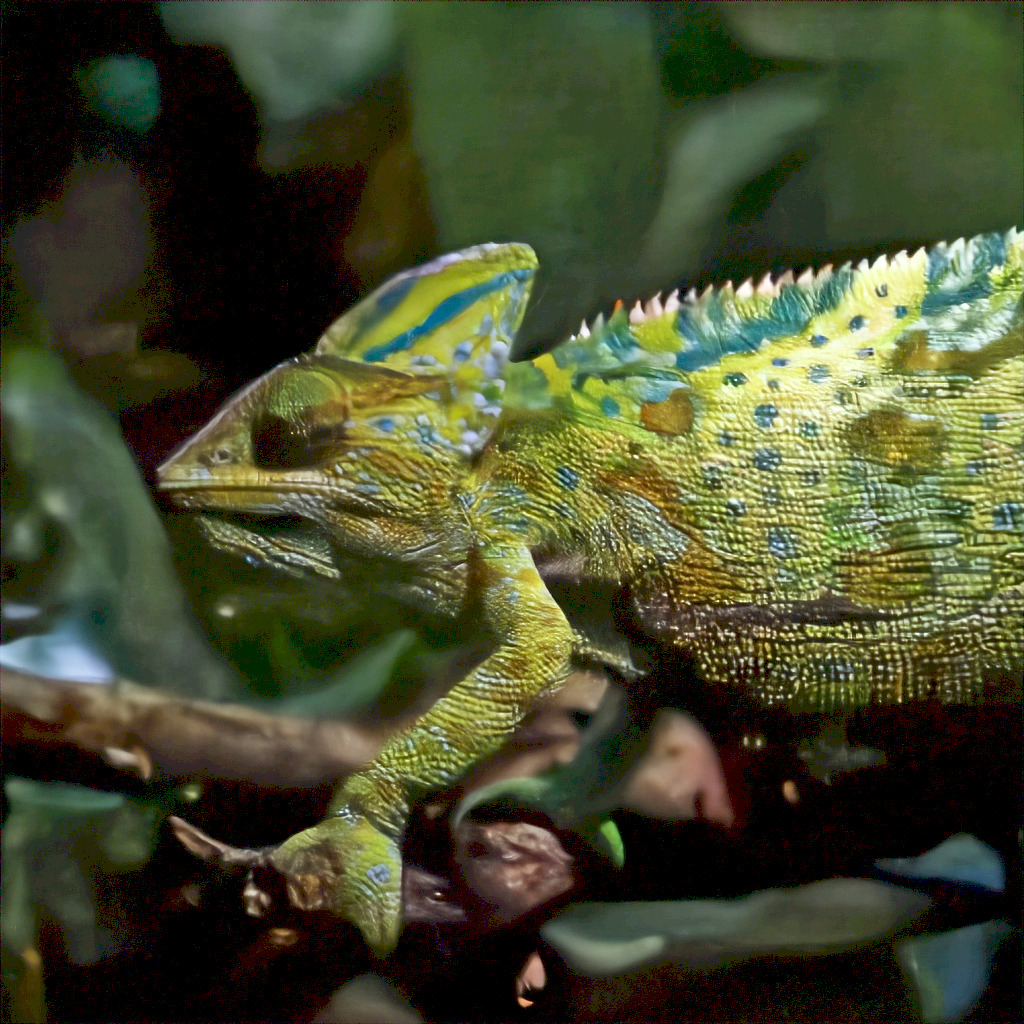}};
            \spy on \spyloc in node [left] at \spyshift;
        \end{tikzpicture} &
        \renewcommand{\spyloc}{(0,-1.2)}
        \begin{tikzpicture}[spy using outlines={red,magnification=\magn,size=\ww}]
            \node {\includegraphics[width=\ww]{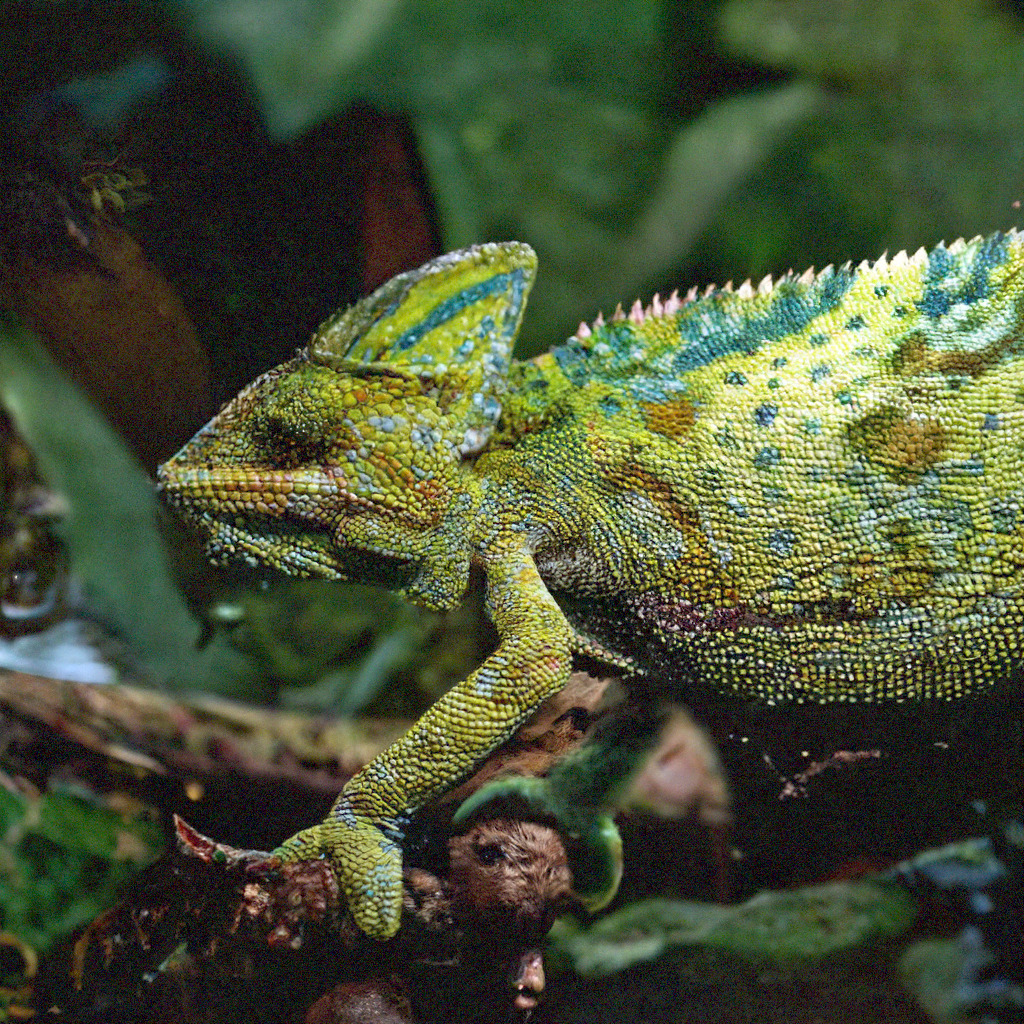}};
            \spy on \spyloc in node [left] at \spyshift;
        \end{tikzpicture} &
        \renewcommand{\spyloc}{(0,-1.2)}
        \begin{tikzpicture}[spy using outlines={red,magnification=\magn,size=\ww}]
            \node {\includegraphics[width=\ww]{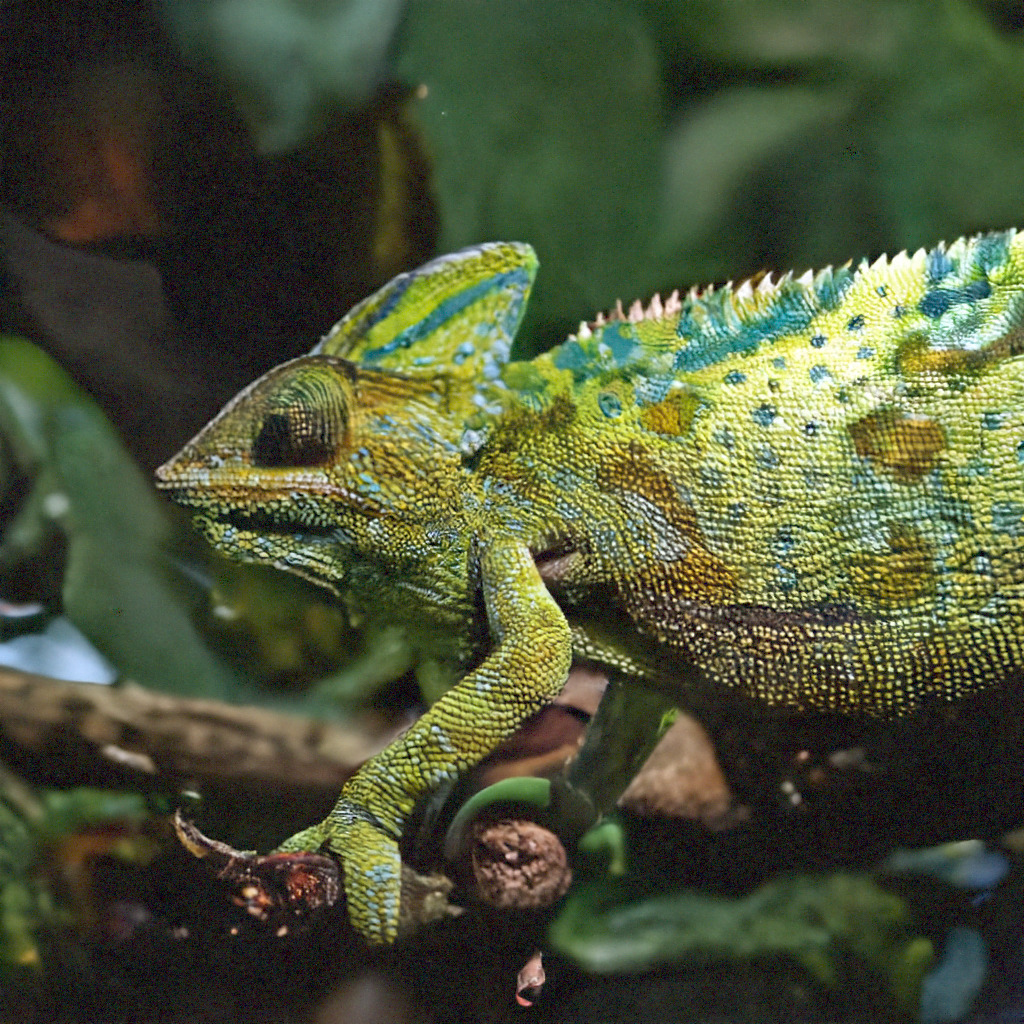}};
            \spy on \spyloc in node [left] at \spyshift;
        \end{tikzpicture} &
        \renewcommand{\spyloc}{(0,-1.2)}
        \begin{tikzpicture}[spy using outlines={red,magnification=\magn,size=\ww}]
            \node {\includegraphics[width=\ww]{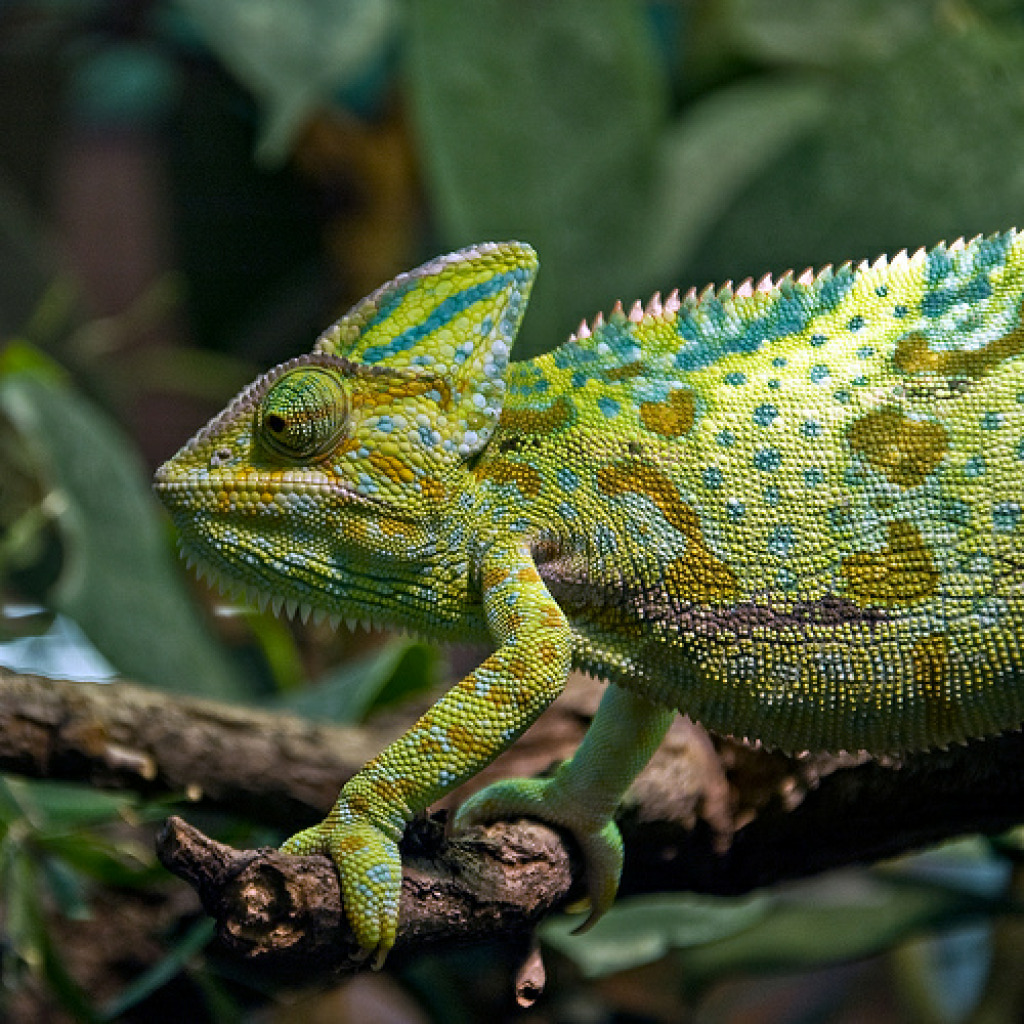}};
            \spy on \spyloc in node [left] at \spyshift;
        \end{tikzpicture} \\
\end{tabular}
\end{center}
\vspace{-1em}
\caption{Comparison between different denoising methods on images with noise gain of 20.} 
\label{fig:imagenet_comparison_20}
\end{figure}

In the lowest noise level (\figref{fig:imagenet_comparison_1}), the noise is mild and all models give comparable results. However, in darker areas (like the owl feathers), one can still identify some over-smootihng in the result of HINet, and some remaining noise in the baseline result. 
In the middle noise level (\figref{fig:imagenet_comparison_4_1}), these artifacts are visible across all images, in both darker and brighter areas. We see that our model manages to balance between the generation of intricate details and the elimination of the noise. These phenomena are most evident in the highest noise level, depicted in \figref{fig:imagenet_comparison_20}.

\end{document}